%% file: Template.tex
\documentclass[journal]{IEEEtran}
\ifCLASSINFOpdf
\usepackage{cite}
\usepackage{hyperref}
\usepackage{dblfloatfix} 
\usepackage[pdftex]{graphicx}
\usepackage[export]{adjustbox}
\usepackage{kantlipsum}

\graphicspath{{Figures/}}
\else
\fi
\usepackage{amsmath}
\usepackage{amssymb}
\usepackage{amsthm}
\usepackage{mathtools}
\usepackage{stackengine}
\usepackage{multicol}
\usepackage{diagbox}
\usepackage{graphicx}
\usepackage{ragged2e}
\usepackage{tabularx}

\newcolumntype{s}{>{\hsize=.1\hsize}X}
\newcolumntype{f}{>{\hsize=.2\hsize}X}

\makeatletter
\def\thickhline{%
 \noalign{\ifnum0=`}\fi\hrule \@height \thickarrayrulewidth \futurelet
 \reserved@a\@xthickhline}
\def\@xthickhline{\ifx\reserved@a\thickhline
 \vskip\doublerulesep
 \vskip-\thickarrayrulewidth
 \fi
 \ifnum0=`{\fi}}
\makeatother

\newlength{\thickarrayrulewidth}
\usepackage{arydshln}
\usepackage[skip=5pt]{caption}

\DeclareMathOperator*{\argmin}{arg\,min}

\usepackage{hyperref}
\hypersetup{
	colorlinks,
	linkcolor={green!70!black},
	citecolor={green!70!black},
	urlcolor={green!70!black}
}

\usepackage{algorithmic, algorithm}

\usepackage{booktabs}

\usepackage{calc}
\usepackage{xcolor}

\definecolor{mygreen}{RGB}{152, 191, 140}
\definecolor{mypink}{RGB}{233, 127, 213}
\definecolor{myyellow}{RGB}{255, 229, 127}
\definecolor{myorange}{RGB}{255, 127, 127}

\newcommand\dunderline[3][-1pt]{{%
 \sbox0{#3}%
 \ooalign{\copy0\cr\rule[\dimexpr#1-#2\relax]{\wd0}{#2}}}}

\usepackage{dsfont}
\label{comments}

\numberwithin{theorem}{section} 

\usepackage{booktabs}
\usepackage{multirow}
\definecolor{myBlue}{RGB}{219, 48, 122}

\usepackage{pgf,tikz,pgfplots}
\usetikzlibrary{arrows,shapes}
\usetikzlibrary{plotmarks}
\usepgfplotslibrary{groupplots}
\pgfplotsset{compat=newest}
\pgfplotsset{plot coordinates/math parser=false}
\usetikzlibrary{positioning,spy}
\usetikzlibrary{backgrounds}
\usepackage{comment}
\usepackage{float}

\newcommand{\dcheck}[1]{{\color{black}{#1}}}

\begin{document}

\title{Mixture-Net: Low-Rank Deep Image Prior Inspired by Mixture Models for Spectral Image Recovery}

\author{Tatiana Gelvez-Barrera,~\IEEEmembership{Member,~IEEE} and Jorge Bacca,~\IEEEmembership{Member,~IEEE,} and Henry Arguello,~\IEEEmembership{Senior~Member,~IEEE}
\thanks{T.Gelvez-Barrera and J.Bacca both are co-first authors with equal contribution. T. Gelvez-Barrera is with the Department of Electrical Engineering, Universidad Industrial de Santander, Bucaramanga, Colombia, 680002, e-mail: tatiana.gelvez@correo.uis.edu.co. J. Bacca and H. Arguello are with the Department of Computer Science,
Universidad Industrial de Santander, Bucaramanga, Colombia, 680002 e-mail:
jorge.bacca1@correo.uis.edu.co; henarfu@uis.edu.co. This work was supported by the Sistema General de Regal\'ias (SGR) under Project 8933.}}

\markboth{IEEE TRANSACTIONS ON IMAGE PROCESSING}%
{Shell \MakeLowercase{\textit{et al.}}: Bare Demo of IEEEtran.cls for IEEE Journals}
	
\maketitle


\begin{abstract}
This paper proposes a non-data-driven deep neural network for spectral image recovery problems such as denoising, single hyperspectral image super-resolution, and compressive spectral imaging reconstruction. Unlike previous methods, the proposed approach, dubbed Mixture-Net, implicitly learns the prior information through the network. Mixture-Net consists of a deep generative model whose layers are inspired by the linear and non-linear low-rank mixture models, where the recovered image is composed of a weighted sum between the linear and non-linear decomposition. Mixture-Net also provides a low-rank decomposition interpreted as the spectral image abundances and endmembers, helpful in achieving \dcheck{remote sensing tasks} without running additional routines. The experiments show the Mixture-Net effectiveness outperforming state-of-the-art methods in recovery quality with the advantage of architecture interpretability.
\end{abstract}
	

\begin{IEEEkeywords}
Spectral Image Recovery, Model-based Optimization, Deep Image Prior, Low-rank Spectral Mixture Models.
\end{IEEEkeywords}


\section{Introduction}
Spectral imaging sensors acquire spatial-spectral information of a scene, organized in a datacube known as a spectral image (SI), which is helpful in remote sensing applications such as unmixing and material identification~\cite{vargas2019spectral}. The SI acquisition through specialized optical systems can be affected by artifacts linked with external conditions and optical aberrations that corrupt, degrade, or diminish the number of measurements. Therefore, recovering the underlying scene is a critical step in the SI analysis~\cite{amigo2020preprocessing}. 
	
The atmospheric conditions, light source nature, and photon efficiency can add noise, outlines, and striping, yielding a denoising problem~\cite{ cao2019hyperspectral, xue2019nonlocal}. The technology limitation inhibits the acquisition of high spatial-spectral resolution images, leading to a hyperspectral super-resolution (HSI-SR) problem~\cite{guilloteau2020hyperspectral, zhang2020deep}. The compressive spectral imaging (CSI) paradigm provides a small number of projected measurements, yielding a reconstruction problem~\cite{bacca2021deep,kar2019compressive}. Using prior information is crucial in current recovery approaches to solve the aforementioned ill-posed problems, including \textit{greedy} algorithms, \textit{model-based} optimization, \textit{data-driven} and \textit{learning-based} strategies.
	
Greedy algorithms include the prior information by following a sequence of reasonable steps, where only locally optimal solutions can be guaranteed~\cite{blumensath2012greedy}.
Meanwhile, model-based optimization considers the prior information by designing hand-crafted regularizers to reduce the feasible search set and find globally optimal solutions~\cite{mullah2020fast,gelvez2017joint,rasti2013hyperspectral}.
For instance, the low-rank regularizer promotes low-dimensionality by modeling the SI as a low-rank structure~\cite{gelvez2022joint, rasti2017automatic, bacca2019noniterative,dian2018hyperspectral}. Nonetheless, hand-crafted regularizers are often insufﬁcient to handle the wide spectral information variety. Next, data-driven methods such as deep-learning (DL) learn the prior information by training a black-box non-linear mapping from a dataset. The black-box nature has been tackled by connecting model-based iterative algorithms and deep neural networks~\cite{wang2020dnu,ramirez2021ladmm,vu2020unrolling}. Nonetheless, data-driven methods are still impractical because of the expensive acquisition cost of several SI datasets.

Deep image prior (\textit{DIP}) is a non-data-driven DL approach that overcame the data dependency limitation, showing that a single generator network is sufficient to capture the low-level SI statistics~\cite{ulyanov2018deep,bacca2021compressive, zhang2019hyperspectral}. 
\dcheck{In particular, DIP approaches usually entail autoencoder-based networks that learn the underlying image's coding and decoding process, providing promising results for remote sensing tasks~\cite{sidorov2019deep}. Still, the network structure is a black-box that lacks interpretability.}

\dcheck{Further,~\cite{rasti2021undip,miao2021hyperspectral, gelvez2021interpretable, rasti2022misicnet} proposed network architectures including physical factors as the abundance and the endmembers with a linear mixture model, mitigating the lack of interpretability. However, the non-linear interactions and the benefits of using an interpretable network to solve different tasks have not been considered.} Therefore, this paper proposes a non-data-driven SI recovery method that, contrary to previous methods, considers the non-linear interactions and adds interpretability to the network using linear and non-linear mixture models.

The proposed SI recovery method contains three main components, the input structure, the network architecture, and the customized loss function. The input structure uses a low-rank decomposition of the network input based on the SI low-dimensionality prior. The network architecture, termed Mixture-Net, consists of a sequence of interpretable deep-blocks composed of three trainable block-layers representing the abundance, the endmember, and the non-linear model learning. The customized loss function consists of each deep-block loss sum with custom regularizers considering spatial-spectral correlations. \dcheck{Remark that, Mixture-Net is interpretable in that the learned features and weights can be interpreted as the abundances and endmembers, supporting remote sensing tasks such as unmixing and material identification without running complementary routines.} Simulations over six datasets and three SI recovery tasks demonstrate the effectiveness of the proposed interpretable Mixture-Net architecture.

The contributions of the paper are summarized as follows, 
\begin{itemize}
\item \dcheck{Mixture-Net: An interpretable network architecture inspired by linear and non-linear low-rank mixture models, where the learned weights and features can be interpreted as the SI abundances and endmembers (Section~\ref{sec4:interpretable}).}
\item A scheme of multiple deep-blocks, whose loss functions contain particular model-based regularizers to improve the SI learned spatial-spectral correlations (Section~\ref{sec4:multipleloss}). 
\item A non-data-driven DIP-based approach for SI denoising, HSI-SR, and CSI reconstruction (Section~\ref{sec4:proposal}).
\item A significant improvement for SI recovery in terms of recovery quality while reducing the processing complexity through the non-data-driven approach (Section~\ref{sec5:simulaciones}).
\end{itemize}


\section{Mathematical Spectral Imaging Background}
\label{sub:mixture_models}
Let $\mathbf{f} \in \mathbb{R}^{n}, n = N \times N \times L$, be the vector form of a SI with $N \times N$ spatial pixels and $L$ spectral bands. A SI with hundred of bands is referred to as a hyperspectral image (HSI)~\cite{gelvez2020nonlocal}. \dcheck{A SI usually contains a few number $r \ll L$ of different materials uniquely represented by the spectral response. Then, the $i^{th}$ spatial pixel $\mathbf{f}_{i}\in \mathbb{R}^{L}$ can be modeled as the linear combination of the form $\mathbf{f}_{i} = \mathbf{E}\mathbf{a}_{i}$, known as the linear mixture model (LMM). $\mathbf{E} \in \mathbb{R}^{L \times r}$ denotes an endmember matrix, whose columns contain a unique spectral response, and $\mathbf{a}_i \in \mathbb{R}^{r}$ denotes an abundance vector, whose elements contain the fractional proportions of each endmember at the $i^{th}$ spatial pixel. Consequently, $\mathbf{f}$ can be low-rank represented as
\begin{equation}
\mathbf{f} = (\mathbf{E} \otimes \mathbf{I}_{N^2}) \bar{\mathbf{a}} = \bar{\mathbf{E}} \bar{\mathbf{a}},
\label{eq:LinearModel}
\end{equation}
where $\bar{\mathbf{a}} \in \mathbb{R}_{+}^{N^2r} = [ \mathbf{a}_1^T \hspace{1mm} \ldots \hspace{1mm} \mathbf{a}_{i}^{T} \hspace{1mm} \ldots \hspace{1mm}
\mathbf{a}_{N^2}^{T}]^T$ stacks the abundances; $\bar{\mathbf{E}} \in \mathbb{R}_{+}^{N^2L \times N^2r}$ encompasses the endmembers spanning the SI; $\mathbf{I}_{N^2}$ is an identity matrix of size $N^2 \times N^2$; and $\otimes$ denotes the Kronecker product, introduced to apply the endmembers along the spatial pixels in vector notation.
\dcheck{For details of the low-rank representation refer to~\cite{gelvez2017joint}.}}
Usually, the LMM does not reliably describe the complex spatial-spectral endmember interactions~\cite{bioucas2012hyperspectral}. Therefore, non-linear mixture models (NLMMs) aim to consider non-linear interactions and scattering factors via flexible models described by
\begin{equation}
\mathbf{f} = \mathcal{N}(\mathbf{E}, \bar{\mathbf{a}}), 
\end{equation} 
where $\mathcal{N}$ stands for an implicit function that defines non-linear interactions between the endmembers and the abundances~\cite{wang2019nonlinear}. 

The mixture models impose some physical constraints over components $\mathbf{E}$ and $\bar{\mathbf{a}}$. Precisely, the non-negativity considering the nature of reflected light in spectral signatures, so that the entries of the endmember matrix $\mathbf{E}$ have to be non-negative values, and the sum-to-one constraint considering the fractional proportions of each endmember at each pixel has to be one, 
i.e., $\mathbf{a}_{i}^T \mathbf{1} = 1, \forall i \in N^2$, where $\mathbf{1}$ is a vector with all values in $1$. 

\subsection{Spectral Image Recovery Problems}
The SI acquisition process usually produces degraded and noisy measurements, described mathematically through the general vector forward linear model given by
\begin{equation}
\mathbf{y}=\boldsymbol{\Phi}\mathbf{f} + \boldsymbol{\omega},
\label{eq:sensing_model}
\end{equation}
where $\mathbf{y} \in \mathbb{R}^{m}$ stands for the measurements, $\boldsymbol{\Phi} \in \mathbb{R}^{m \times n}$ stands for the sensing matrix, and $\boldsymbol{\omega}\in \mathbb{R}^{m}$ stands for added noise.
This paper addresses the following problems,

\subsubsection{Denoising} when $m = n$, $\boldsymbol{\Phi} = \mathbf{I}_n \in \mathbb{R}^{n \times n}$ denotes the identity matrix, and $\mathbf{y}$ models a noisy SI. This problem appears by external factors of the optical system and environmental conditions during the acquisition process.

\subsubsection{Single Hyperspectral Super-Resolution} \label{sub:super-resolution} 	when $m = n/d$, $\boldsymbol{\Phi} = \mathbf{DB} \in \mathbb{R}^{m \times n}$, with $\mathbf{B} \in \mathbb{R}^{n \times n}$ being a blurring matrix, and $\mathbf{D} \in \mathbb{R}^{m \times n}$ being a spatial downsampling matrix with downsampling factor $d \in \mathbb{Z}_{++}$, and $\mathbf{y}$ models a spatially degraded SI, referred to as low-resolution HSI (LR-HSI). This problem occurs by the limited incident light affecting the spatial resolution when acquiring several spectral bands.

\subsubsection{Compressive Spectral Imaging Reconstruction} when $m \ll n$, $\boldsymbol{\Phi} = \mathbf{H} \in \mathbb{R}^{m \times n}$ denotes a compressive sensing matrix, and $\mathbf{y}\in \mathbb{R}^{m}$ models the SI compressed measurements. This problem is related to the compressive sensing theory that simultaneously acquire and compress a signal.
	
The SI can be recovered from the noisy, LR-HSI or compressed measurements by solving the inverse problem
\begin{align}
\hat{\mathbf{f}} \in \underset{\mathbf{f} \in \mathbb{R}^{n}}{\mbox{argmin}} & \hspace{2mm} F(\mathbf{f}| \mathbf{y})+\lambda R(\mathbf{f}),
\label{eq:traditional_inverse}
\end{align}
where $F(\cdot):\mathbb{R}^{n}\times \mathbb{R}^m \rightarrow \mathbb{R}$ denotes a data fidelity term, $R(\cdot):\mathbb{R}^{n} \rightarrow \mathbb{R}$ denotes a regularizer that promotes SI prior information, and $\lambda > 0$ denotes the regularization parameter.


\section{Proposed Spectral Image Recovery Method}
\label{sec4:proposal}

\input{tikz/figure1}

The proposed SI recovery method aims to include the prior information implicitly in the architecture of a deep model that generates the SI measurements by minimizing
\begin{align}
\hat{\theta} \in \underset{\theta}{\mbox{argmin}} \hspace{2mm} \mathcal{L}\left(\mathbf{y}, \boldsymbol{\Phi}\mathcal{M}_{\theta}(\mathbf{f}^0)\right),
\label{eq:main_ecuation}
\end{align}
where $\hat{\mathbf{f}} := \mathcal{M}_{\hat{\theta}}(\mathbf{f}^0)$ denotes the recovered SI, $\mathcal{M}_{\theta}(\cdot): \mathbb{R}^{n} \rightarrow \mathbb{R}^{n}$ denotes the deep generative network with $\theta$ adjustable weights, $\mathcal{L}\left(\cdot\right): \mathbb{R}^{m}\times \mathbb{R}^{m} \rightarrow \mathbb{R}$ denotes a customized loss function, and $\mathbf{f}^0 \in \mathbb{R}^{n}$ denotes the model input. Notice that~\eqref{eq:main_ecuation} only requires the measurements $\mathbf{y}$, the acquisition operator $\boldsymbol{\Phi}$, and the input $\mathbf{f}^0$, i.e., the proposed method is non-data-driven.

Figure~\ref{fig:my_proposed_network} schematizes the three components of the proposed SI recovery method. (\romannumeral 1) The input structure is based on tensor decomposition. (\romannumeral 2) The Mixture-Net interpretable architecture generates the recovered SI, the learned features interpreted as the abundances, and the adjusted weights interpreted as the endmembers. (\romannumeral 3) The customized loss function includes the forward model, the losses of each deep-block, and the mixture model physical constraints.
	
\subsection{Input structure component}
\label{sub:input_segment}

The input component determines the input structure for the network $\mathcal{M}_{\theta}(\cdot)$. \dcheck{The proposed method computes the input as}
an adjustable variable from a blind representation by solving
\begin{align}
\{\hat{\theta},\hat{\mathbf{f}^0}\} \in & \hspace{2mm}
\underset{\theta, \mathbf{f}^0 \in \mathbb{R}^{n}}{\mbox{argmin}} \hspace{2mm} \mathcal{L}\left(\mathbf{y}, \boldsymbol{\Phi}\mathcal{M}_{\theta}(\mathbf{f}^0)\right).
\label{eq:main_ecuation_2}
\end{align}
Authors in~\cite{bacca2021compressive} suggested that imposing a low-dimensional input $\mathbf{f}^0$ will force obtaining a low-dimensional output, \dcheck{helpful for capturing the spectral data structure even in the first layer of the model~\cite{huang2022spectral}}. Therefore, the input is learned according to the Tucker decomposition as $\mathbf{f}^0=\mbox{vec}(\boldsymbol{\mathcal{Z}})$, $\boldsymbol{\mathcal{Z}} = \boldsymbol{\mathcal{Z}}_0 \times_1 \mathbf{U} \times_2 \mathbf{V} \times_3 \mathbf{W}$, where the variables $\boldsymbol{\mathcal{Z}}_0 , \mathbf{U}, \mathbf{V}, \mathbf{W}$ are fitted by minimizing~\eqref{eq:main_ecuation_2}. The Tucker decomposition maintains the SI 3D structure and guarantees low-dimensionality given that $\boldsymbol{\mathcal{Z}}_o\in\mathbb{R}^{N_{\rho}\times N_{\rho}\times L_{\rho}}$ stands for a 3D low-rank tensor, where $N_\rho<N$ and $L_\rho<L$, and $\rho$ is a scale factor.

\subsection{Mixture-Net: Network architecture component}
\label{sec4:interpretable}
To implicitly capture the SI prior information, Mixture-Net is composed of a sequence of $K$ interpretable deep-blocks inspired by low-rank mixture models, containing an abundance, an endmember, and a non-linearity block-layer.

\subsubsection{Abundance block-layer}
The first block-layer consists of a CNN that filters the input with the size of the reference SI to obtain an output whose structure and dimensions should match for being interpreted as the abundances. The abundance block-layer of the $k$-th interpretable deep-block can be expressed as
\begin{equation}
\dcheck{\bar{\mathbf{a}}}^{k} = \mathcal{A}_{\theta}^k(\mathbf{f}^{k-1}),
\label{eq:ab_block_layer}
\end{equation}
where $\mathcal{A}_{\theta}^k(\cdot): \mathbb{R}^{N^2L} \rightarrow \mathbb{R}^{N^2r} $ connotes the CNN with $r$ being a tunable hyper-parameter related to the SI rank. The block-layer also includes the non-negativity and sum-to-one physical constraints described in Section~\ref{sub:mixture_models}. Precisely, the \textit{sigmoid} function is used as the activation of the last layer, and the following regularization term is included in the loss function
\begin{equation}
R(\bar{\mathbf{a}}^{k}) = \sum_{i=1}^{N^2}\left(\left(\mathbf{a}^{k}_{i}\right)^T\mathbf{1} - 1\right)^2.
\label{eq:regularization}
\end{equation}

\subsubsection{Endemember block-layer} The second block-layer consists of an operator that performs the matrix multiplication between the learned features in the abundance block-layer $\dcheck{\bar{\mathbf{a}}}^k$ and the adjusted model weights in the endmember block-layer $\mathbf{E}^k$, whose dimensions should match to be interpreted as the endmembers. \dcheck{The endmember block-layer can be expressed as the linear component of the low-rank mixture model as}
\begin{equation}
\mathbf{L}^{k} = \mathcal{E}_{\mathbf{E}^k} \left(\dcheck{\bar{\mathbf{a}}}^{k}\right) = (\mathbf{E}^{k} \otimes \mathbf{I}_{N^2})\dcheck{\bar{\mathbf{a}}}^{k},
\label{eq:Endmember_block}
\end{equation}
where $\mathcal{E}_{\mathbf{E}^k}(\cdot): \mathbb{R}^{N^2r} \rightarrow \mathbb{R}^{N^2L}$ models the fully connected endmember block-layer following the LMM. \dcheck{The adjusted weights $\mathbf{E}^{k}$ are meant to be interpreted as the endmembers, so that their entries must be non-negative~(Section \ref{sub:mixture_models}). An activation function will not guarantee non-negativity since the activations affect the layers' outputs, not the weights. Therefore, a projection to the positive real numbers is imposed over the adjusted weights at each gradient step to guarantee the weights interpretability, i.e., to project $\mathbf{E}^{k}$ into $\mathbb{R}_{+}$.}

\subsubsection{Non-Linearity block-layer} The third block-layer consists of a convolutional operator determining the non-linear transformation applied to the linear component obtained in the endmember block-layer $\mathbf{L}^k$. The block-layer learns the non-linear interactions providing an output whose structure and dimension should match to be interpreted as the recovered SI. \dcheck{The third block-layer can be expressed as the non-linear component of the low-rank mixture model as}
\begin{equation}
\mathbf{NL}^{k} = \mathcal{N}_{\theta^{k}}(\mathcal{E}_{\mathbf{E}^k} \left(\dcheck{\bar{\mathbf{a}}}^{k}\right)),
\label{eq:non-linear-mapping}
\end{equation}
where $\mathcal{N}_{\theta}^{k}(\cdot):\mathbb{R}^{N^2r} \rightarrow \mathbb{R}^{N^2L}$ connotes a CNN that determines the non-linear transformation following the NLMM.

In summarizing, the $k^{th}$ interpretable deep-block estimates the recovered SI, generating three outputs given by
\begin{equation}
\begin{split}
\mathcal{M}_{\theta^k} (\mathbf{f}^{k-1}) = \hspace{3mm} \mathbf{f}^{k} & = \hspace{1mm} (1-\lambda)\mathbf{L}^{k} + \lambda \mathbf{NL}^{k}, \label{eq:eq7}
\\ \mathbf{L}^{k} & = \hspace{1mm} \mathcal{E}_{\mathbf{E}^{k}}(\mathcal{A}_{\theta}^{k}(\mathbf{f}^{k-1})),
\\ \mathbf{NL}^{k} & = \hspace{1mm} \mathcal{N}_{\theta^k}(\mathcal{E}_{\mathbf{E}^{k}}(\mathcal{A}_{\theta}^{k}(\mathbf{f}^{k-1})),
\end{split}
\end{equation}
for $k = 1, \ldots, K$ interpretable deep-blocks.
\dcheck{Note that each $\mathbf{f}^{k}$ follows a low-rank representation since it results from an affine combination between $\mathbf{L}^{k}$ in~\eqref{eq:Endmember_block} and $\mathbf{NL}^{k}$ in~\eqref{eq:non-linear-mapping}, interpreted as the linear and non-linear components of the low-rank mixture model balanced by the parameter $0 \leq \lambda \leq 1$.}

\subsection{Customized loss function component}
\label{sec4:multipleloss}
The loss function is decisive for the learning effectiveness in each SI recovery problem. Therefore, the proposed method admits a customized loss function considering the forward model, the independent deep-blocks, and the physical constraints. First, the forward operator $\boldsymbol{\Phi}$ is applied at each deep-block output to predict the measurements. Subsequently, the weights $\boldsymbol{\theta}$ are adjusted by minimizing a loss that \dcheck{measures the predicted measurements fidelity to the observed measurements through a chosen data-fidelity term, $F(\mathbf{f}^{k}| \mathbf{y}, \boldsymbol{\Phi})$}, and contains the mixture model physical constraint\dcheck{, $R(\mathcal{A}_{\theta}^k(\mathbf{f}^{k})) = \sum_{i=1}^{N^2}\left(\left(\mathbf{a}^{k}_{i}\right)^T\mathbf{1} - 1\right)^2$, where the sum of the fractional proportions at each spatial location must sum one}. \dcheck{Lastly, the loss function is composed of the sum of the single losses at each deep-block as follows}
\begin{equation}
\begin{split}
& \{\boldsymbol{\theta}^*\} \in \argmin_{ \boldsymbol{\theta}} \sum_{k} \tau_k \hspace{0.1em}\mathcal{L}^{k}\left( \mathbf{f}^{k}| \hspace{1mm} \mathbf{y},\boldsymbol{\Phi} \right), \label{eq:regularization1} \\ 
& \mathcal{L}^{k}(\mathbf{f}^{k} | \hspace{1mm} \mathbf{y},\boldsymbol{\Phi}) = F(\mathbf{f}^{k}| \mathbf{y}, \dcheck{\boldsymbol{\Phi}}) + \gamma^{k}R(\mathcal{A}_{\theta}^k(\mathbf{f}^{k})),
\end{split}
\end{equation}
where, $\mathbf{f}^{k} := \mathcal{M}_{\theta^{k}}(\mathbf{f}^{k-1})$, $\mathcal{M}_{\theta^{k}}(\cdot): \mathbb{R}^{N^2 L} \rightarrow \mathbb{R}^{N^2 L}$ stands for the $k^{th}$ interpretable deep-block, $\tau_k>0$ denotes the $k^{th}$ loss function relative weight, and $\gamma_k>0$ denotes the $k^{th}$ regularization parameter that controls the trade-off between the \dcheck{data-fidelity term} $F(\mathbf{f}^{k}| \mathbf{y}, \dcheck{\boldsymbol{\Phi}})$ and the abundance sum-to-one constraint defined in \eqref{eq:regularization}. The output at each deep-block could be interpreted as the recovered SI; therefore, this paper studies two options (i) using the last {deep-block} output, i.e., $\hat{\mathbf{f}} := \mathbf{f}^K$ and (ii) using the average between the last two {deep-block}s outputs, i.e., $\hat{\mathbf{f}}:=({\mathbf{f}}^K+{\mathbf{f}}^{K-1})/2$.

 
\section{Simulations and Results}
\label{sec5:simulaciones}
The experiments to evaluate the Mixture-Net performance were conducted over six publicly available spectral datasets. 

The Pavia University dataset\footnote{Available in~\href{https://rslab.ut.ac.ir/data}{Remote Sensing Datasets}. Accessed: 09-Sep-2022.}, acquired with the ROSIS sensor spanning the ($0.43-0.86$) $\mu$m spectral range. It contains $103$ spectral bands, and $610 \times 610$ spatial pixels. A sub-region of $400\times 200$ pixels is used following the setup in~\cite{nguyen2020sure}.

The hyperspectral Pavia Center dataset$^{1}$, acquired with the ROSIS sensor. It contains $102$ spectral bands and $1096 \times 1096$ spatial pixels. \dcheck{\textbf{Remark:} Since only one image is adopted, the training process follows the splitting scheme in~\cite{jiang2020learning}.
Precisely, the image's central region (1096 $\times 715$) is cropped and divided into training and testing data. The left sub-region is extracted to form the testing data, with four non-overlapped HSIs with $224\times224$ pixels. The remaining region is extracted to form the training data with overlapped HSIs. Finally, the ten percent of the training data is included as a validation set.}

The multispectral Stuff-Toys and Fake images taken from the CAVE dataset\footnote{Available in~\href{https://www.cs.columbia.edu/CAVE/databases/multispectral/}{Multispectral Image Dataset}. Accessed: 09-Sep-2022.} that contains $32$ images of everyday objects with $512 \times 512$ spatial pixels, and $31$ spectral bands ranging from $400$nm to $700$nm at $10$nm steps~\cite{yasuma2010generalized}. 

The KAIST\footnote{Available in~\href{http://vclab.kaist.ac.kr/siggraphasia2017p1/}{KAIST Dataset}. Accessed: 09-Sep-2022.} and the ARAD datasets\footnote{Available in~\href{https://competitions.codalab.org/competitions/22226}{ARAD dataSet}. Accessed: 09-Sep-2022.} with $512 \times 512$ spatial pixels and $31$ spectral bands~\cite{choi2017high,arad2020ntire}. ARAD contains $480$ images ranging from $400$nm to $700$nm at $10$nm steps. 

The Jasper Ridge dataset$^{1}$ with $512 \times 614$ spatial pixels and $224$ spectral bands, ranging from $380$nm to $2500$nm at $9.46$nm steps. Jasper Ridge contains four land classes: Road, Soil, Water, and Tree,~\cite{zhu2014structured}. A sub-region of $100 \times 100 \times 66$ spatial-spectral dimension aligned at the $ (105,269)^{th}$ location in the original image is used following the setup in~\cite{zhu2014structured}.

Mixture-Net was tested in three SI recovery tasks, SI denoising, HSI-SR, and CSI reconstruction. For HSI-SR and CSI reconstruction, the $\ell_2$-norm was used as the data fidelity term, given by $F(\mathbf{f}^{k}| \mathbf{y}, \dcheck{\boldsymbol{\Phi}}) = \left\| \mathbf{y}- \boldsymbol{\Phi}\mathbf{f}^{k}\right\|^2_2$. For SI denoising, the \dcheck{state-of-the-art SURE loss was chosen to avoid over-fitting for high levels of noise, described in~\cite{nguyen2020sure} as}
\begin{equation}
\begin{split}
F(\mathbf{f}^{k}| \mathbf{y}) = \hspace{2mm}\left\| \mathbf{y}- \boldsymbol{\Phi} \mathbf{f}^{k} \right\|^2_2 - \sigma^2 + \hspace{2mm} \dfrac{2\sigma}{N^2L} \mbox{div}_{\mathbf{y}}( \boldsymbol{\Phi}\mathbf{f}^{k}),
\end{split}
\end{equation}
where $\mbox{div}_{\mathbf{y}}(\boldsymbol{\Phi}\mathbf{f}^{k})$ is the divergence of $\boldsymbol{\Phi}\mathbf{f}^{k} := \boldsymbol{\Phi}\mathcal{M}_{\theta^k}(\mathbf{f}^{0})$ computed with the Monte-Carlo SURE strategy in~\cite{ramani2008monte} as
\begin{equation}
\begin{split}
\mbox{div}_{\mathbf{y}}( \boldsymbol{\Phi}\mathbf{f}^{k})) \approx \mathbf{b}^T \left( \dfrac{\boldsymbol{\Phi}(\mathcal{M}_{\theta^k}(\mathbf{f}^{0}+ \epsilon) ) - \boldsymbol{\Phi}\mathcal{M}_{\theta^k}(\mathbf{f}^{0}) }{\epsilon} \right).
\end{split}
\end{equation}
$\mathbf{b} \in \mathbb{R}^{N^2L}$ is an i.i.d. Gaussian distribution with zero mean and unit variance. \dcheck{The SURE loss entails two hyperparameters $\epsilon$ and $\sigma$. $\epsilon$ is a small value with order $1 \times 10^{-5}$, and $\sigma$ is the noise level calculated by the band-wise mean of the median absolute deviation estimator of the $2$D wavelet transform~\cite{ramani2008monte}.}

\dcheck{The hyper-parameters $\tau_k$ and $\gamma_k$ are tuned for each evaluated scenario following the sweeping methodology described in Supplementary Material, Section I.}

The quality improvement is quantified through the spectral angle mapper (SAM), the root mean squared error (RMSE), the dimensionless global relative error of synthesis (ERGAS), the peak signal-to-noise ratio (PSNR), and the structural similarity (SSIM) metrics calculated as in~\cite{gelvez2020nonlocal, gelvez2017joint}. Simulations were run on an Intel Xeon W-$3223$ with $64$GB of memory, and an NVIDIA RTX $3090$ GPU with $24$GB of memory\footnote{{The source code is publicly available in~\href{https://github.com/TatianaGelvez/Mixture-Net} {Mixture-Net code}}}. 


\subsection{Characterization of Mixture-Net}
This section studies the components and hyper-parameters tuning affecting Mixture-Net. (\romannumeral 1) The input strategy, varying the network input structure. (\romannumeral 2) The abundance block-layer scheme, varying the number of layers and filters. (\romannumeral 3) The non-linearity block-layer, varying the rank. (\romannumeral 4) \dcheck{The number of deep-blocks $K$ with single and multiple losses.} The SI-SR task is carried out over Pavia Center to visualize the characterization results.

\subsubsection{Input structure}
\dcheck{This experiment evaluates the influence of learning the structure of the input $\mathbf{f}^0\in \mathbb{R}^{n}$ against using a fixed input, i.e., the entries do not change during the learning process. Precisely, four fixed input strategies imposing a different structure were considered: A \textit{Constant} input, using a tensor with all values equal to $0.5$, i.e., $\boldsymbol{\mathcal{Z}} \in \{0.5\}^{N \times N \times L} $; A \textit{Random} input, generating a random tensor with a normal Gaussian distribution; A \textit{Mesh-grid} input, initializing the tensor as in~\cite{ulyanov2018deep}. And, an \textit{Estimated} input, roughly estimating the image as $vec(\boldsymbol{\mathcal{Z}}) = \boldsymbol{\Phi}^T\mathbf{y}$. The proposed \emph{Learned} strategy learns a low-dimensional Tucker Decomposition from random noise using $\rho=0.4$ as described in Section~\ref{sub:input_segment}.}

\input{tikz/figure2}

The input $\mathbf{f}^0$ is perturbed at each iteration of the learning process to emulate external noise, improving the results quality as shown in~\cite{ulyanov2018deep}. The perturbation is given by 
\begin{equation}
\mathbf{f}^{(0)} = \text{vec}(\boldsymbol{\mathcal{Z}}) + \beta \boldsymbol{\eta}, 
\end{equation}
where vec($\cdot$) denotes a vectorization operator, $\beta \in \mathbb{R}_{+}$ is a parameter controlling the perturbation level, and $\boldsymbol{\eta}_{i,j} \sim \mathcal{N}(0,1)$ stands for Gaussian additive noise. 
 
Figure~\ref{fig:input-characterization} shows the Mixture-Net quality results varying the input structures. The \emph{Constant} and \emph{Random} inputs are far away from being good inputs, where even the \emph{Mesh-grid} or \emph{Estimated} strategies obtain a better quality. The proposed \emph{Learned} input emerges as the best strategy, with a variance no greater than $0.47$ [dBs] of PSNR and $0.27$ degrees of SAM.

\subsubsection{Abundance-block-layer}
\label{section:Abundance_block}
The abundance block-layer tuning evaluates three architectures to learn the spatial features leading to the estimated abundance maps. The \emph{Convolutional} refers to a sequence of $2$D convolutional layers with padding as in~\cite{goodfellow2016deep}. The \emph{Auto-encoder} refers to a sequence of $2$D convolutional layers, where the first half increase at the double at each feature, i.e., $[\ell, \hspace{1mm} 2\ell, \hspace{1mm} \ldots, \hspace{1mm} 2\upsilon\ell]$, and the remaining half decrease $[2\upsilon\ell, \hspace{1mm} (2\upsilon-1)\ell, \hspace{1mm} \ldots, \hspace{1mm}\ell]$ as in~\cite{choi2017high}. The \textit{ResNet} refers to a sequence of residual neural layers, where the first and the last convolutional layers are concatenated as in~\cite{bacca2021compressive}. Figure~\ref{fig:abundance_structure} shows the PSNR and SAM metrics across the architectures, varying the number of layers and the learned features per layer. It can be observed that the \emph{Convolutional} architecture provides the highest quality, showing the most stable behavior with minimal variances when using thirty two features or more.
 
\input{tikz/figure3}

\subsubsection{Non-linearity block-layer varying the rank}
\label{subse:nonlinearityblocklayer}
\dcheck{This experiment evaluates the effect of introducing the non-linearity block-layer and varying the rank. The rank can be addressed as a hyper-parameter related to the number of different materials in the SI. Figure~\ref{fig:rank_behaviour} shows a quantitative comparison between Mixture-Net with just the LMM ($\lambda = 0$ in~\eqref{eq:eq7}) and Mixture-Net when the non-linearity block-layer is included, i.e., $\lambda>0$. A clear improvement in the quality can be observed, especially for small values of the rank. This result indicates that using many endmembers compensates for the non-linear relations in the underlying scene. However, applications such as material identification should be aware of just a few ground truth endmembers that interact in non-linear ways to form the scene instead of using a large number of endmembers that do not match the spectral response of any material~\cite{bioucas2012hyperspectral}.}
 
\input{tikz/figure4}

\subsubsection{\dcheck{Number of interpretable deep-blocks with single and multiple losses}}
\dcheck{This experiment evaluates the influence of concatenating multiple interpretable deep-blocks and of using a single or multiple losses scheme. The single scheme includes the non-linearity block-layer and one loss only at the end of the whole network; meanwhile, the multiple scheme includes a loss at the end of each intermediate deep-block. 

Mixture-Net is evaluated in two scenarios: (\romannumeral 1) varying the number of interpretable deep-blocks and (\romannumeral 2) fixing five deep-blocks, calculating the quality of each intermediate deep-block output $\mathbf{f}^{k}$, interpreted as a recovered image. 

Figure~\ref{fig:my_deepblocks} (\textit{left}) plots the resulting quality in PSNR as a function of the number of interpretable deep-blocks for the single and multiple schemes. An increasing relationship can be observed, where using multiple losses consistently achieves higher performance than using a single loss. Figure~\ref{fig:my_deepblocks} (\textit{right}) plots the quality at the intermediate deep-blocks when fixing a total number of five deep-blocks. The multiple scheme obtains a remarkably higher performance than the single loss scheme from the first intermediate deep-block. 

In both scenarios, the quality obtained with two deep-blocks is comparable to the quality when using three, four, and five deep-blocks. Therefore, the experiments in the rest of the manuscript employ two deep-blocks, a trade-off between the recovered image quality and the method complexity.}

\input{tikz/figure5}


\subsection{Mixture-Net Performance for Spectral Image Recovery}
This section compares the Mixture-Met performance against state-of-the-art methods for denoising, HSI-SR, and CSI.

\subsubsection{Denoising}
\dcheck{A SI can be corrupted by different types of noise such as additive Gaussian, Poisson, and pepper~\cite{rasti2018noise}. This experiment evaluates the case of additive Gaussian noise over Pavia University. Mixture-Net uses
the SURE loss to avoid over-fitting in the learning process.} The comparison methods cover the block-matching and 3D filtering denoiser (BM3D)~\cite{dabov2007image}, the first order spectral roughness penalty for denoising (FORPDN)~\cite{rasti2013hyperspectral}, the hyperspectral restoration (HyRes)~\cite{rasti2017automatic}, the hyperspectral-DIP~\cite{sidorov2019deep}, the SURE-CNN~\cite{choi2017high}, and the unsupervised disentangled spatio-spectral deep priors method (DS2DP)~\cite{miao2021hyperspectral}.

Table~\ref{tab:denoising} summarizes the quantitative results regarding PSNR and SSIM metrics, \dcheck{where the Input column refers to the noisy image quality.} Mixture-Net obtains a significant improvement, especially in the SSIM metric, indicating that the recovered SI intrinsic structures with Mixture-Net are improved over the state-of-the-art methods as observed in Fig.~\ref{fig:denoisingRGB}. Further, the zoomed version visualizes a noise reduction in smooth regions.

\input{Tables/table1}

\input{tikz/figure6}

\subsubsection{Single Hyperspectral Super-Resolution}
This experiment employs the Pavia Center and Stuff-Toys datasets for two downsampling factors, $d = [4, 8]$. The comparison methods cover the bicubic interpolation; three gray$/$RGB image SR methods, EDSR~\cite{lim2017enhanced}, RCAN~\cite{zhang2018image}, and SAN~\cite{dai2019second}; three single HSI-SR methods, 3DCNN~\cite{mei2017hyperspectral}, GDRRN~\cite{li2018single}, and SSPSR~\cite{jiang2020learning}; and the non-data-driven Hyperspectral-DIP method~\cite{sidorov2019deep}. \dcheck{The optimization uses the Adam algorithm with $lr = 1e^{-3}$ and $\gamma^{k} = 0.5$. The rank $r$ and the number of deep-blocks $K$ were set to $r=6$ and $K=3$, and $r=12$ and $K=4$, for Pavia Center and Stuff-Toys, respectively.}

Tables~\ref{tab:sematic_depth} and~\ref{tab:sematic_depth_CAVE} compare the quantitative results for the single HSI-SR over the evaluated methods. Our non-data-driven Mixture-Net outperforms or achieves competitive quality even against the data-driven methods such as the SSPSR for both datasets, both downsampling factors, and all evaluated metrics. The spectral quality measured with the SAM metric results remarkably improved for the Pavia Center dataset. Thus, the intrinsic low-rank prior is more substantial when a higher number of correlated spectral bands are considered. Figure~\ref{fig:visualPavia} shows a visual comparison of the RGB mapping of some super-resolved images for the Pavia Center dataset, where Mixture-Net improves the spatial quality, particularly for high down-sampling factors.

\input{Tables/table2_3}
\input{tikz/figure7}
\input{tikz/figure8}
\input{Tables/table4}
\setcounter{figure}{9}
\input{tikz/figure10}
\setcounter{figure}{8}
\input{tikz/figure9}

\subsubsection{Compressive Spectral Imaging Reconstruction for KAIST and ARAD datasets} This experiment compares the Mixture-Net effectiveness for CSI reconstruction against data-driven and non-data-driven state-of-the-art methods, including the non-data-driven Plug-and-Play (PnP)~\cite{yuan2020plug} and TL-DIP~\cite{bacca2021compressive}, and the data-driven Deep Non-local Unrolling (DNU)~\cite{wang2020dnu}, \dcheck{Autoencoder (AE)~\cite{choi2017high}, and Joint non-linear Representation and Recovery Network (JR2Net)~\cite{monroy2022jr2net}}. \dcheck{The CSI reconstruction is carried out over the KAIST and ARAD datasets, where the DNU and AE methods were trained with the publicly available codes using a training dataset of $27$ and $450$ images for the KAIST and ARAD datasets, respectively.}

\dcheck{Table~\ref{tab:compressive_spectral_reconstruction} summarizes the average reconstruction quality across all testing images for each database with the standard deviation. For all metrics, Mixture-Net outperforms state-of-the-art methods, showing the effectiveness to recover the spatial and spectral details.} Figure~\ref{fig:comparison} visualizes the recovered images with the reconstruction quality measured in PSNR and SSIM. Mixture-Net improves the quality of reconstruction by up to $7$ [dB] and obtains the highest SSIM, outperforming even the methods employing data during the training step. \dcheck{Figure~\ref{fig:SpectralComparison} shows the recovered spectral signatures at a random spatial location. The absolute errors at right confirm that Mixture-Net obtains the best response estimation for both databases.}

\subsection{\dcheck{Remote Sensing Tasks} Experiments}
 
This experiment aims to evaluate the potential application of Mixture-Net to perform \dcheck{remote sensing tasks} such as unmixing and material identification without running complementary routines. For this, Mixture-Net takes advantage of its architecture's interpretability, providing two outputs interpreted as the abundances and endmembers.

The unmixing experiment is carried out using one single shot of the Dual Disperser Coded Aperture Snapshot Spectral Imaging (DD-CASSI)~\cite{gehm2007single} over the Jasper Dataset. \dcheck{The hyperparameters are the same as the experiment in Section~\ref{section:Abundance_block}, with $\lambda = 0$, i.e., considering only the linear component.} \dcheck{Figure~\ref{fig:unmixing} depicts a false RGB representation of Jasper's ground-truth, the Mixture-Net estimation, and the learned features and adjusted weights. The learned features in the abundance block-layer and the adjusted weights in the endmember block-layer converge to a rough estimation of the abundance maps and endmembers, demonstrating that Mixture-Net could be used for the unmixing problem at any additional cost.}

The material identification experiment is carried out using a single shot of the DD-CASSI and the Fake image containing a real and a fake plastic fruit with similar shape and color. The image is spatially sub-sampled to $256 \times 256$ pixels. Figure~\ref{fig:kaistcomp} shows the ground truth and a false RGB mapping of the Mixture-Net estimation with $r=15$. At the right top, the figure shows two obtained features that can be interpreted as abundances. Then, thresholding is applied over the abundances obtaining the binary maps at the right bottom to determine if the materials of both objects are the same. The color checker shows that the obtained abundance can determine that the materials are different, identifying the fake object without running additional routines.

\setcounter{figure}{10}
\input{tikz/figure11}

\subsection{\dcheck{Real-Data Experiment}}
\input{tikz/figure12}
\dcheck{This experiment evaluates the proposed Mixture-Net for the super-resolution task with real-data, acquired in the Optics Laboratory at Universidad Industrial de Santander with a test-bed composed of a SI and an RGB arm.}

\dcheck{The SI arm is composed of an adjustable mechanical slit (Thorlabs VA100C-30 mm, 8-32 Tap) in which the $4$f system focuses the scene. The entrance slit has a height of $13.59$mm and a width of $20\mu$m. Then, a relay lens at $100$ mm from the slit forms a parallel beam that reaches the $600$ grooves/mm transmission grating that diffracts the light rays onto the sensor. The push-broom imagery spectrogram illuminates the slit with $1032$ wavelengths in the spectral range $420-700$nm with steps of $2.7$nm. The SI acquisition performs $50$ horizontal steps with a resolution of $48\mu$m, so that the acquired low-resolution spectral image, referred to as LR-SI, has $100 \times 50$ spatial pixels and $1032$ spectral bands.} 

\dcheck{The RGB arm is composed of an RGB camera, where the $4$f system focuses the scene onto its detector array, so we employ a sub-region of $1200\times 600$ spatial pixels as a reference RGB image with high spatial resolution, referred to as HR-RGB.} 

\dcheck{Figure~\ref{fig:RealDataCoregistered1} (left) compares the super-resolved image, referred to as HR-SI, against the LR-SI and the HR-RGB. The spatial structure of the HR-SI results comparable to the observed in the HR-RGB, and the HR-SI spectral signatures are estimated accordingly to the LR-SI. In particular, the HR-SI contains smoother spectral signatures in comparison to the LR-SI.}

\dcheck{
Figure~\ref{fig:RealDataCoregistered1} (right) illustrates the Mixture-Net interpretability, where the depicted learned features match the features of a set of abundance maps, and the plotted adjusted weights follow the behavior of a set of endmembers. Therefore, besides super-resolving the image, Mixture-Net provides a set of abundance and endmembers maps that can be used for unmixing or material identification purposes.}

\section{Conclusions}
This paper proposes a non-data-driven spectral image recovery method based on deep image prior, where the low-rank mixture models inspire the network dubbed Mixture-Net. Beyond previous methods, Mixture-Net  includes prior knowledge implicitly in the architecture, addressing the black-box nature of standard deep learning. Mixture-Net is divided into three components, the input, imposing a low-dimensional structure through the learning of a Tucker decomposition; the interpretable Mixture-Net, following a sequence of multiple deep-blocks to estimate the abundances, the endmembers, and the spectral image with non-linear relations; and the custom loss function that considers a regularization  related to physical constraints. The non-data-driven Mixture-Net effectiveness was demonstrated over three spectral image recovery problems: spectral image denoising, single hyperspectral super-resolution, and compressive spectral imaging reconstruction in terms of different metrics, outperforming even data-driven methods requiring the training of a vast amount of data. Along with the experiments, we found that imposing a low-dimensional structure over the input improves the quality of the recovered image. On the other hand, the non-linearity block-layer drastically improved the obtained quality by considering the non-linear relationships between the endmembers. We remark that Mixture-Net can be extended to any other spectral image recovery task, even using different loss functions, where the interpretable advantage of Mixture-Net allows to execute further \dcheck{remote sensing tasks} as linear unmixing and material identification without using additional routines.

\ifCLASSOPTIONcaptionsoff
\newpage
\fi

\bibliographystyle{IEEEtran}
\bibliography{report.bib}

\begin{IEEEbiography}[{\vspace{-0.15in}\includegraphics[width=1in,height=1.45in,clip,keepaspectratio]{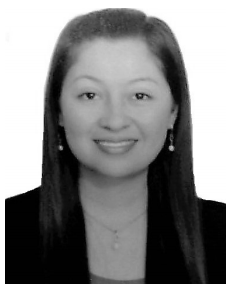}}]{Tatiana Gelvez} (S'17) received the B.S. degree in industrial engineering and systems engineering from the Universidad Industrial de Santander, Bucaramanga, Colombia, in 2016, where she is currently pursuing the Ph.D. degree with the Department of Electrical Engineering. During the second semester of 2019 and 2020 she was an intern at Tampere University, Tampere, Finland. Her research interests include numerical optimization, high-dimensional signal processing, spectral imaging, and compressive sensing.
\end{IEEEbiography}
\vspace{-3.0em}
\begin{IEEEbiography}[{\vspace{-0.15in}\includegraphics[width=1in,height=1.45in,clip,keepaspectratio]{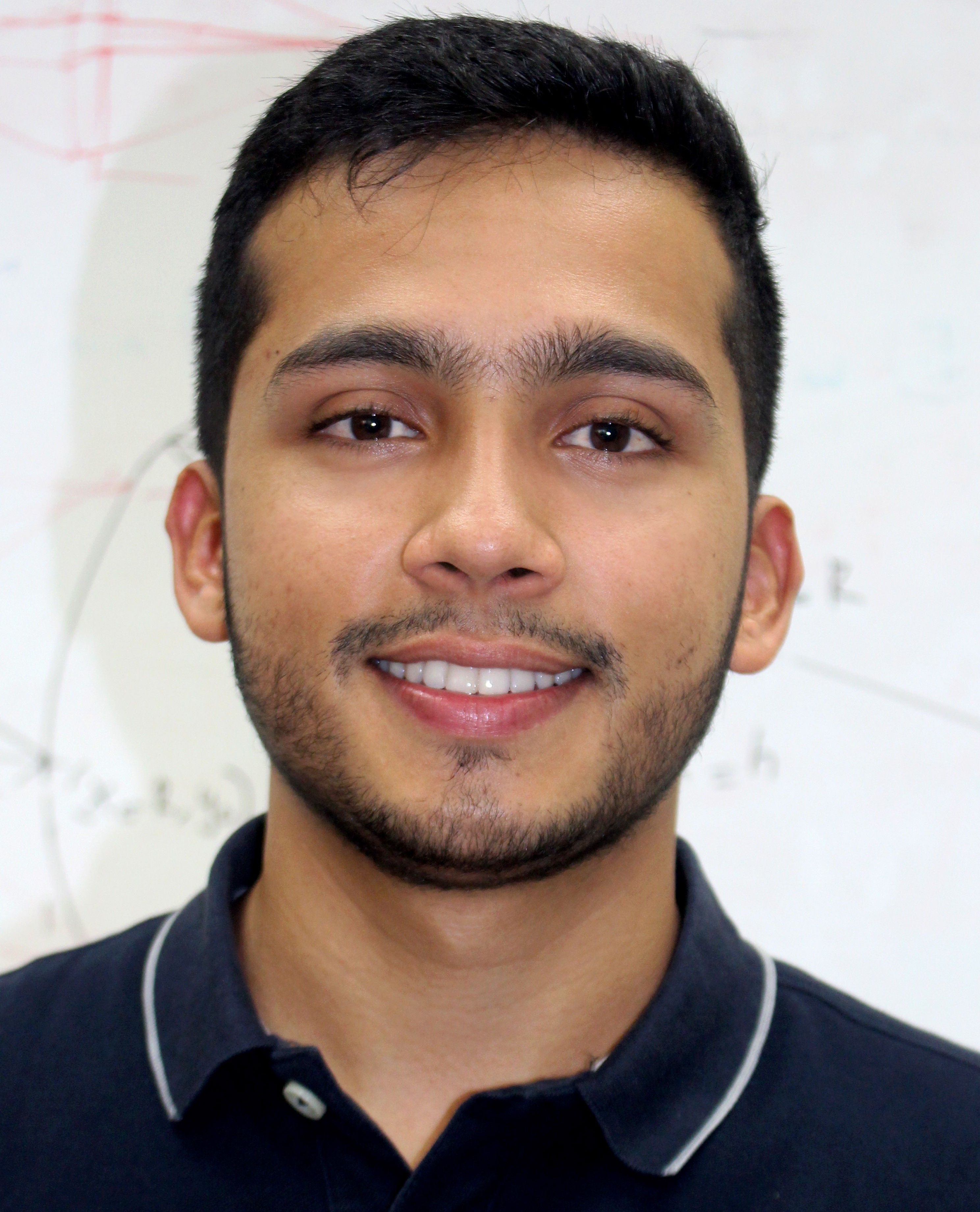}}]{Jorge Bacca}(S'17-M'17) received the B.S. degree in
computer science and the Ph.D. degree in Computer Science from the Universidad Industrial de Santander, Bucaramanga, Colombia in 2017 and 2021, respectively. His current research interests include inverse problem, deep learning methods, optical imagining, compressive sensing, phase retrieval, hyperspectral imaging and numerical optimization.
\end{IEEEbiography}
\vspace{-3.0em}
\begin{IEEEbiography}[{\vspace{-0.05in}\includegraphics[width=1in,height=1.45in,clip,keepaspectratio]{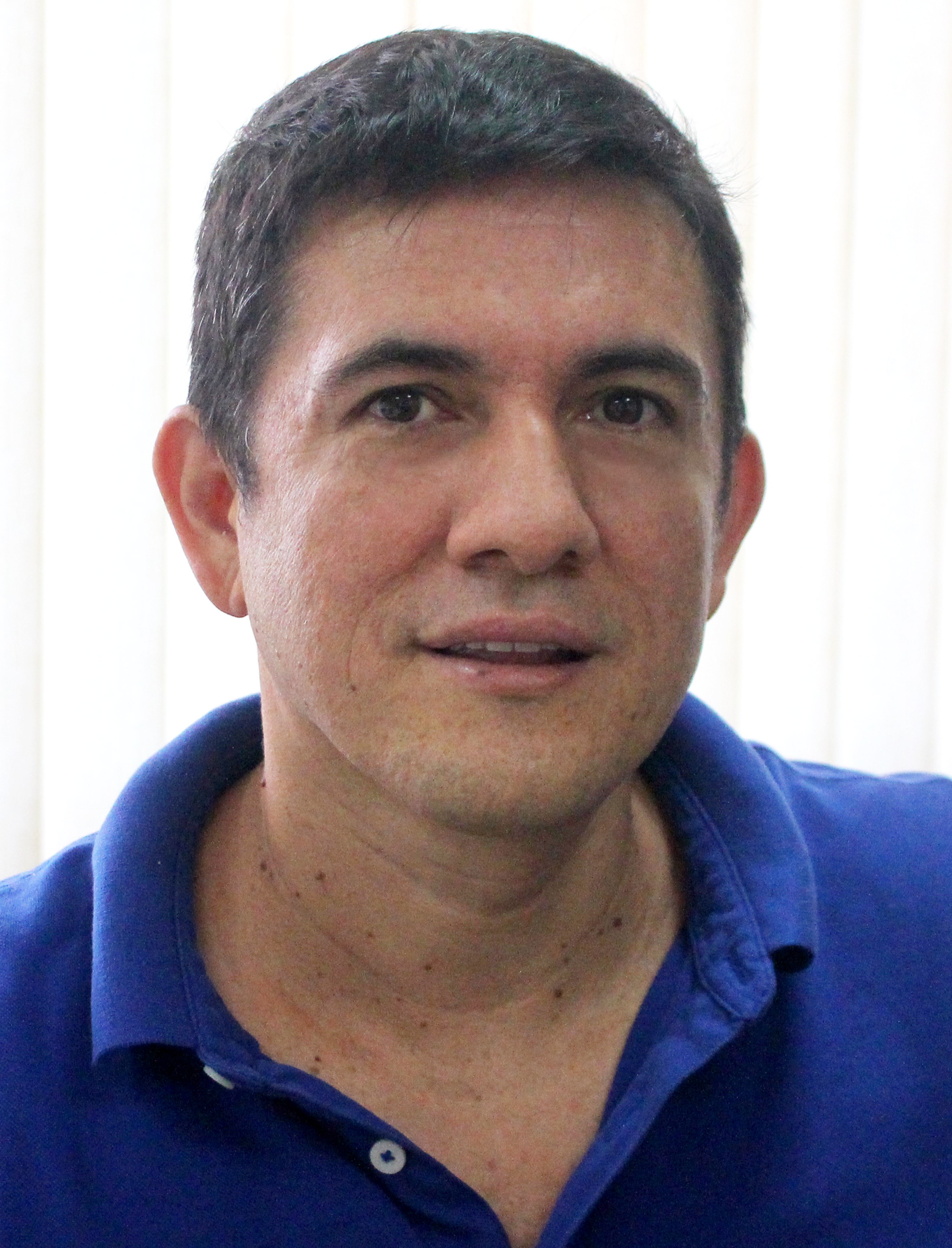}}]{Henry Arguello}(S'11–M'13–SM'17) received the B.Sc. Eng. degree in electrical engineering and the M.Sc. degree in electrical power from the Universidad Industrial de Santander, Bucaramanga, Colombia, in 2000 and 2003, respectively, and the Ph.D. degree in electrical engineering from the University of Delaware, Newark, DE, USA, in 2013. He is currently an Associate Professor with the Department of Systems Engineering, Universidad Industrial de Santander. In first semester 2020, he was a Visiting Professor with Stanford University, Stanford, CA, USA, funded by Fulbright. His research interests include high-dimensional signal processing, optical imaging, compressed sensing, hyperspectral imaging, and computational imaging.
\end{IEEEbiography}
\vfill

\end{document}

%% file: tikz/figure1.tex
\begin{figure*}[hbt!]
\begin{tikzpicture}
\tikzstyle{every node}=[font=\footnotesize]

\filldraw [fill=blue!4, draw=blue!90 , rounded corners, dashed, thick] (-0.49\linewidth,0.55) rectangle (0.49\linewidth,-3.0); 

\node[inner sep=0pt] (Abun) at (0\linewidth,0)
{\includegraphics[width=0.98\linewidth]{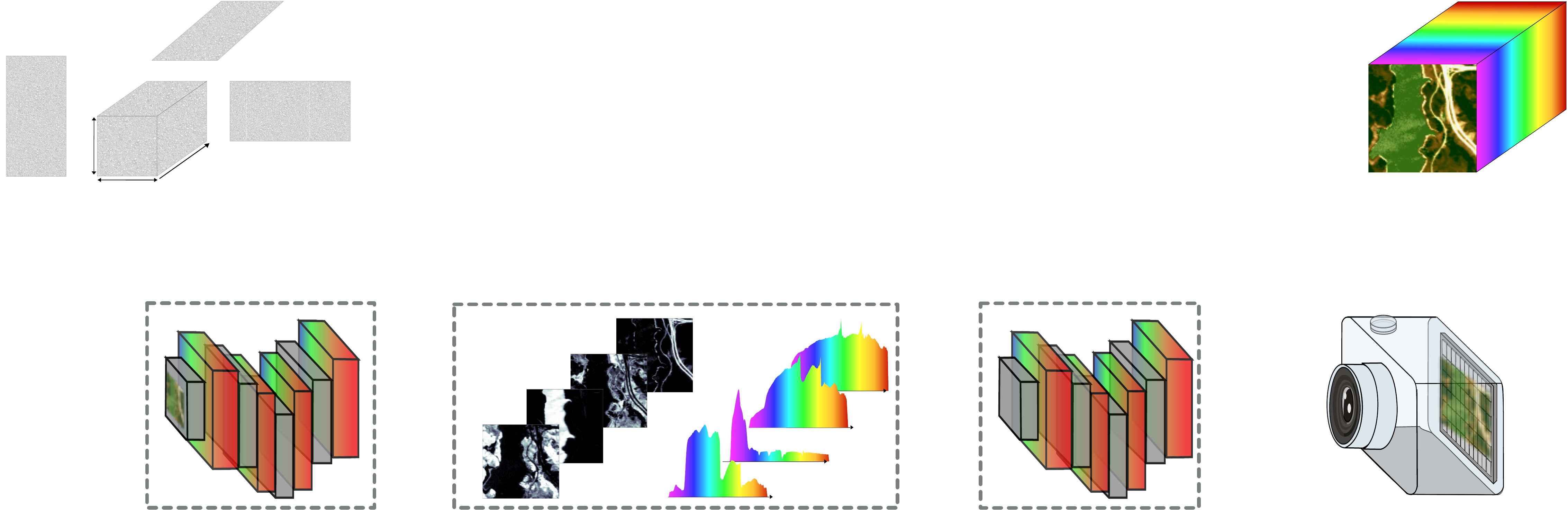}};
 
\draw[black, thick] (-0.22\linewidth,1.3) rectangle (-0.08\linewidth,2.5);
\filldraw[fill=blue!4, draw=blue!90, rounded corners, thick, dashed] (-0.03\linewidth,1.3) rectangle (0.11\linewidth,2.5);
\draw[black, thick] (0.16\linewidth,1.3) rectangle (0.30\linewidth,2.5);
\draw[blue, thick, dashed] (-0.03\linewidth,1.3) -- (-0.49\linewidth,0.55);
\draw[blue, thick, dashed] (0.11\linewidth,1.3) -- (0.49\linewidth,0.55);

\node at (-0.38\linewidth,3.0cm) {(\romannumeral 1) Input};
\node at (-0.0\linewidth,3.0cm) { (\romannumeral 2) Mixture-Net: Network with multiple interpretable deep-blocks $\mathcal{M}_\theta(\cdot)$};
\node at (0.45\linewidth,3.0cm) {Output};
\node at (0.01\linewidth,0.75cm) {Interpretable deep-block};
\node at (-0.33\linewidth,0.35cm) {Abundance block-layer};
\node at (-0.33\linewidth,0.05cm) {$\mathcal{A}_{\theta}^{k}(\mathbf{f}^{k-1})$};
\node at (-0.06\linewidth,0.35cm) {Endmember block-layer};
\node at (-0.06\linewidth,0.05cm) {$\mathcal{E}_{\mathbf{E}}(\bar{\mathbf{a}}^{k})$};
\node at (0.19\linewidth,0.35cm) {Non-linearity block-layer};
\node at (0.19\linewidth,0.05cm) {$\mathcal{N}_{\theta^k}(\mathcal{E}_{\mathbf{E}^{k}}(\bar{\mathbf{a}}^{k})$};
\node at (0.41\linewidth,0.35cm) {loss function};
\node at (0.41\linewidth,0.05cm) {$\mathcal{L}^{k}(\mathbf{f}^{k} | \mathbf{y},\boldsymbol{\Phi})$};
\node[anchor=west] at (-0.4825\linewidth,1.5cm) {$\mathbf{U}$};
\node[anchor=west] at (-0.43\linewidth,1.3cm) {$\boldsymbol{\mathcal{Z}}_0$};
\node[anchor=west] at (-0.325\linewidth,1.65cm) {$\mathbf{V}$};
\node[anchor=west] at (-0.37\linewidth,2.5cm) {$\mathbf{W}$};
\node[rotate=90, anchor=west] at (-0.44\linewidth,0.9cm) {${\rho}N$};
\node[anchor=west] at (-0.4275\linewidth,0.7cm) {${\rho}N$};
\node[anchor=west] at (-0.385\linewidth,0.95cm) {${\rho}L$};

\node[rotate=90, anchor=west] at (0.355\linewidth,1.3cm) {$N$};
\node[anchor=west] at (0.385\linewidth,0.8cm) {$N$};
\node[anchor=west] at (0.46\linewidth,1.15cm) {$L$};

\node at (-0.15\linewidth,1.90cm) {$\mathcal{M}_{\theta^{1}}(\cdot)$};
\node at (0.04\linewidth,1.90cm) {$\mathcal{M}_{\theta^{k}}(\cdot)$};
\node at (0.23\linewidth,1.90cm) {$\mathcal{M}_{\theta^{K}}(\cdot)$};

\filldraw [fill=white, draw=black] (-0.49\linewidth,-3.1) rectangle (0.49\linewidth,-3.6) node[xshift=-0.5\linewidth, yshift=0.25cm] {(\romannumeral 3) Loss function \; $\mathcal{L}(\mathbf{f} | \hspace{1mm} \mathbf{y},\boldsymbol{\Phi}) = \sum_{k} \tau_k \hspace{0.1em}\mathcal{L}^{k}\left(\mathcal{M}_{\theta^{k}}(\mathbf{f}^{k-1}) | \hspace{1mm} \mathbf{y},\boldsymbol{\Phi} \right),$ \hspace{2mm} $\mathcal{L}^{k}\left(\mathcal{M}_{\theta^{k}}(\mathbf{f}^{k-1}) | \hspace{1mm} \mathbf{y},\boldsymbol{\Phi} \right) = \left\| \mathbf{y}- \boldsymbol{\Phi}\mathbf{f}^{k}\right\|^2_2 + \gamma^{k}\sum_{i=1}^{N^2}\left(\left(\mathbf{a}^{k}_{i}\right)^T\mathbf{1} - 1\right)^2.$};

\node[anchor=west] at (-0.49\linewidth,-1.6cm) {$\mathbf{f}^{k-1}$};
\node[anchor=west] at (0.45\linewidth,-1.2cm) {$\mathbf{f}^{k}$};
\node[anchor=west] at (0.35\linewidth,-0.5cm) {$\boldsymbol{\Phi}$};
\node[anchor=west] at (0.40\linewidth,-2.5cm) {$\mathbf{y}^{k}$};

\draw[gray, ultra thick, ->] (-0.26\linewidth,2.0) -- (-0.23\linewidth,2.0);
\draw[red, ultra thick, <-] (-0.26\linewidth,1.75) -- (-0.23\linewidth,1.75);
\draw[gray, ultra thick,->] (-0.07\linewidth,2.0) -- (-0.04\linewidth,2.0);
\draw[red, ultra thick,<-] (-0.07\linewidth,1.75) -- (-0.04\linewidth,1.75);
\draw[black, thick, dotted] (-0.07\linewidth,2.25) -- (-0.04\linewidth,2.25);
\draw[gray, ultra thick,->] (0.12\linewidth,2.0) -- (0.15\linewidth,2.0);
\draw[red, ultra thick,<-] (0.12\linewidth,1.75) -- (0.15\linewidth,1.75);
\draw[black, thick,dotted] (0.12\linewidth,2.25) -- (0.15\linewidth,2.25);
\draw[gray, ultra thick,->] (0.31\linewidth,2.0) -- (0.34\linewidth,2.0);
\draw[red, ultra thick,<-] (0.31\linewidth,1.75) -- (0.34\linewidth,1.75);
\draw[gray, ultra thick, ->] (-0.44\linewidth,-1.50) -- (-0.41\linewidth,-1.50);
\draw[red, ultra thick, <-] (-0.44\linewidth,-1.65) -- (-0.41\linewidth,-1.65);
\draw[gray, ultra thick, ->] (-0.245\linewidth,-1.50) -- (-0.215\linewidth,-1.50);
\draw[red, ultra thick, <-] (-0.245\linewidth,-1.65) -- (-0.215\linewidth,-1.65);
\draw[gray, ultra thick,->] (0.085\linewidth,-1.50) -- (0.115\linewidth,-1.50);
\draw[red, ultra thick,<-] (0.085\linewidth,-1.65) -- (0.115\linewidth,-1.65);
\draw[gray, ultra thick] (0.10\linewidth,-1.50) -- (0.10\linewidth,-0.4);
\draw[gray, ultra thick] (0.10\linewidth,-0.4) -- (0.29\linewidth,-0.4);
\draw[gray, ultra thick, ->] (0.29\linewidth,-0.4) -- (0.29\linewidth,-1.5);
\draw[gray, ultra thick,->] (0.26\linewidth,-1.50) -- (0.28\linewidth,-1.50);
\node at (0.29\linewidth,-1.7) {$\oplus$};
\node at (0.27\linewidth,-1.8cm) {$\lambda$};
\node[rotate=90] at (0.28\linewidth,-1.0cm) {$1-\lambda$};
\draw[gray, ultra thick,->] (0.305\linewidth,-1.50) -- (0.335\linewidth,-1.50);
\draw[red, ultra thick,<-] (0.305\linewidth,-1.65) -- (0.335\linewidth,-1.65);
\draw[gray, ultra thick,->] (0.455\linewidth,-1.50) -- (0.485\linewidth,-1.50);
\draw[red, ultra thick,<-] (0.455\linewidth,-1.65) -- (0.485\linewidth,-1.65);

\node at (-0.115\linewidth,-0.7cm) {$\mathbf{a}_{r}^{k}$};
\node at (-0.135\linewidth,-1.0cm) {$\cdot$};
\node at (-0.155\linewidth,-1.3cm) {$\cdot$};
\node at (-0.175\linewidth,-1.6cm) {$\cdot$};
\node at (-0.195\linewidth,-1.9cm) {$\mathbf{a}_{1}^{k}$};
\node at (0.01\linewidth,-0.7cm) {$\mathbf{e}_{r}^{k}$};
\node at (-0.01\linewidth,-1.0cm) {$\cdot$};
\node at (-0.03\linewidth,-1.3cm) {$\cdot$};
\node at (-0.05\linewidth,-1.6cm) {$\cdot$};
\node at (-0.07\linewidth,-1.9cm) {$\mathbf{e}_{1}^{k}$};

\end{tikzpicture}
\caption{Proposed SI recovery method scheme. (\romannumeral 1) The input $\mathbf{f}_{0}$ is learned as a tensor decomposition to impose a low-rank structure. (\romannumeral 2) Mixture-Net is composed by a sequence of multiple interpretable deep-blocks $\mathcal{M}_{\theta^{k}}(\cdot)$. The $k^{th}$ interpretable deep-block contains an abundance block-layer $\mathcal{A}_{\theta}^{k}(\mathbf{f}^{k-1})$, consisting of a CNN to learn the spatial correlations, an endmember block-layer $\mathcal{E}_{\mathbf{E}}(\bar{\mathbf{a}}^{k})$ performing the matrix product according to the LMM, and a non-linearity block-layer $\mathcal{N}_{\theta^k}(\mathcal{E}_{\mathbf{E}^{k}}(\bar{\mathbf{a}}^{k})$ containing a CNN to learn the NLMM operator. (\romannumeral 3) The customized loss function \dcheck{$\mathcal{L}(\mathbf{f} | \hspace{1mm} \mathbf{y},\boldsymbol{\Phi})$} \dcheck{is formulated as a weighted} sum of the single losses employed at each interpretable deep-block \dcheck{$\mathcal{L}^{k}\left(\mathcal{M}_{\theta^{k}}(\mathbf{f}^{k-1}) | \hspace{1mm} \mathbf{y},\boldsymbol{\Phi} \right)$}, \dcheck{whose first term corresponds to the data fidelity term $ \left\| \mathbf{y}- \boldsymbol{\Phi}\mathbf{f}^{k}\right\|^2_2$, and second term relates the abundance sum-to-one constraint.}}
\label{fig:my_proposed_network}
\end{figure*}

%% file: tikz/figure2.tex
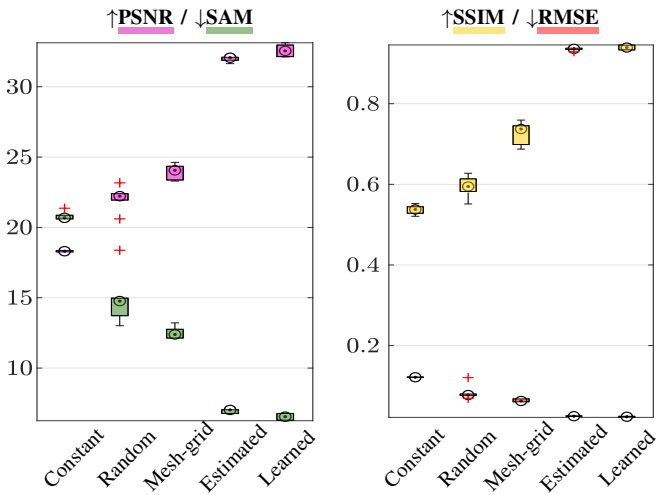
\begin{figure}[b!]
\centering
\begin{tikzpicture}
\tikzstyle{every node}=[font=\footnotesize]
\node[inner sep=0pt] (Abun) at (0\linewidth,0){\includegraphics[height=0.57\linewidth,width=0.46\linewidth]{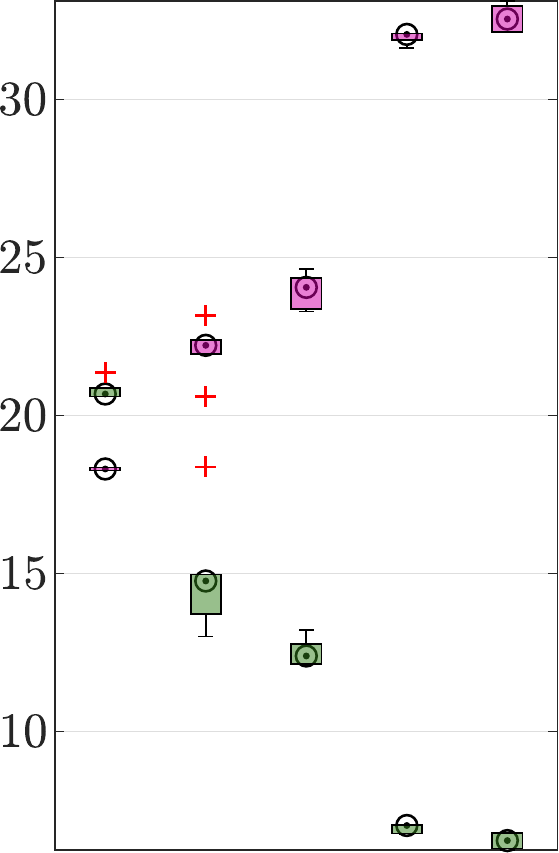}};
\node at (0.03\linewidth,2.8cm) {$\uparrow$\textbf{\textcolor{mypink}{\dunderline{2.5pt}{\textcolor{black}{PSNR}}} / $\downarrow$\textcolor{mygreen}{\dunderline{2.5pt}{\textcolor{black}{SAM}}}}};
\node[rotate = 45] at (-0.13\linewidth,-3.0cm) {Constant};
\node[rotate = 45] at (-0.05\linewidth,-3.0cm) {Random};
\node[rotate = 45] at (0.03\linewidth,-3.0cm) {Mesh-grid};
\node[rotate = 45] at (0.11\linewidth,-3.0cm) {Estimated};
\node[rotate = 45] at (0.19\linewidth,-3.0cm) {Learned};
\end{tikzpicture}
\begin{tikzpicture}
\tikzstyle{every node}=[font=\footnotesize]
\end{tikzpicture}
\begin{tikzpicture}
\tikzstyle{every node}=[font=\footnotesize]
\node[inner sep=0pt] (Abun) at (0\linewidth,0){\includegraphics[height=0.57\linewidth, width=0.46\linewidth]{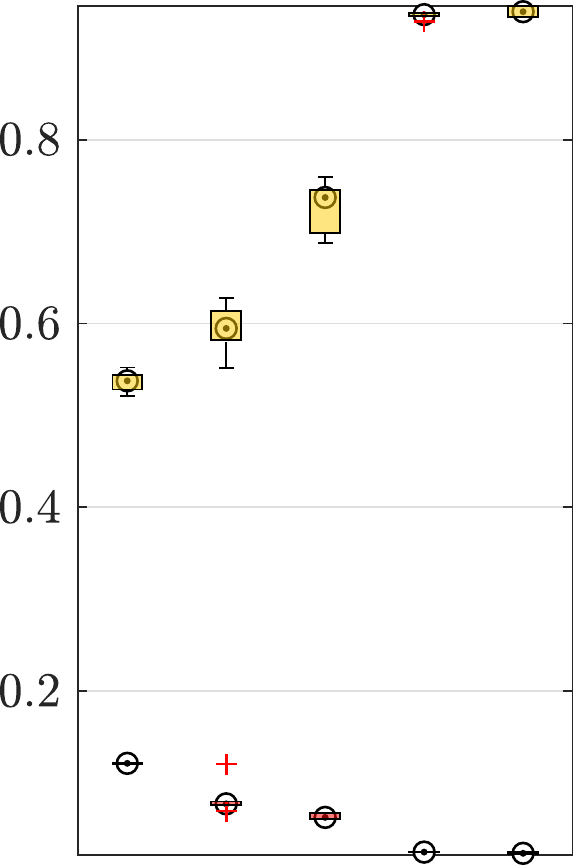}};
\node at (0.03\linewidth,2.8cm){\textbf{$\uparrow$\textcolor{myyellow}{\dunderline{2.5pt}{\textcolor{black}{SSIM}}} / $\downarrow$\textcolor{myorange}{\dunderline{2.5pt}{\textcolor{black}{RMSE}}}}};
\node[rotate = 45] at (-0.13\linewidth,-3.0cm) {Constant};
\node[rotate = 45] at (-0.05\linewidth,-3.0cm) {Random};
\node[rotate = 45] at (0.03\linewidth,-3.0cm) {Mesh-grid};
\node[rotate = 45] at (0.11\linewidth,-3.0cm) {Estimated};
\node[rotate = 45] at (0.19\linewidth,-3.0cm) {Learned};
\end{tikzpicture}
\caption{Quantitative performance box plot measured in \textit{(left)} PSNR (purple) - SAM (green) and \textit{(right)} SSIM (yellow) - RMSE (red) metrics for the five strategies imposing a structure in the input of Mixture-Net. \dcheck{The level of perturbation was varied across the values $[0,\hspace{1mm} 1e^{-2},\hspace{1mm} 3e^{-2},\hspace{1mm} 5e^{-2},\hspace{1mm} 8e^{-2},\hspace{1mm} 1e^{-1}]$. The Abundance block-layer is a CNN with six layers using thirty two filters, and the hyper-parameters were set to $lr = 5e^{-3}$, $\lambda = 0$ (only the LMM), and $r=8$.}}
\label{fig:input-characterization}
\end{figure}
 

%% file: tikz/figure3.tex
\begin{figure}[b!]
\centering
\begin{tikzpicture}
\tikzstyle{every node}=[font=\footnotesize]
\node[inner sep=0pt] (Abun) at (0.0\linewidth,0){\includegraphics[width=0.95\linewidth]{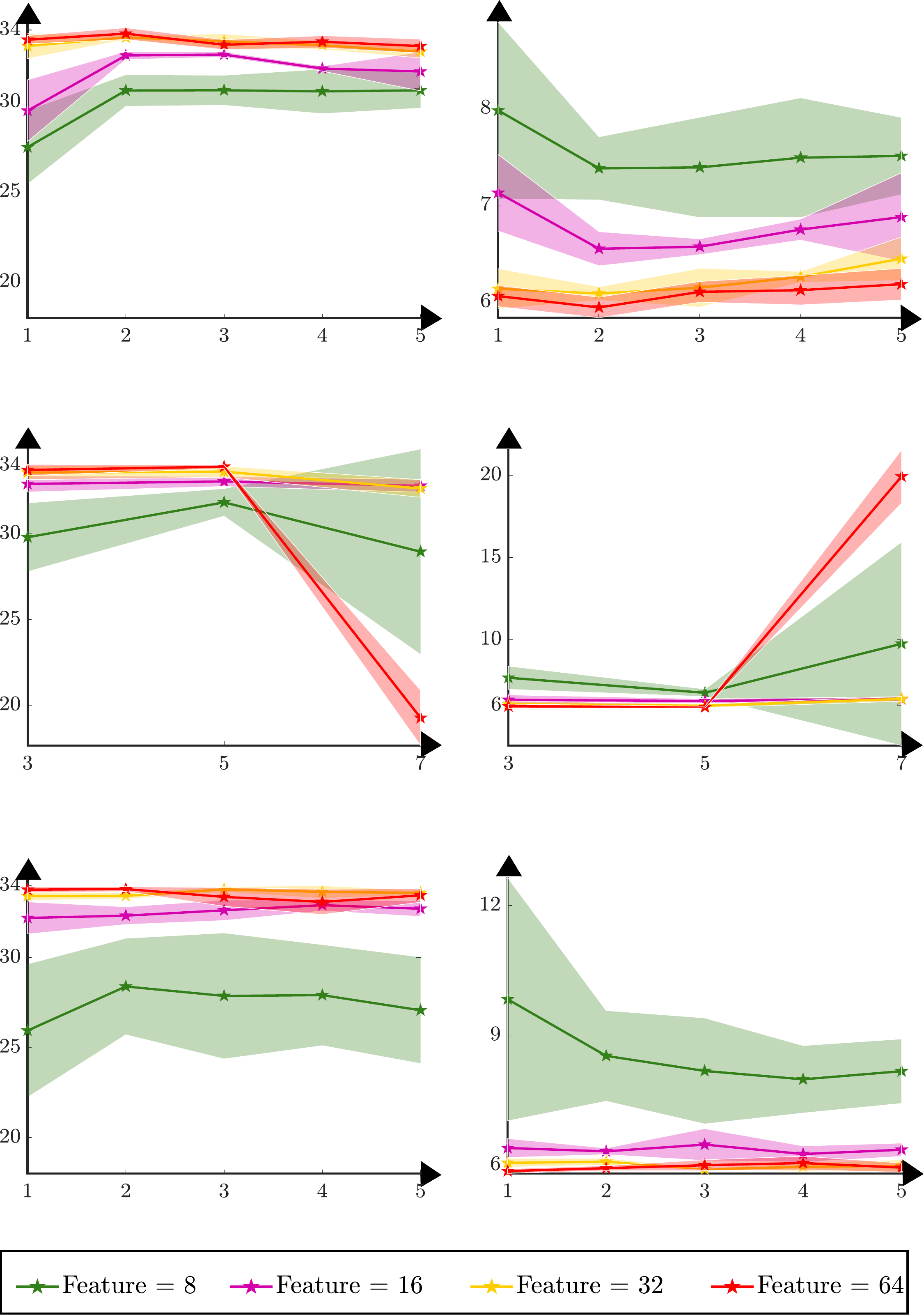}};
\node[rotate=90] at (-0.5\linewidth,4.3cm){\textbf{\textit{Convolutional}}};
\node at (-0.25\linewidth,6.1cm) {$\uparrow$ \textbf{PSNR [dBs]}};
\node at (0.25\linewidth,6.1cm) {$\downarrow$ \textbf{SAM [degrees]}};
\node at (-0.25\linewidth,2.7cm) {Number of layers};
\node at (0.25\linewidth,2.7cm) {Number of layers};
\draw[gray, thick, densely dashdotted] (-0.5\linewidth,2.5) -- (0.5\linewidth,2.5);
\node[rotate=90] at (-0.5\linewidth,0.5cm){\textbf{\textit{Auto-encoder}}};
\node at (-0.25\linewidth,2.2cm) {$\uparrow$ \textbf{PSNR [dBs]}};
\node at (0.25\linewidth,2.2cm) {$\downarrow$ \textbf{SAM [degrees]}};
\node at (-0.25\linewidth,-1.2cm) {Number of layers};
\node at (0.25\linewidth,-1.2cm) {Number of layers};
\draw[gray, thick, densely dashdotted] (-0.5\linewidth,-1.4) -- (0.5\linewidth,-1.4);
\node[rotate=90] at (-0.5\linewidth,-3.3cm){\textbf{\textit{Res-Net}}};
\node at (-0.25\linewidth,-1.7cm) {$\uparrow$ \textbf{PSNR [dBs]}};
\node at (0.25\linewidth,-1.7cm) {$\downarrow$ \textbf{SAM [degrees]}};
\node at (-0.25\linewidth,-5.1cm) {Number of layers};
\node at (0.25\linewidth,-5.1cm) {Number of layers};
\end{tikzpicture}
\caption{Quantitative performance in terms of PSNR \textit{(left)} and SAM \textit{(right)} across three architectures. \dcheck{The hyper-parameters are varied as follows: the number of layers varies in the range $[1-5]$ for the \emph{Convolutional} and \emph{Res-Net} architectures, and in the range $[3, 5, 7]$ for the \emph{Auto-encoder}; and the number of features per layer varies in the range $[8, 16, 32, 64]$. The best learning-rate was found to be $lr = 1e^{-3}$, and the noise perturbation of the input was set to be $\beta = 0.7$.}}
\label{fig:abundance_structure}
\end{figure}
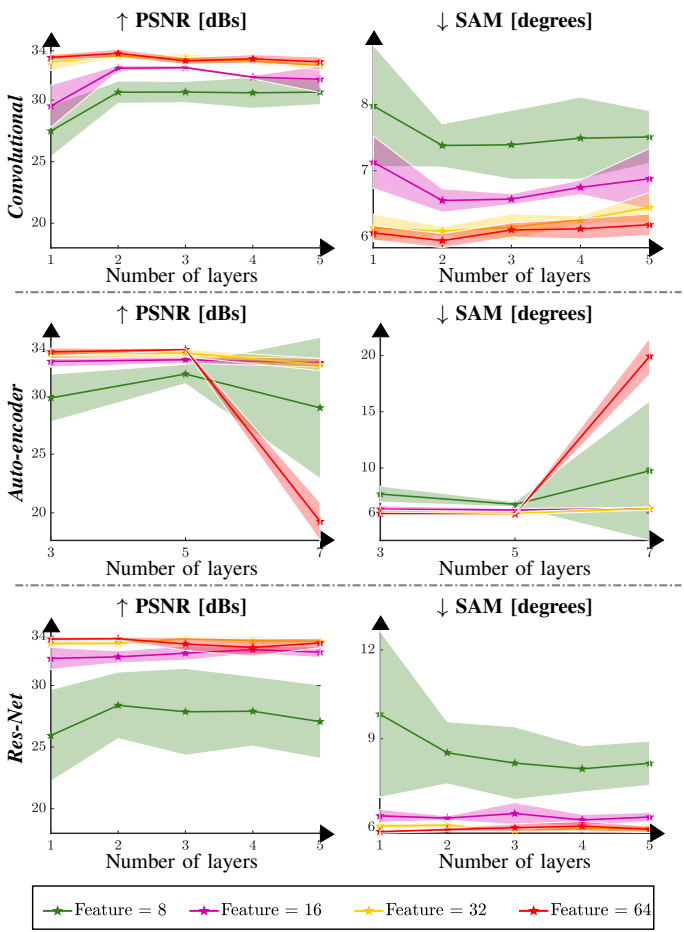
 

%% file: tikz/figure4.tex
\begin{figure}[b!]
\centering
\begin{tikzpicture}
\tikzstyle{every node}=[font=\footnotesize]
\node[inner sep=0pt] (Abun) at (0\linewidth,0){\includegraphics[width=1\linewidth]{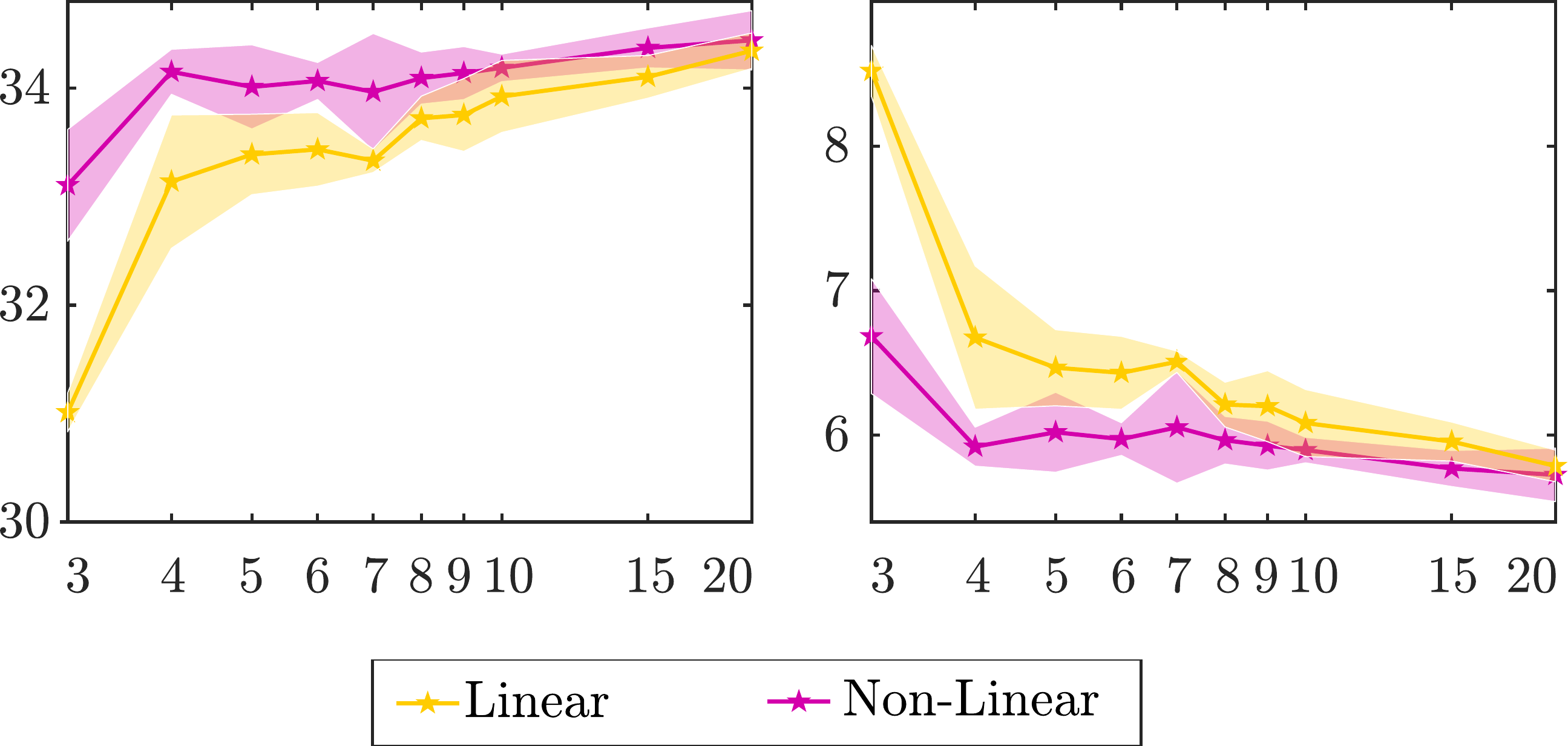}};
\node at (-0.25\linewidth,2.3cm){$\uparrow$\textbf{PSNR [dBs] }};
\node at (0.25\linewidth,2.3cm){$\downarrow$\textbf{SAM [Degrees] }};
\node at (-0.25\linewidth,-1.4cm) {Rank };
\node at (0.25\linewidth,-1.4cm) {Rank};
\end{tikzpicture}
\caption{\dcheck{Quantitative performance in terms of PSNR {(\textit{left})} and SAM {(\textit{right})} when using the linear block-layer, i.e., $\lambda=0$ in~\eqref{eq:eq7}, and when including the non-linearity block-layer with {$\lambda=0.7$. The rank value varies in the range $[3,4,5,6,7,8,9,10, 15, 20]$. The non-linearity block-layer is composed of two consecutive spatial-spectral networks presented in~\cite{wang2019hyperspectral}, with the learning rate set to $lr = 1e^{-3}$.}} 
}
\label{fig:rank_behaviour}
\end{figure}
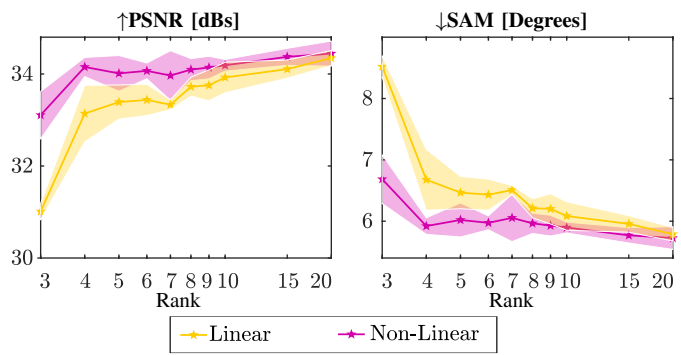

%% file: tikz/figure5.tex
\begin{figure}[ht!]
\centering
\includegraphics[width=0.48\linewidth]{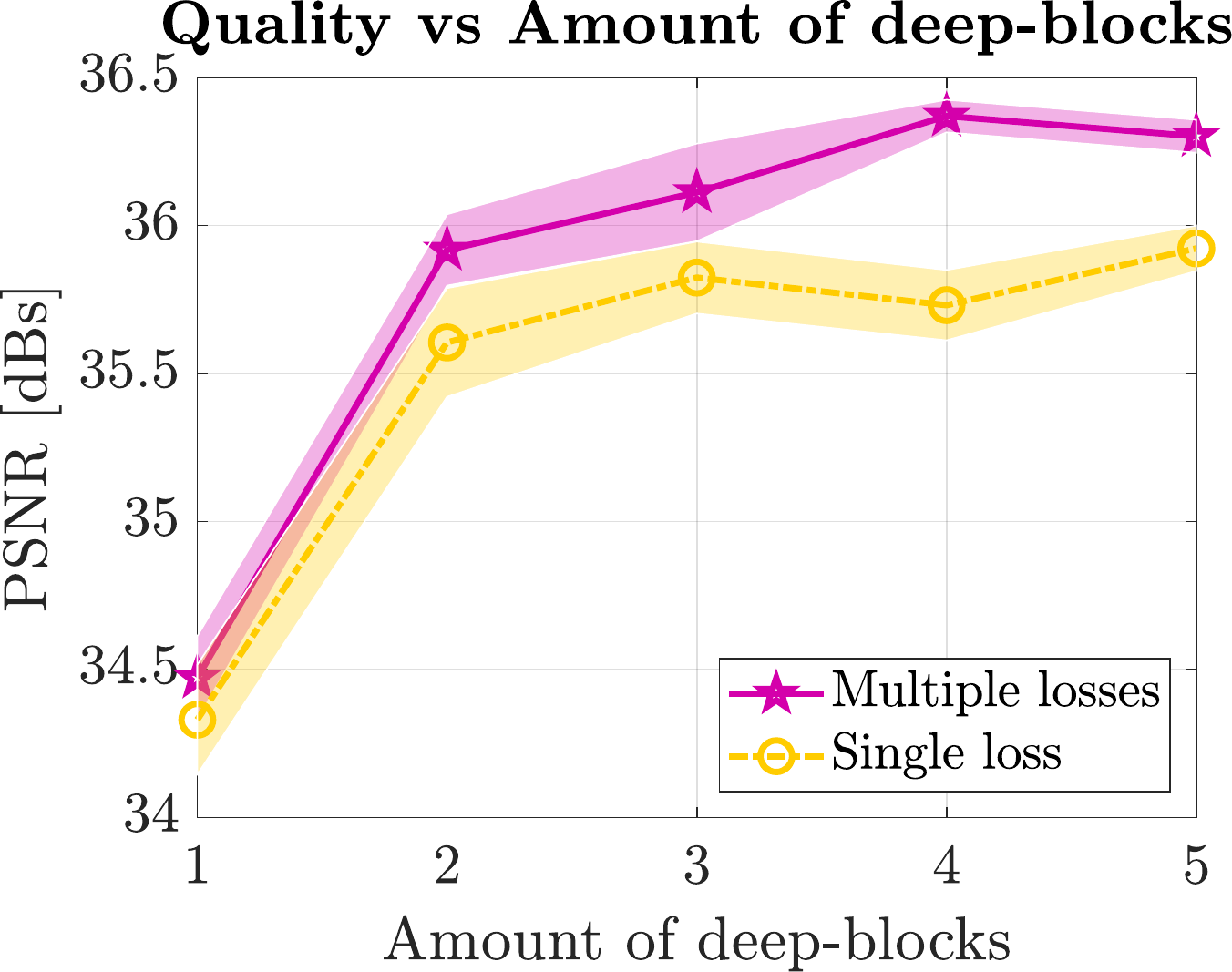}
\includegraphics[width=0.48\linewidth]{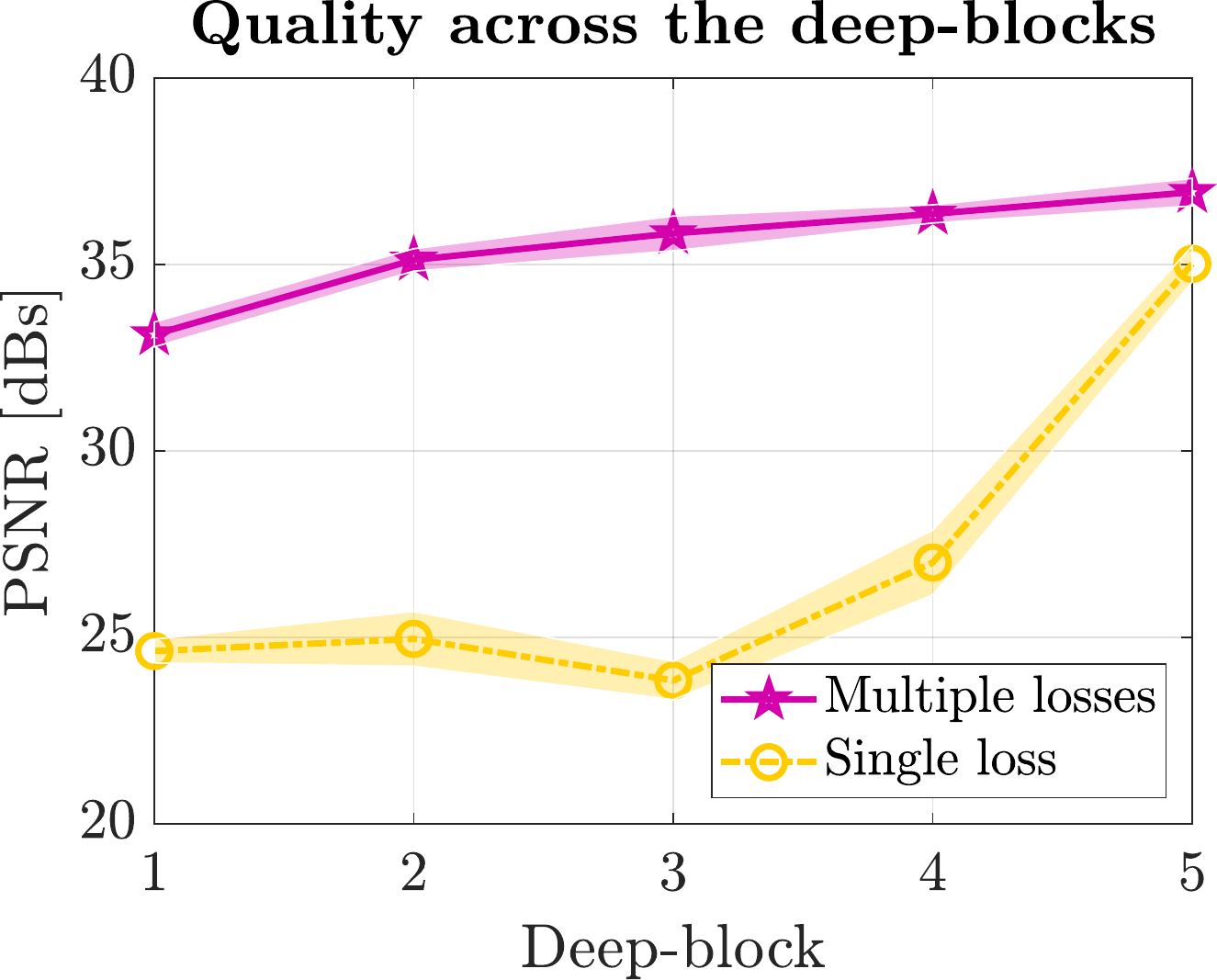}
\caption[justification=justified]{\dcheck{(\textit{left}) Mixture-Net performance, varying the number of interpretable deep-blocks in the range $[1-5]$ for single and multiple losses. (\textit{right}) Recovered image quality at each intermediate deep-block when fixing five deep-blocks.}}
\label{fig:my_deepblocks}
\end{figure}

%% file: Tables/table1.tex
\begin{table}[b!]
\scriptsize
\renewcommand{\arraystretch}{1.0}
\caption{Denoising Quantitative Results for Pavia University}
\label{tab:denoising}
\centering
\resizebox{1\columnwidth}{!}{%
\setlength\tabcolsep{0.05cm}
\begin{tabular}{llllllllll} 
\toprule 
\begin{tabular}[c]{@{}c@{}}Noise\\ $\sigma$\end{tabular} & Metric &\dcheck{Input} & BM3D & FORPDN & HyRes & DIP &\multicolumn{1}{c}{\begin{tabular}[c]{@{}c@{}}SURE\\ CNN\end{tabular}} & DS2DP &\multicolumn{1}{c}{\begin{tabular}[c]{@{}c@{}}Mixture\\-Net\end{tabular}}\\
\bottomrule
\midrule
\multirow{2}{*}{{$\frac{100}{255}$}} & $\uparrow$PSNR &\dcheck{$8.130$} & {$29.14$} & $26.03$ & $28.41$ & $26.47$ &\underline{$29.62$} & $27.54$ & $\mathbf{30.95}$\\ 
&$\downarrow$SSIM &\dcheck{$0.025$} & {$0.754$} & $0.597$ & $0.738$ & $0.683$ &\underline{$0.802$} &$0.718$ & $\mathbf{0.869}$\\ 
\midrule 
\multirow{2}{*}{$\frac{50}{255}$} & $\uparrow$PSNR &\dcheck{$14.15$} & {$32.97$} & $30.44$ & $31.78$ & $30.69$ &\underline{$33.29$} & $32.12$ & $\mathbf{34.47}$\\ 
&$\downarrow$SSIM &\dcheck{$0.102$} & $0.881$ & $0.799$ & $0.855$ & $0.846$ &\underline{$0.905$} & $0.896$ & $\mathbf{0.942}$\\ 
\midrule 
\multirow{2}{*}{$\frac{25}{255}$} & $\uparrow$PSNR &\dcheck{$20.17$} &\underline{$36.48$} & $34.34$ & $35.35$ & $34.48$ & {$36.09$} & $35.55$ & $\mathbf{36.99}$\\ 
&$\downarrow$SSIM &\dcheck{$0.286$} & $0.942$ & $0.906$ & $0.927$ & $0.917$ & $0.945$ &\underline{$0.951$}& $\mathbf{0.965}$\\ 
\bottomrule
\end{tabular} }
\end{table}

%% file: tikz/figure6.tex
\begin{figure}[th!]
\begin{tikzpicture}
\tikzstyle{every node}=[font=\footnotesize]
\begin{scope}[node distance = 0mm, inner sep = 0pt, outer sep = 1,spy using outlines={rectangle, red, magnification=3.5, every spy on node/.append style={thick}}]
\node[align = center](img){\includegraphics[width= .155\linewidth]{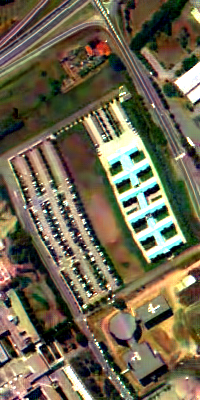}};
\spy [cyan, height=0.08\linewidth, width= .155\linewidth] on (0.05,-0.5) in node at (0,-1.8);
\node at (0,1.6) {BM3D};
\end{scope} 
\filldraw [fill=white, draw=black] (-0.03\linewidth,-0.95) rectangle (0.07\linewidth,-1.35);
\node[anchor=west] at (-0.03\linewidth,-1.15){\footnotesize{$32.97$}};
\end{tikzpicture}
\begin{tikzpicture}
\tikzstyle{every node}=[font=\footnotesize]
\begin{scope}[node distance = 0mm, inner sep = 0pt, outer sep = 1,spy using outlines={rectangle, red, magnification=3.5, every spy on node/.append style={thick}}]
\node[align = center](img){\includegraphics[width= .155\linewidth]{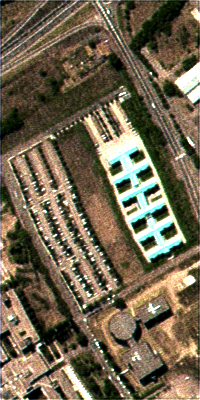}};
\spy [cyan, height=0.08\linewidth, width= .155\linewidth] on (0.05,-0.5) in node at (0,-1.8);
\node at (0,1.6) {HSI-DIP};
\end{scope} 
\filldraw [fill=white, draw=black] (-0.03\linewidth,-0.95) rectangle (0.07\linewidth,-1.35);
\node[anchor=west] at (-0.03\linewidth,-1.15){\footnotesize{$30.69$}};
\end{tikzpicture}
\begin{tikzpicture}
\tikzstyle{every node}=[font=\footnotesize]
\begin{scope}[node distance = 0mm, inner sep = 0pt, outer sep = 1,spy using outlines={rectangle, red, magnification=3.5, every spy on node/.append style={thick}}]
\node[align = center](img){\includegraphics[width= .155\linewidth]{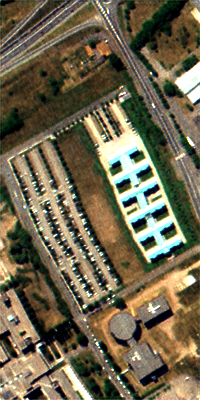}};
\spy [cyan, height=0.08\linewidth, width= .155\linewidth] on (0.05,-0.5) in node at (0,-1.8);
\node at (0,1.6) {SURE};
\end{scope} 
\filldraw [fill=white, draw=black] (-0.03\linewidth,-0.95) rectangle (0.07\linewidth,-1.35);
\node[anchor=west] at (-0.03\linewidth,-1.15){\footnotesize{$32.29$}};
\end{tikzpicture}
\begin{tikzpicture}
\tikzstyle{every node}=[font=\footnotesize]
\begin{scope}[node distance = 0mm, inner sep = 0pt, outer sep = 1,spy using outlines={rectangle, red, magnification=3.5, every spy on node/.append style={thick}}]
\node[align = center](img){\includegraphics[width= .155\linewidth]{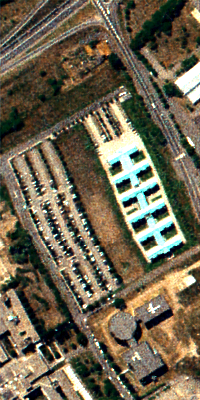}};
\spy [cyan, height=0.08\linewidth, width= .155\linewidth] on (0.05,-0.5) in node at (0,-1.8);
\node at (0,1.6) {DS2DP};
\end{scope} 
\filldraw [fill=white, draw=black] (-0.03\linewidth,-0.95) rectangle (0.07\linewidth,-1.35);
\node[anchor=west] at (-0.03\linewidth,-1.15){\footnotesize{$32.12$}};
\end{tikzpicture}
\begin{tikzpicture}
\tikzstyle{every node}=[font=\footnotesize]
\begin{scope}[node distance = 0mm, inner sep = 0pt, outer sep = 1,spy using outlines={rectangle, red, magnification=3.5, every spy on node/.append style={thick}}]
\node[align = center](img){\includegraphics[width= .155\linewidth]{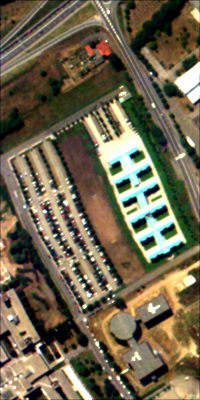}};
\spy [cyan, height=0.08\linewidth, width= .155\linewidth] on (0.05,-0.5) in node at (0,-1.8);
\node at (0,1.6) {Mixture-Net};
\end{scope} 
\filldraw [fill=white, draw=black] (-0.03\linewidth,-0.95) rectangle (0.07\linewidth,-1.35);
\node[anchor=west] at (-0.03\linewidth,-1.15){\footnotesize{$34.47$}};
\end{tikzpicture}
\begin{tikzpicture}
\tikzstyle{every node}=[font=\footnotesize]
\begin{scope}[node distance = 0mm, inner sep = 0pt, outer sep = 1,spy using outlines={rectangle, red, magnification=3.5, every spy on node/.append style={thick}}]
\node[align = center](img){\includegraphics[width= .155\linewidth]{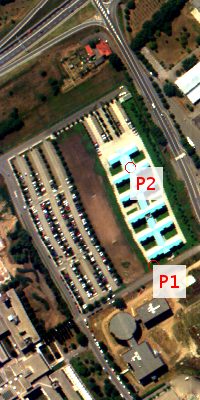}};
\spy [cyan, height=0.08\linewidth, width= .155\linewidth] on (0.05,-0.5) in node at (0,-1.8);
\node at (0,1.6) {GT};
\end{scope}
\filldraw [fill=white, draw=black] (-0.03\linewidth,-0.95) rectangle (0.07\linewidth,-1.35);
\node[anchor=west] at (-0.03\linewidth,-1.15){\footnotesize{PSNR}};
\end{tikzpicture}
 
\begin{tikzpicture}
\tikzstyle{every node}=[font=\footnotesize]
\begin{scope}[node distance = 0mm, inner sep = 5pt, outer sep = 1pt ,spy using outlines={circle, red, magnification=4.5, every spy on node/.append style={thick}}]
\node[align = center](img){\includegraphics[width= 1\linewidth]{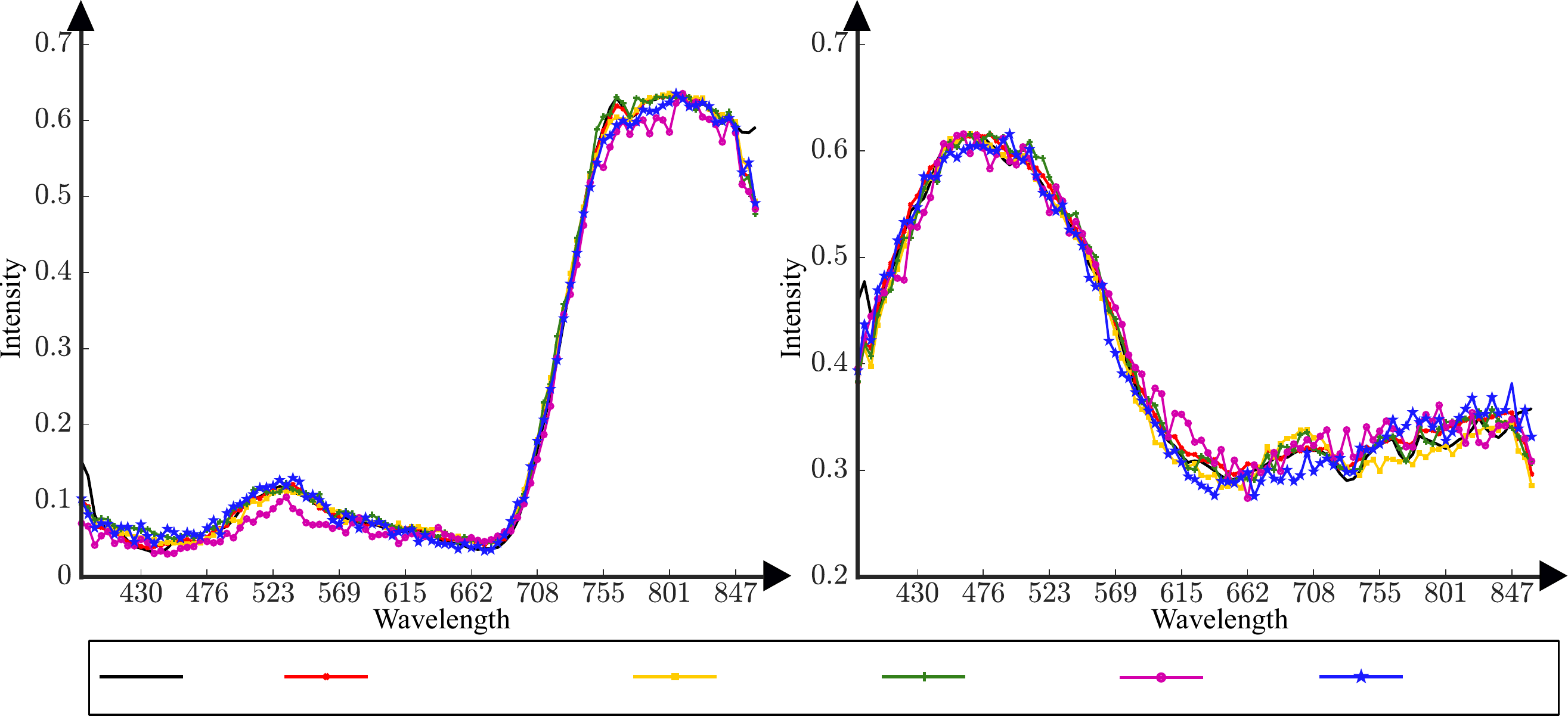}};
\spy [cyan, height=0.25\linewidth, width= .25\linewidth] on (-0.10\linewidth, 1.35) in node at (-2.55,0.7);
\spy [cyan, height=0.25\linewidth, width= .25\linewidth] on (0.27\linewidth, -0.6) in node at (3,0.75);
\node at (-0.25\linewidth,2.2) {Spectral Signature P1};
\node at (0.25\linewidth,2.2) {Spectral Signature P2};
\node at (-0.36\linewidth,-1.8) {GT};
\node at (-0.185\linewidth,-1.8) {Mixture-Net};
\node at (-0.00\linewidth,-1.8) {DS2DP};
\node at (0.16\linewidth,-1.8) {SURE};
\node at (0.295\linewidth,-1.8) {DIP};
\node at (0.445\linewidth,-1.8) {BM3D};
\end{scope}
\end{tikzpicture}
\caption{\dcheck{Visual RGB mapping of denoised Pavia University with noise level $\sigma = 50/255$. The white box at the bottom of each image shows the quantitative quality in PSNR. The zoomed sub-region shows that Mixture-Net suppresses the Gaussian noise in smooth regions while maintaining the shapes and structures. Unlike, the noise is still visible in HSI-DIP, SURE, and DS2DP methods, and the edges and structure are blurred in the BM3D case. A spectral quality comparison is shown at the bottom for two spatial pixels $\mathbf{P1}$ and $\mathbf{P2}$.}}
\label{fig:denoisingRGB}
\end{figure}
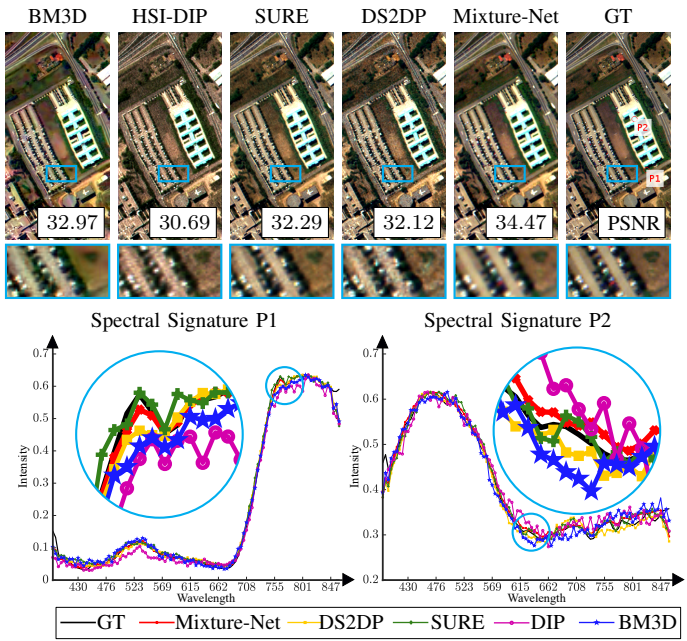

%% file: Tables/table2_3.tex
\begin{table}[ht!]
\scriptsize
\renewcommand{\arraystretch}{1.00}
\caption{Single Hyperspectral Super-Resolution Quantitative Results for Pavia Center}
\label{tab:sematic_depth}
\centering
\resizebox{1\columnwidth}{!}{%
\setlength\tabcolsep{0.1cm}
\begin{tabular}{lcccccc} 
\toprule 
Method & $d$ & SAM$\downarrow$ & RMSE$\downarrow$ & ERGAS$\downarrow$ & PSNR$\uparrow$ & SSIM $\uparrow$\\
\midrule 
Bicubic & 4 & 6.1399 & 0.0437 & 6.8814 & 27.5874 & 0.6961\\\midrule 
EDSR & 4 & 5.8657 &0.0379 & 6.0199 & 28.7981& 0.7722\\ 
RCAN & 4 & 5.9785 &0.0376 & 6.0485 & 28.8165& 0.7719\\ 
SAN & 4 & 5.9590 &0.0374 & 5.9903 & 28.8554& 0.7740\\ 
\midrule 
3DCNN & 4 & 5.8669 &0.0396 & 6.2665 & 28.4114& 0.7501\\ 
GDRRN & 4 & 5.4750 &0.0393 & 6.2264 & 28.4726& 0.7530\\ 
SSPSR & 4 &\underline{5.4612} &\underline{0.0362} &\textbf{5.8014} &\underline{29.1581} &\underline{0.7903}\\
\midrule 
DIP& 4 & 6.2665 &0.0410 & 6.4845 & 28.1061& 0.7365\\ 
Mixture-Net & 4 &\textbf{4.2120} &\textbf{0.0352} &\underline{5.8084} &\textbf{29.914}&\textbf{0.8396}\\
\bottomrule 
\midrule 
Bicubic & 8 & 7.8478 & 0.0630 & 4.8280 & 24.5972 & 0.4725\\\midrule 
EDSR & 8 & 7.8594 & 0.05983 & 4.6359 & 25.0041 & 0.5130\\
RCAN & 8 & 7.9992 & 0.0604 & 4.6930 & 24.9183 & 0.5086\\
SAN & 8 & 8.0371 & 0.0609 & 4.7646 & 24.8485 & 0.5054\\
\midrule 
3DCNN & 8 & 7.6878 & 0.0605 & 4.6469 & 24.9336 & 0.5038\\
GDRRN & 8 & 7.3531 & 0.0607 & 4.6220 & 24.8648 & 0.5014\\
SSPSR &8 &\underline{7.3312} &\underline{0.0586} &\underline{4.5266} &\underline{25.1985} &\underline{0.5365}\\
\midrule 
DIP & 8 &7.9281 &0.0618 &4.7366 &24.7252 &0.4963\\
Mixture-Net & 8 &\textbf{6.7855} &\textbf{0.0485} &\textbf{4.0015} &\textbf{26.9041} &\textbf{0.7148}\\
\bottomrule 
\end{tabular} }
\end{table}
\begin{table}[ht!]
\scriptsize
\renewcommand{\arraystretch}{1.00}
\caption{ Single Hyperspectral Super-Resolution Quantitative Results for Stuff-Toys }
\label{tab:sematic_depth_CAVE}
\centering
\resizebox{1\columnwidth}{!}{%
\setlength\tabcolsep{0.1cm}
\begin{tabular}{lcccccc} 
\toprule 
Method & $d$ & SAM$\downarrow$ & RMSE$\downarrow$ & ERGAS$\downarrow$ & PSNR$\uparrow$ & SSIM $\uparrow$\\
\midrule 
Bicubic & 4 & 4.1759 & 0.0212 & 5.2719 & 34.7214 & 0.9277\\\midrule 
EDSR & 4 & 3.5499 & 0.0149 & 3.5921 & 38.1575 & 0.9522\\
RCAN & 4 & 3.6050 & 0.0142 & 3.4178 & 38.7585 & 0.9530\\
SAN & 4 & 3.5951 & 0.0143 & 3.4200 & 38.7188 & 0.9531\\
\midrule 
3DCNN & 4 & 3.3463 & 0.0154 & 3.7042 & 37.9759 & 0.9522\\
GDRRN & 4 &\underline{3.4143} & 0.0145 & 3.5086 & 38.4507 & 0.9538\\
SSPSR & 4 &\textbf{3.1846} & 0.0138 & 3.3384 &\underline{39.0892} & 0.9553\\
\midrule
DIP & 4 & 8.4935 &\underline{0.0124} &\underline{ 2.5358} & 38.1329 &\underline{0.9631}\\
Mixture-Net &4& 5.3285&\textbf{0.0110} &\textbf{2.1997} &\textbf{39.1640} &\textbf{0.9821}\\
\bottomrule 
\midrule 
Bicubic & 8 & 5.8962 & 0.0346 & 4.2175 & 30.2056 & 0.8526\\\midrule 
EDSR & 8 & 5.6865 & 0.0279 & 3.3903 & 32.4072 & 0.8842\\
RCAN & 8 & 5.9771 & 0.0268 & 3.1781 & 32.9544 & 0.8884\\
SAN & 8 & 5.8683 & 0.0267 & 3.1437 & 33.0012 & 0.8888\\
\midrule 
3DCNN & 8 &\underline{5.0948} & 0.0292 & 3.5536 & 31.9691 & 0.8863\\ 
GDRRN & 8 & 5.3597 & 0.0280 & 3.3460 & 32.5763 & 0.8890\\
SSPSR & 8 &\textbf{4.4874} & 0.0257 & 3.0419 &
\underline{33.4340} & 0.9010\\
\midrule
DIP & 8 & 8.3342 &\underline{0.0231} &\underline{2.3697}& 32.7324&\underline{0.9291}\\
Mixture-Net &8& 5.5027&\textbf{0.0208} &\textbf{2.1061} &\textbf{33.6270} &\textbf{0.9432}\\
\bottomrule 
\end{tabular} }
\end{table}

%% file: tikz/figure7.tex
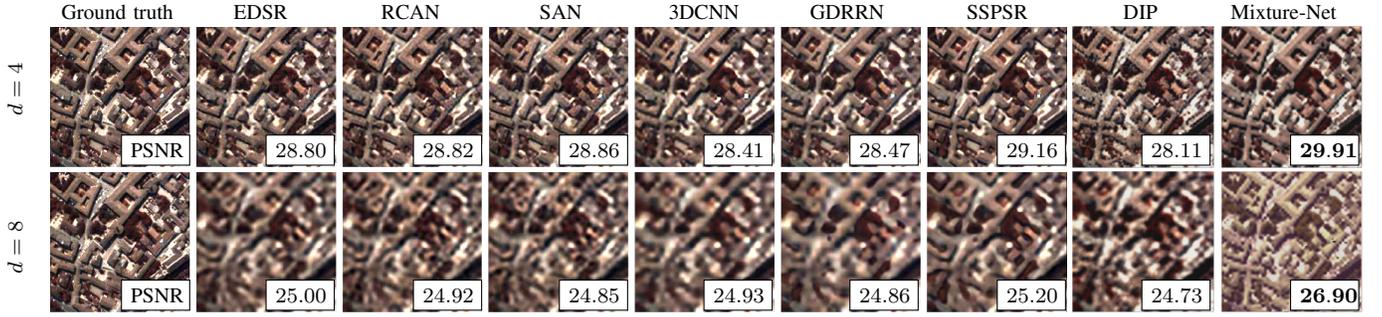
\begin{figure*}[htb!]
\begin{tikzpicture}
\tikzstyle{every node}=[font=\footnotesize]
\node[inner sep=0pt] (Abun) at (0\linewidth,0){\includegraphics[width=1\linewidth]{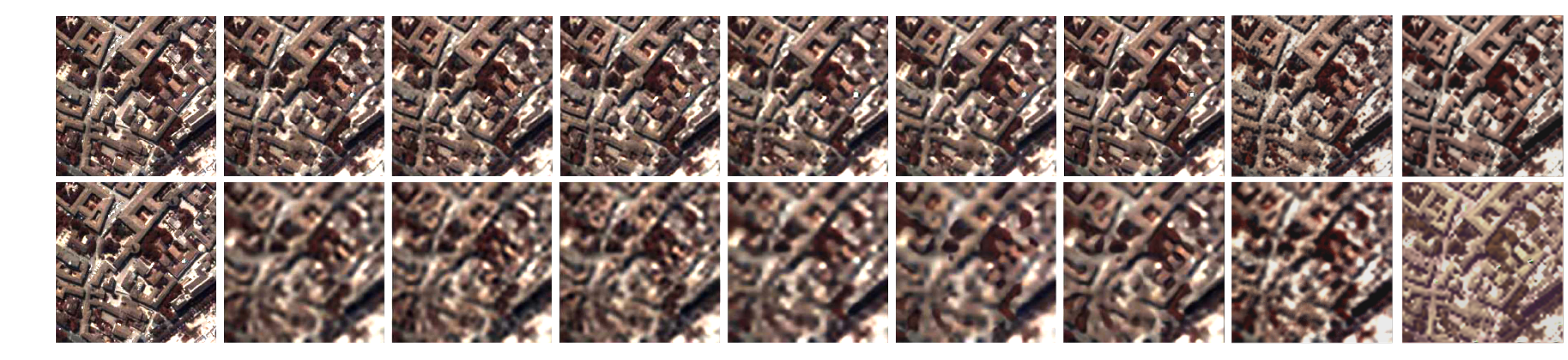}};
\node[rotate=90, anchor=west] at (-0.49\linewidth,0.6cm){$d=4$};
\node[rotate=90, anchor=west] at (-0.49\linewidth,-1.5cm){$d=8$};
\node at (-0.415\linewidth,2.1cm) {Ground truth};
\node at (-0.31\linewidth,2.1cm) {EDSR};
\node at (-0.20\linewidth,2.1cm) {RCAN};
\node at (-0.09\linewidth,2.1cm) {SAN};
\node at (0.015\linewidth,2.1cm) {3DCNN};
\node at (0.12\linewidth,2.1cm) {GDRRN};
\node at (0.23\linewidth,2.1cm) {SSPSR};
\node at (0.335\linewidth,2.1cm) {DIP};
\node at (0.44\linewidth,2.1cm) {Mixture-Net};
\filldraw [fill=white, draw=black] (-0.412\linewidth,-1.47) rectangle (-0.362\linewidth,-1.87);
\node[anchor=west] at (-0.412\linewidth,-1.67){\footnotesize{PSNR}};
\filldraw [fill=white, draw=black] (-0.412\linewidth,0.46) rectangle (-0.362\linewidth,0.06);
\node[anchor=west] at (-0.412\linewidth,0.26){\footnotesize{PSNR}};
\filldraw [fill=white, draw=black] (-0.306\linewidth,-1.47) rectangle (-0.256\linewidth,-1.87);
\node[anchor=west] at (-0.306\linewidth,-1.67){\footnotesize{$25.00$}};
\filldraw [fill=white, draw=black] (-0.306\linewidth,0.46) rectangle (-0.256\linewidth,0.06);
\node[anchor=west] at (-0.306\linewidth,0.26){\footnotesize{$28.80$}};
\filldraw [fill=white, draw=black] (-0.198\linewidth,-1.47) rectangle (-0.148\linewidth,-1.87);
\node[anchor=west] at (-0.198\linewidth,-1.67){\footnotesize{$24.92$}};
\filldraw [fill=white, draw=black] (-0.198\linewidth,0.46) rectangle (-0.148\linewidth,0.06);
\node[anchor=west] at (-0.198\linewidth,0.26){\footnotesize{$28.82$}};
\filldraw [fill=white, draw=black] (-0.091\linewidth,-1.47) rectangle (-0.041\linewidth,-1.87);
\node[anchor=west] at (-0.091\linewidth,-1.67){\footnotesize{$24.85$}};
\filldraw [fill=white, draw=black] (-0.091\linewidth,0.46) rectangle (-0.041\linewidth,0.06);
\node[anchor=west] at (-0.091\linewidth,0.26){\footnotesize{$28.86$}};
\filldraw [fill=white, draw=black] (0.015\linewidth,-1.47) rectangle (0.065\linewidth,-1.87);
\node[anchor=west] at (0.015\linewidth,-1.67){\footnotesize{$24.93$}};
\filldraw [fill=white, draw=black] (0.015\linewidth,0.46) rectangle (0.065\linewidth,0.06);
\node[anchor=west] at (0.015\linewidth,0.26){\footnotesize{$28.41$}};
\filldraw [fill=white, draw=black] (0.122\linewidth,-1.47) rectangle (0.172\linewidth,-1.87);
\node[anchor=west] at (0.122\linewidth,-1.67){\footnotesize{$24.86$}};
\filldraw [fill=white, draw=black] (0.122\linewidth,0.46) rectangle (0.172\linewidth,0.06);
\node[anchor=west] at (0.122\linewidth,0.26){\footnotesize{$28.47$}};
\filldraw [fill=white, draw=black] (0.23\linewidth,-1.47) rectangle (0.28\linewidth,-1.87);
\node[anchor=west] at (0.23\linewidth,-1.67){\footnotesize{$25.20$}};
\filldraw [fill=white, draw=black] (0.23\linewidth,0.46) rectangle (0.28\linewidth,0.06);
\node[anchor=west] at (0.23\linewidth,0.26){\footnotesize{$29.16$}};
\filldraw [fill=white, draw=black] (0.336\linewidth,-1.47) rectangle (0.386\linewidth,-1.87);
\node[anchor=west] at (0.336\linewidth,-1.67){\footnotesize{$24.73$}};
\filldraw [fill=white, draw=black] (0.336\linewidth,0.46) rectangle (0.386\linewidth,0.06);
\node[anchor=west] at (0.336\linewidth,0.26){\footnotesize{$28.11$}};
\filldraw [fill=white, draw=black] (0.445\linewidth,-1.47) rectangle (0.495\linewidth,-1.87);
\node[anchor=west] at (0.445\linewidth,-1.67){\footnotesize{$\mathbf{26.90}$}};
\filldraw [fill=white, draw=black] (0.445\linewidth,0.46) rectangle (0.495\linewidth,0.06);
\node[anchor=west] at (0.445\linewidth,0.26){\footnotesize{$\mathbf{29.91}$}};
\end{tikzpicture}
\caption{RGB representation of the reconstructed composite images of Pavia Center datatest for spatial downsampling factor $d = 4$ (\textit{top}) and $d=8$ (\textit{bottom}). Notice that, the spatial quality is especially improved for a factor $d=8$.}
\label{fig:visualPavia}
\end{figure*}

%% file: tikz/figure8.tex
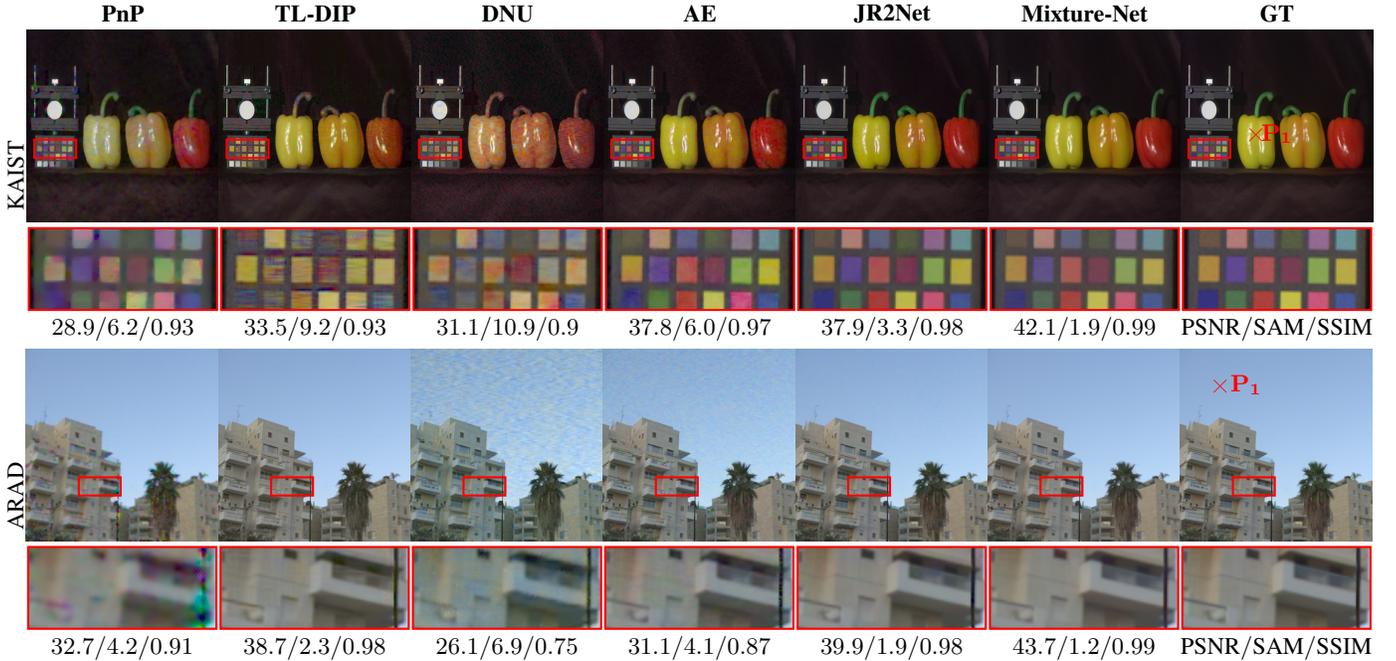
\begin{figure*}[htb!]
\begin{tikzpicture}[outer sep=1pt,inner sep=1pt]
\tikzstyle{every node}=[font=\small]
\node[rotate=90] at (0.0\linewidth,0cm){\hspace{17mm}KAIST};
\end{tikzpicture}\begin{tikzpicture}[outer sep=0pt,inner sep=1pt]\tikzstyle{every node}=[font=\small]
\begin{scope}[node distance = 0mm, inner sep = 0pt, outer sep = 1,spy using outlines={rectangle, red, magnification=4.5, every spy on node/.append style={thick}}]
\node[align = center](img){\includegraphics[width= .141\linewidth]{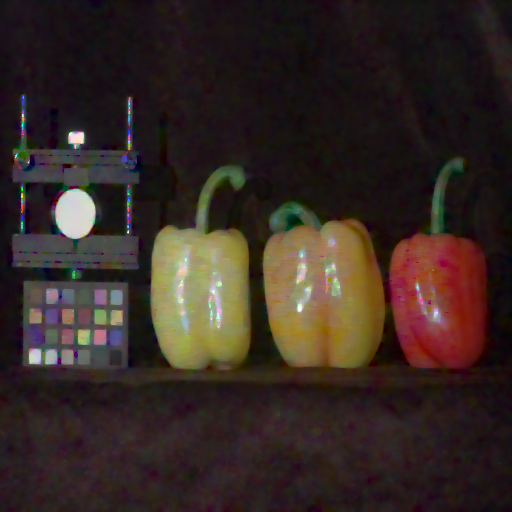}};
\spy [red, height=0.06\linewidth, width= .138\linewidth] on (-0.9,-0.3) in node at (0,-1.9);
\node at (0,-2.7) {$28.9 / 6.2 / 0.93$};
\node at (0\linewidth,1.5) {\textbf{PnP}};
\end{scope}
\end{tikzpicture}\begin{tikzpicture}[outer sep=0pt,inner sep=1pt]\tikzstyle{every node}=[font=\small]
\begin{scope}[node distance = 0mm, inner sep = 0pt, outer sep = 1,spy using outlines={rectangle, red, magnification=4.5, every spy on node/.append style={thick}}]
\node[align = center](img){\includegraphics[width= .141\linewidth]{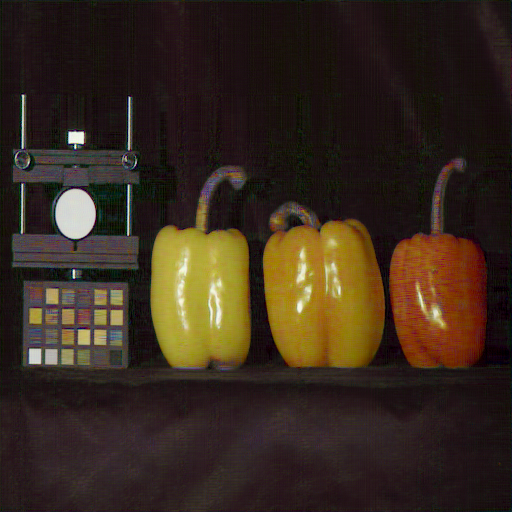}};
\spy [red, height=0.06\linewidth, width= .138\linewidth] on (-0.9,-0.3) in node at (0,-1.9);
\node at (0,-2.7) {{$33.5 / 9.2 / 0.93$}};
\node at (0\linewidth,1.5) {\textbf{TL-DIP}};
\end{scope}
\end{tikzpicture}\begin{tikzpicture}[outer sep=0pt,inner sep=1pt]\tikzstyle{every node}=[font=\small]
\begin{scope}[node distance = 0mm, inner sep = 0pt, outer sep = 1,spy using outlines={rectangle, red, magnification=4.5, every spy on node/.append style={thick}}]
\node[align = center](img){\includegraphics[width= .141\linewidth]{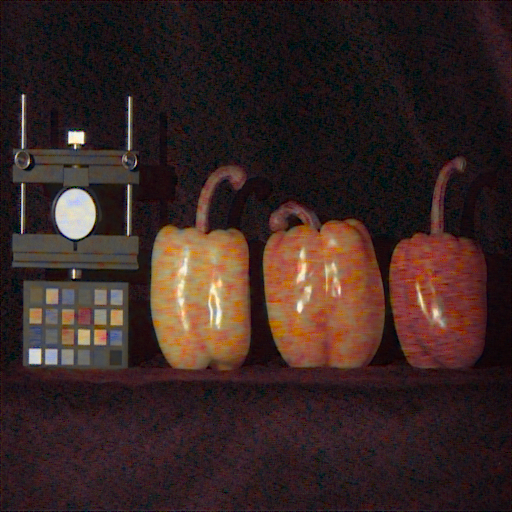}};
\spy [red, height=0.06\linewidth, width= .138\linewidth] on (-0.9,-0.3) in node at (0,-1.9);
\node at (0,-2.7) {$31.1 / 10.9 / 0.9$};
\node at (0\linewidth,1.5) {\textbf{DNU}};
\end{scope}
\end{tikzpicture}\begin{tikzpicture}[outer sep=0pt,inner sep=1pt]\tikzstyle{every node}=[font=\small]
\begin{scope}[node distance = 0mm, inner sep = 0pt, outer sep = 1,spy using outlines={rectangle, red, magnification=4.5, every spy on node/.append style={thick}}]
\node[align = center](img){\includegraphics[width= .141\linewidth]{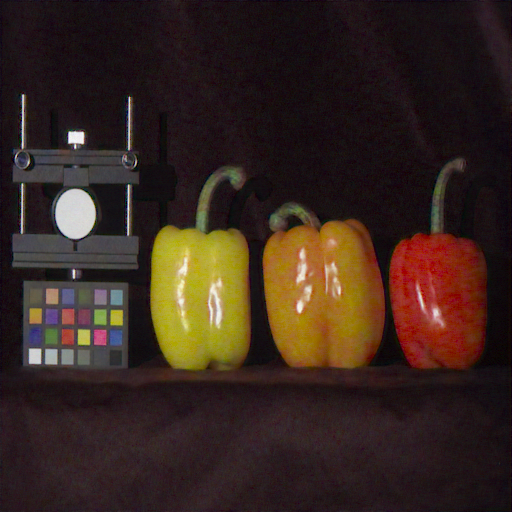}};
\spy [red, height=0.06\linewidth, width= .138\linewidth] on (-0.9,-0.3) in node at (0,-1.9);
\node at (0,-2.7) {$37.8 / 6.0 / 0.97$};
\node at (0\linewidth,1.5) {\textbf{AE}};
\end{scope}
\end{tikzpicture}\begin{tikzpicture}[outer sep=0pt,inner sep=1pt]\tikzstyle{every node}=[font=\small]
\begin{scope}[node distance = 0mm, inner sep = 0pt, outer sep = 1,spy using outlines={rectangle, red, magnification=4.5, every spy on node/.append style={thick}}]
\node[align = center](img){\includegraphics[width= .141\linewidth]{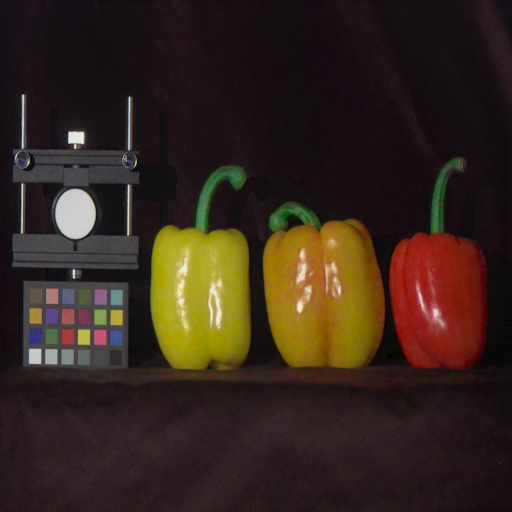}};
\spy [red, height=0.06\linewidth, width= .138\linewidth] on (-0.9,-0.3) in node at (0,-1.9);
\node at (0,-2.7) {$37.9 / 3.3 / 0.98$};
\node at (0\linewidth,1.5) {\textbf{JR2Net}};
\end{scope}
\end{tikzpicture}\begin{tikzpicture}[outer sep=0pt,inner sep=1pt]\tikzstyle{every node}=[font=\small]
\begin{scope}[node distance = 0mm, inner sep = 0pt, outer sep = 1,spy using outlines={rectangle, red, magnification=4.5, every spy on node/.append style={thick}}]
\node[align = center](img){\includegraphics[width= .141\linewidth]{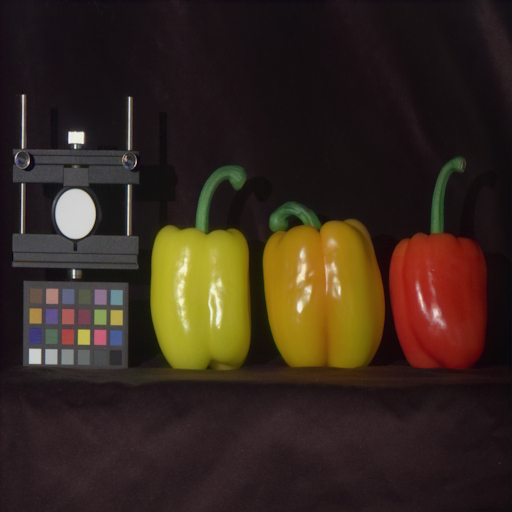}};
\spy [red, height=0.06\linewidth, width= .138\linewidth] on (-0.9,-0.3) in node at (0,-1.9);
\node at (0,-2.7) {${42.1 / 1.9 / 0.99}$};
\node at (0\linewidth,1.5) {\textbf{Mixture-Net}};
\end{scope}
\end{tikzpicture}\begin{tikzpicture}[outer sep=0pt,inner sep=1pt]\tikzstyle{every node}=[font=\small]
\begin{scope}[node distance = 0mm, inner sep = 0pt, outer sep = 1,spy using outlines={rectangle, red, magnification=4.5, every spy on node/.append style={thick}}]
\node[align = center](img){\includegraphics[width= .141\linewidth]{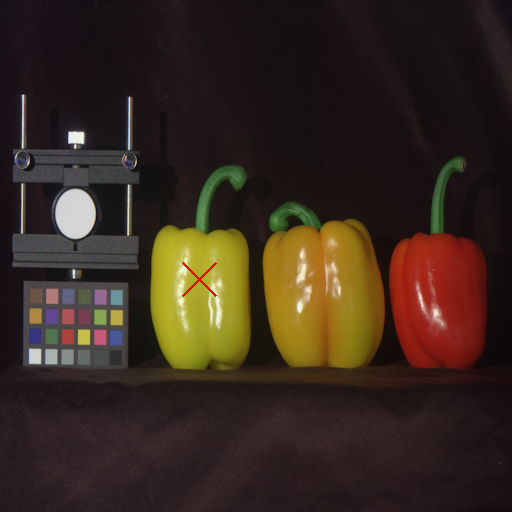}};
\spy [red, height=0.06\linewidth, width= .138\linewidth] on (-0.9,-0.3) in node at (0,-1.9);
\node at (0,-2.7) {$\mbox{PSNR} /\mbox{SAM} /\mbox{SSIM}$};
\node at (0\linewidth,1.5) {\textbf{GT}};
\node at (0.0,-0.1) {\textcolor{red}{$\mathbf{P_1}$}};
\end{scope}
\end{tikzpicture}

\smallbreak
\begin{tikzpicture}[outer sep=1pt,inner sep=1pt]
\tikzstyle{every node}=[font=\small]
\node[rotate=90] at (0.2\linewidth,0cm) {\hspace{17mm}ARAD};
\end{tikzpicture}\begin{tikzpicture}[outer sep=0pt,inner sep=1pt]\tikzstyle{every node}=[font=\small]
\begin{scope}[node distance = 0mm, inner sep = 0pt, outer sep = 1,spy using outlines={rectangle, red, magnification=4.5, every spy on node/.append style={thick}}]
\node[align = center](img){\includegraphics[width= .141\linewidth]{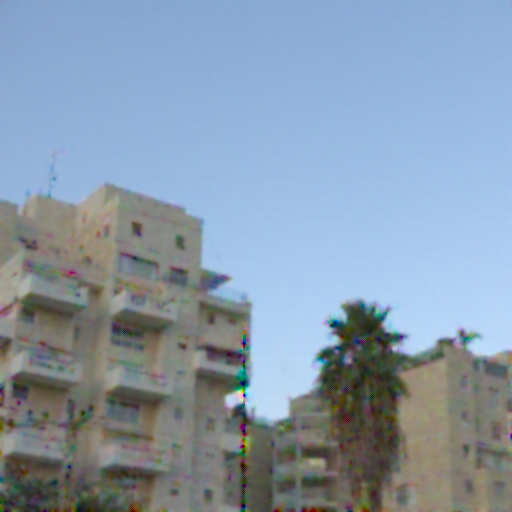}};
\spy [red, height=0.06\linewidth, width= .138\linewidth] on (-0.3,-0.55) in node at (0,-1.9);
\node at (0,-2.7) {$32.7 / 4.2 / 0.91$};
\end{scope}
\end{tikzpicture}\begin{tikzpicture}[outer sep=0pt,inner sep=1pt]\tikzstyle{every node}=[font=\small]
\begin{scope}[node distance = 0mm, inner sep = 0pt, outer sep = 1,spy using outlines={rectangle, red, magnification=4.5, every spy on node/.append style={thick}}]
\node[align = center](img){\includegraphics[width= .141\linewidth]{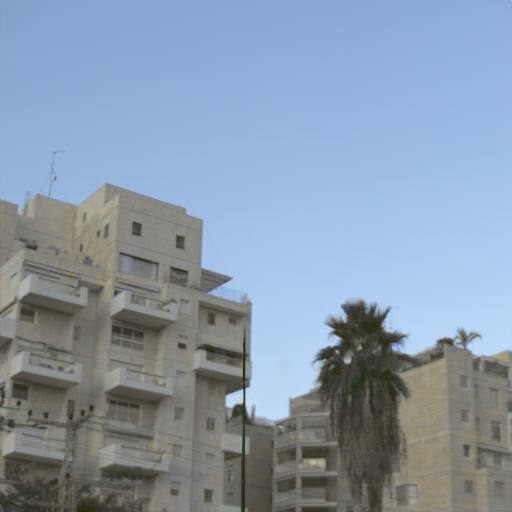}};
\spy [red, height=0.06\linewidth, width= .138\linewidth] on (-0.3,-0.55) in node at (0,-1.9);
\node at (0,-2.7) {{$38.7 / 2.3 / 0.98$}};
\end{scope}
\end{tikzpicture}\begin{tikzpicture}[outer sep=0pt,inner sep=1pt]\tikzstyle{every node}=[font=\small]
\begin{scope}[node distance = 0mm, inner sep = 0pt, outer sep = 1,spy using outlines={rectangle, red, magnification=4.5, every spy on node/.append style={thick}}]
\node[align = center](img){\includegraphics[width= .141\linewidth]{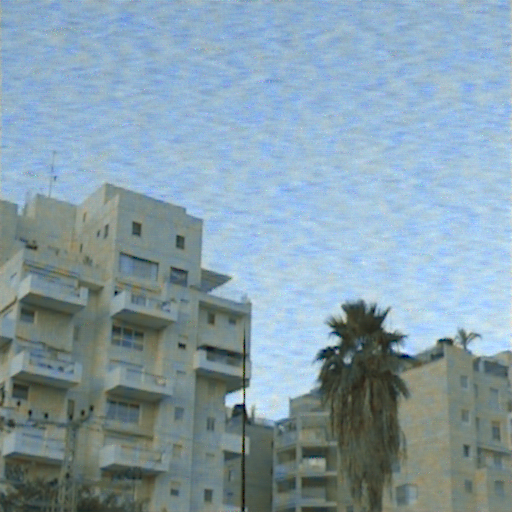}};
\spy [red, height=0.06\linewidth, width= .138\linewidth] on (-0.3,-0.55) in node at (0,-1.9);
\node at (0,-2.7) {$26.1 / 6.9 / 0.75$};
\end{scope}
\end{tikzpicture}\begin{tikzpicture}[outer sep=0pt,inner sep=1pt]\tikzstyle{every node}=[font=\small]
\begin{scope}[node distance = 0mm, inner sep = 0pt, outer sep = 1,spy using outlines={rectangle, red, magnification=4.5, every spy on node/.append style={thick}}]
\node[align = center](img){\includegraphics[width= .141\linewidth]{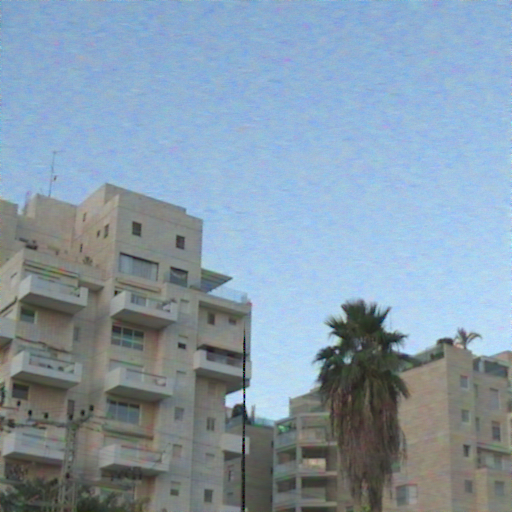}};
\spy [red, height=0.06\linewidth, width= .138\linewidth] on (-0.3,-0.55) in node at (0,-1.9);
\node at (0,-2.7) {$31.1 / 4.1 / 0.87$};
\end{scope}
\end{tikzpicture}\begin{tikzpicture}[outer sep=0pt,inner sep=1pt]\tikzstyle{every node}=[font=\small]
\begin{scope}[node distance = 0mm, inner sep = 0pt, outer sep = 1,spy using outlines={rectangle, red, magnification=4.5, every spy on node/.append style={thick}}]
\node[align = center](img){\includegraphics[width= .141\linewidth]{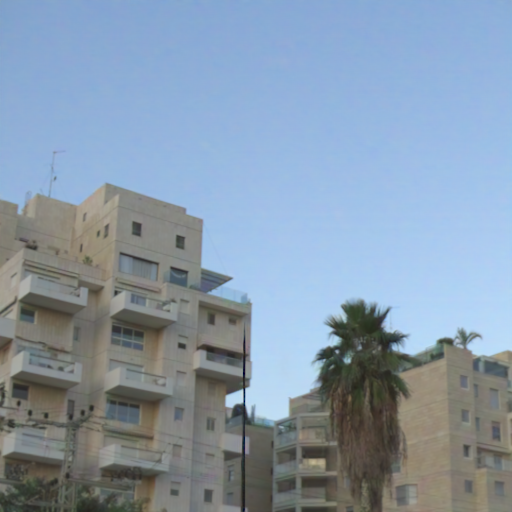}};
\spy [red, height=0.06\linewidth, width= .138\linewidth] on (-0.3,-0.55) in node at (0,-1.9);
\node at (0,-2.7) {$39.9 / 1.9 / 0.98 $};
\end{scope}
\end{tikzpicture}\begin{tikzpicture}[outer sep=0pt,inner sep=1pt]\tikzstyle{every node}=[font=\small]
\begin{scope}[node distance = 0mm, inner sep = 0pt, outer sep = 1,spy using outlines={rectangle, red, magnification=4.5, every spy on node/.append style={thick}}]
\node[align = center](img){\includegraphics[width= .141\linewidth]{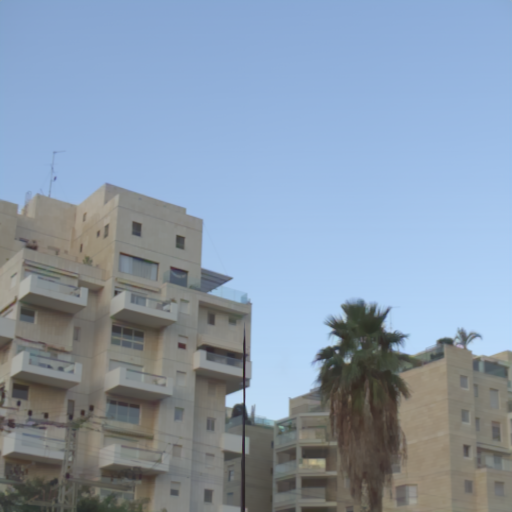}};
\spy [red, height=0.06\linewidth, width= .138\linewidth] on (-0.3,-0.55) in node at (0,-1.9);
\node at (0,-2.7) {$43.7 / 1.2 / 0.99$};
\end{scope}
\end{tikzpicture}\begin{tikzpicture}[outer sep=0pt,inner sep=1pt]\tikzstyle{every node}=[font=\small]
\begin{scope}[node distance = 0mm, inner sep = 0pt, outer sep = 1,spy using outlines={rectangle, red, magnification=4.5, every spy on node/.append style={thick}}]
\node[align = center](img){\includegraphics[width= .141\linewidth]{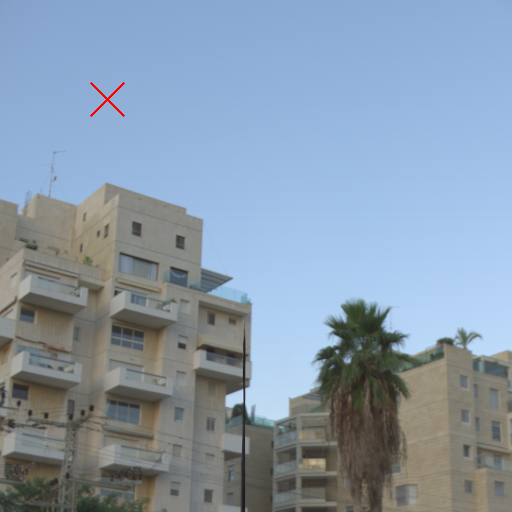}};
\spy [red, height=0.06\linewidth, width= .138\linewidth] on (-0.3,-0.55) in node at (0,-1.9);
\node at (0,-2.7) {$\mbox{PSNR} /\mbox{SAM} /\mbox{SSIM}$};
\node at (-0.4,0.8) {\textcolor{red}{$\mathbf{P_1}$}};
\end{scope}
\end{tikzpicture}
\caption{\dcheck{Compressive sensing reconstruction quality comparison using the non-data-driven PnP and DIP methods, and the training-data dependent DNU, and AE methods against the interpretable method Mixture-Net for the KAIST and ARAD datasets. The quality is measured in terms of PSNR / SSIM.}}
\label{fig:comparison}
\end{figure*}

%% file: Tables/table4.tex
\begin{table}[h!]
\scriptsize
\renewcommand{\arraystretch}{1.25}
\caption{Compressive Spectral Imaging Reconstruction Quantitative Results}
\label{tab:compressive_spectral_reconstruction}
\centering
\resizebox{1\columnwidth}{!}{%
\setlength\tabcolsep{0.05cm}
\begin{tabular}{lllllll} 
\multicolumn{7}{c}{\textbf{KAIST}}\\
\toprule\multicolumn{1}{c}{\textbf{Metric}} &\multicolumn{1}{c}{\textbf{PnP}} &\multicolumn{1}{c}{\textbf{DNU}} &\multicolumn{1}{c}{\textbf{AE}} &\multicolumn{1}{c}{\textbf{TL-DIP}} &\multicolumn{1}{c}{\textbf{\dcheck{JR2net}}} &\multicolumn{1}{c}{\textbf{Mixture-Net}}\\
\bottomrule
\midrule
$\uparrow$PSNR & 
$29.8\pm2.6$ &	
$32.6\pm1.8$ & 
$37.9\pm1.1$ & 
$32.8\pm0.9$ &
\dcheck{\underline{$39.1\pm2.5$}} & 	
$\mathbf{40.3\pm2.6}$ \\ 
$\uparrow$SSIM & 
$0.92\pm0.02$ & 
$0.91\pm0.01$ &	
$0.96\pm0.01$ &		
$0.92\pm0.01$ &
\dcheck{\underline{$0.97\pm0.02$}}& 
$\mathbf{0.99\pm0.01}$ \\ 
$\downarrow$SAM & 
$10.9\pm4.2$ & 
$12.9\pm1.8$ &	
$9.20\pm2.9$ &	
$12.4\pm2.7$ &	
\dcheck{\underline{$6.00\pm2.3$}} &	
$\mathbf{3.26\pm1.1}$\\ 
\midrule 
\multicolumn{7}{c}{\textbf{ARAD}}\\
\toprule
$\uparrow$PSNR & 
$31.5\pm4.5$ &	
$29.5\pm4.5$ &	 
$33.5\pm4.4$ &
$35.8\pm3.8$ &	
\dcheck{\underline{$36.5\pm4.6$}} &	
$\mathbf{39.1\pm4.1}$ \\ 
$\uparrow$SSIM & 
$0.83\pm0.08$ &	
$0.82\pm0.09$ &	
$0.90\pm0.05$ &	
$0.95\pm0.03$ &	
\dcheck{\underline{$0.97\pm0.01$}} &	
$\mathbf{0.98\pm0.01}$ \\ 
$\downarrow$SAM & 
$6.94\pm2.7$ &	
$8.26\pm3.1$ &	
$5.50\pm1.7$ &	
$4.45\pm2.5$ &	
\dcheck{\underline{$2.85\pm0.95$}} &	
$\mathbf{2.04\pm0.53}$\\
\bottomrule
\end{tabular}}
\end{table}

%% file: tikz/figure10.tex
\begin{figure*}[b!]
\begin{tikzpicture}[outer sep=0pt,inner sep=1pt]
\tikzstyle{every node}=[font=\small]
\node[rotate=90] at (0.2\linewidth,0cm) {\hspace{6mm}Ground truth};
\end{tikzpicture}\begin{tikzpicture}[outer sep=0pt,inner sep=1pt]\tikzstyle{every node}=[font=\small]
\node [align = center](img){\includegraphics[width=0.159\textwidth]{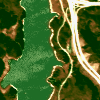}} (img.north west) node[xshift=1.45cm, yshift=0.2cm] {\textbf{False RGB}};
\end{tikzpicture}\begin{tikzpicture}[outer sep=0pt,inner sep=1pt]\tikzstyle{every node}=[font=\small]
\node [align = center](img){\includegraphics[width=0.159\textwidth]{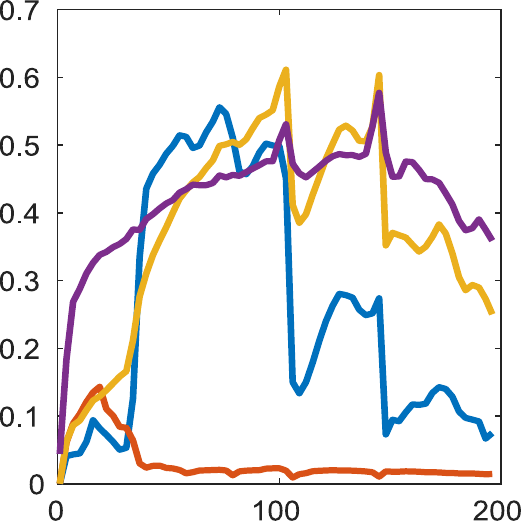}} (img.north west) node[xshift=1.5cm, yshift=0.2cm] {\textbf{Endmembers}};
\end{tikzpicture}\begin{tikzpicture}[outer sep=0pt,inner sep=1pt]\tikzstyle{every node}=[font=\small]
\node [align = center](img){\includegraphics[width=0.159\textwidth]{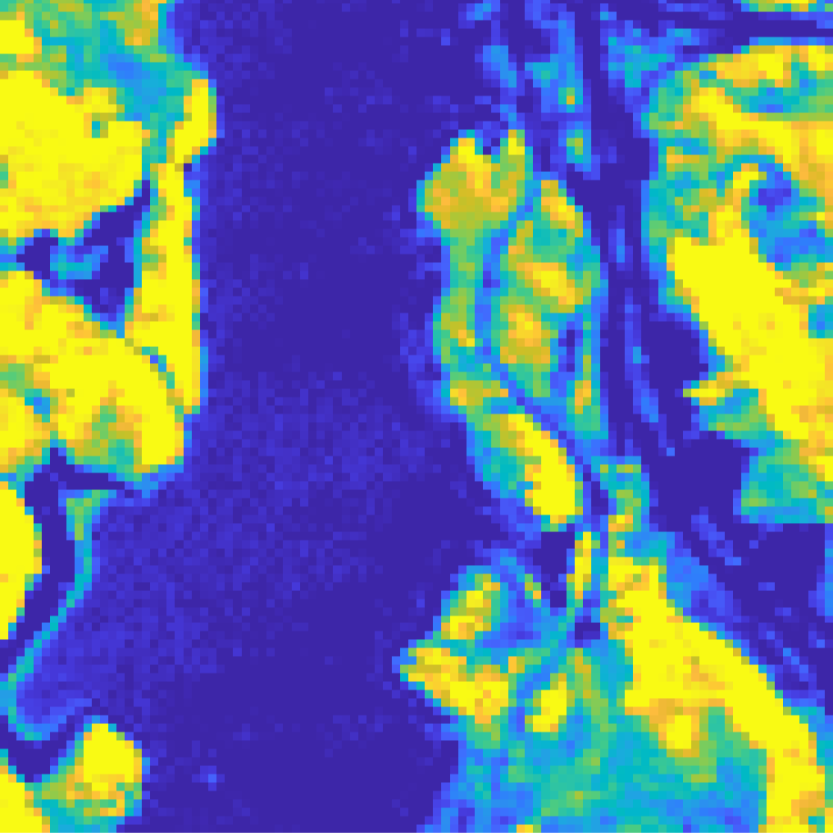}} (img.north west) node[xshift=1.45cm, yshift=0.2cm] {\textbf{Abundance} $1$};
\end{tikzpicture}\begin{tikzpicture}[outer sep=0pt,inner sep=1pt]\tikzstyle{every node}=[font=\small]
\node [align = center](img){\includegraphics[width=0.159\textwidth]{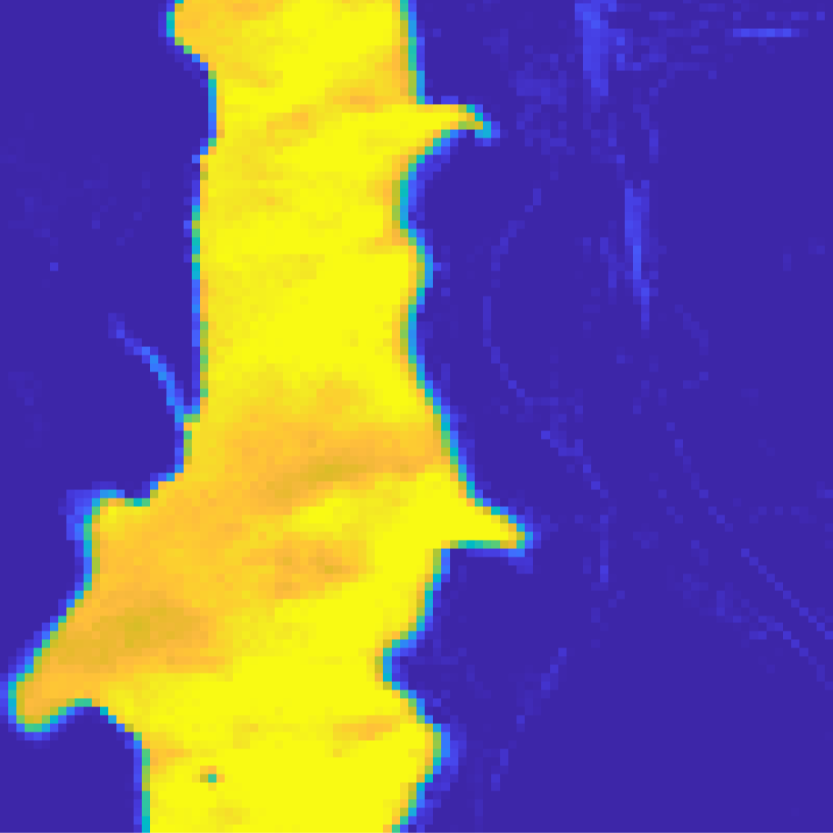}} (img.north west) node[xshift=1.45cm, yshift=0.2cm] {\textbf{Abundance} $2$};
\end{tikzpicture}\begin{tikzpicture}[outer sep=0pt,inner sep=1pt]\tikzstyle{every node}=[font=\small]
\node [align = center](img){\includegraphics[width=0.159\textwidth]{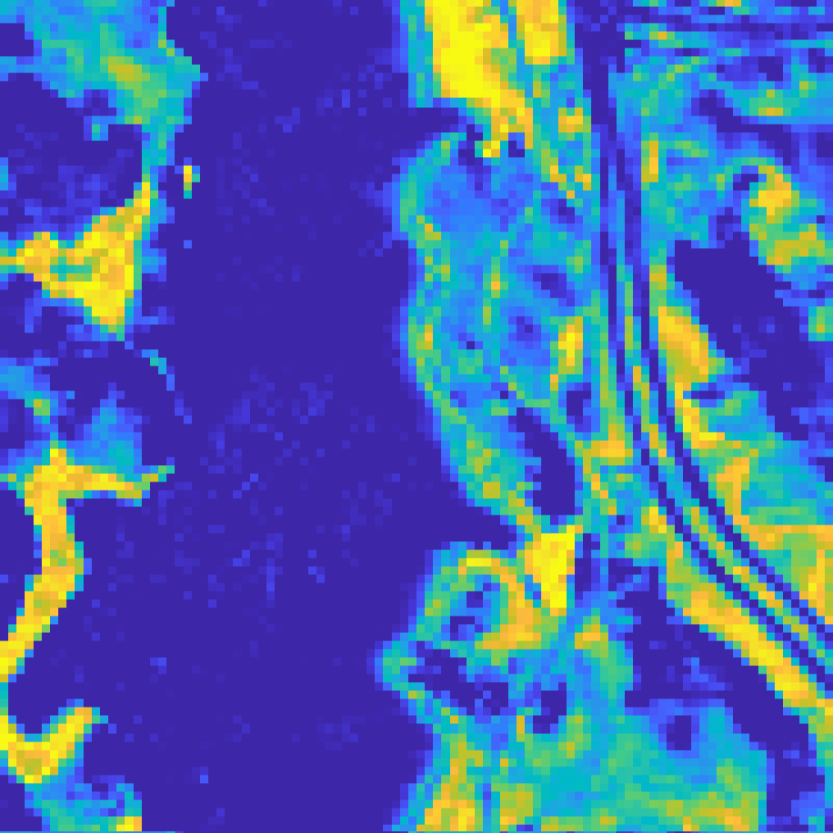}\baselineskip=1pt} (img.north west) node[xshift=1.45cm, yshift=0.2cm] {\textbf{Abundance $3$}};
\end{tikzpicture}\hspace{-2.2pt}\begin{tikzpicture}[outer sep=0pt,inner sep=1pt]\tikzstyle{every node}=[font=\small]
\node [align = center](img){\includegraphics[width=0.159\textwidth]{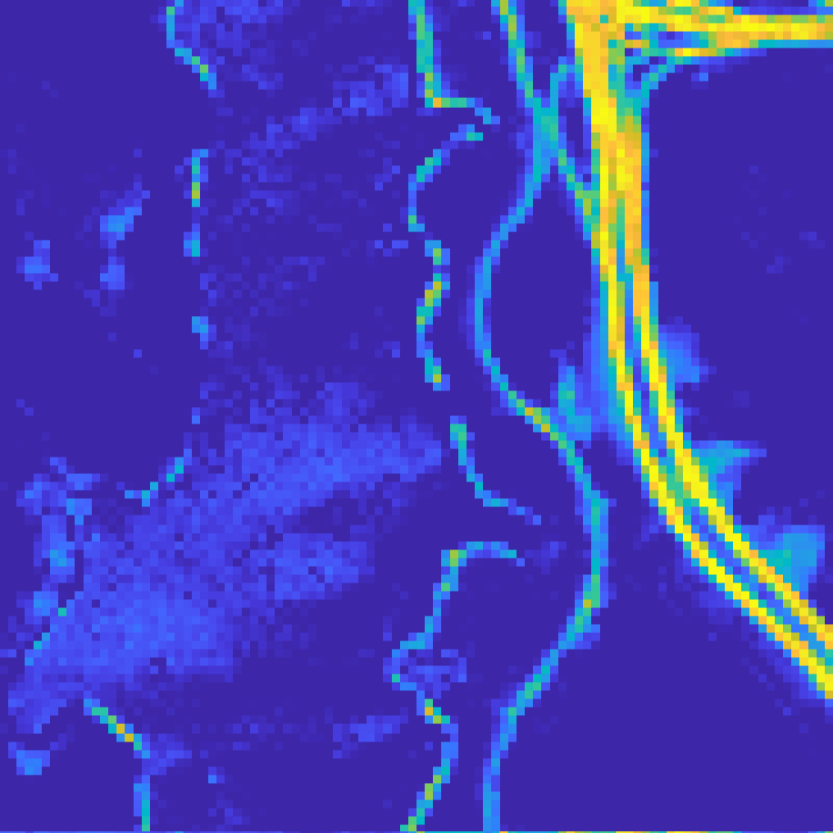}} (img.north west) node[xshift=1.45cm, yshift=0.2cm] {\textbf{Abundance} $4$};
\end{tikzpicture}\smallbreak\vspace{-3mm}\begin{tikzpicture}[outer sep=0pt,inner sep=-0pt]
\draw[dashdotted] (0,0)--(\linewidth,0);
\end{tikzpicture}

\smallbreak
\begin{tikzpicture}[outer sep=0pt,inner sep=1pt]
\tikzstyle{every node}=[font=\small]
\node[rotate=90] at (0.2\linewidth,0cm) {\hspace{8mm}Estimated};
\end{tikzpicture}\begin{tikzpicture}[outer sep=0pt,inner sep=1pt]\tikzstyle{every node}=[font=\small]
\node [align = center](img){\includegraphics[width=0.159\textwidth]{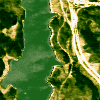}};
\filldraw [fill=white, draw=black] (0.2,-0.8) rectangle (1.35,-1.3);
\node at (0.2,-1.05) [anchor=west] {$29.23$dB};
\end{tikzpicture}\begin{tikzpicture}[outer sep=0pt,inner sep=1pt]\tikzstyle{every node}=[font=\small]
\node [align = center](img){\includegraphics[width=0.159\textwidth]{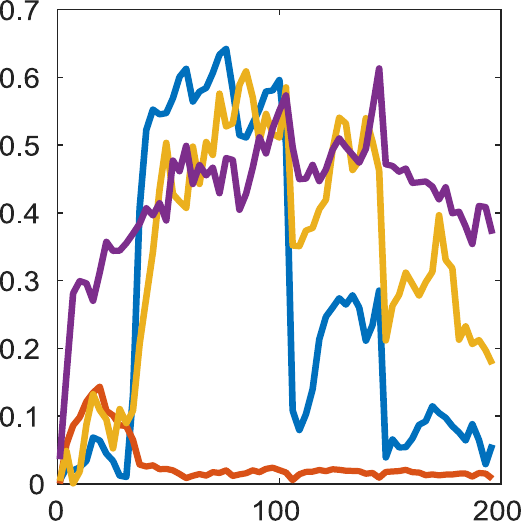}};\end{tikzpicture}\begin{tikzpicture}[outer sep=0pt,inner sep=1pt]\tikzstyle{every node}=[font=\small]
\node [align = center](img){\includegraphics[width=0.159\textwidth]{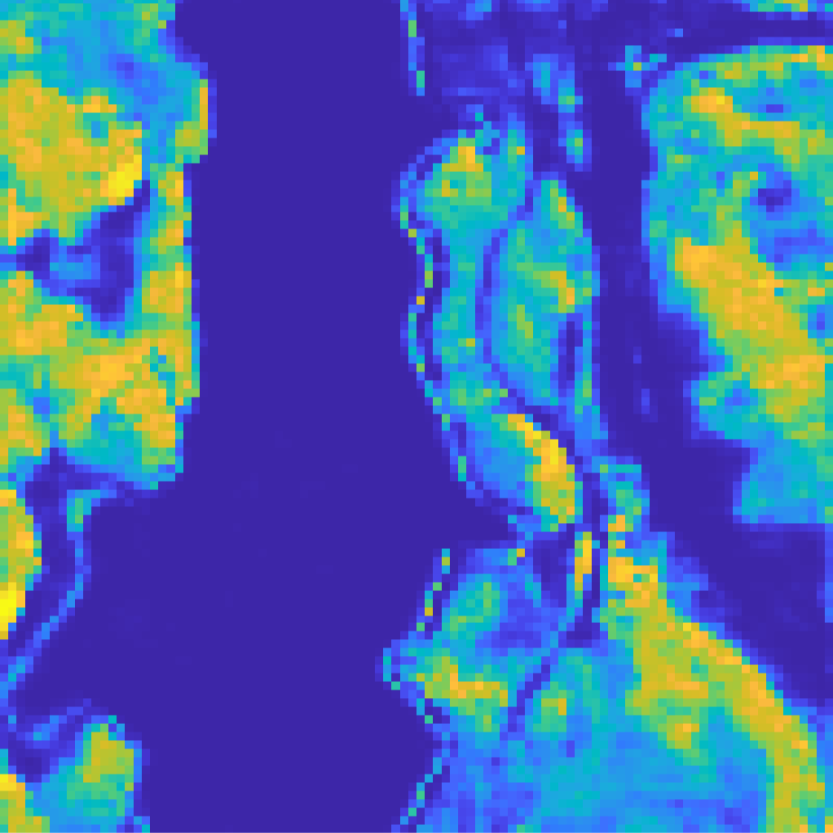}};
\filldraw [fill=white, draw=black] (0.2,-0.8) rectangle (1.35,-1.3);
\node at (0.2,-1.05) [anchor=west] {$15.29$dB};
\end{tikzpicture}\begin{tikzpicture}[outer sep=0pt,inner sep=1pt]\tikzstyle{every node}=[font=\small]
\node [align = center](img){\includegraphics[width=0.159\textwidth]{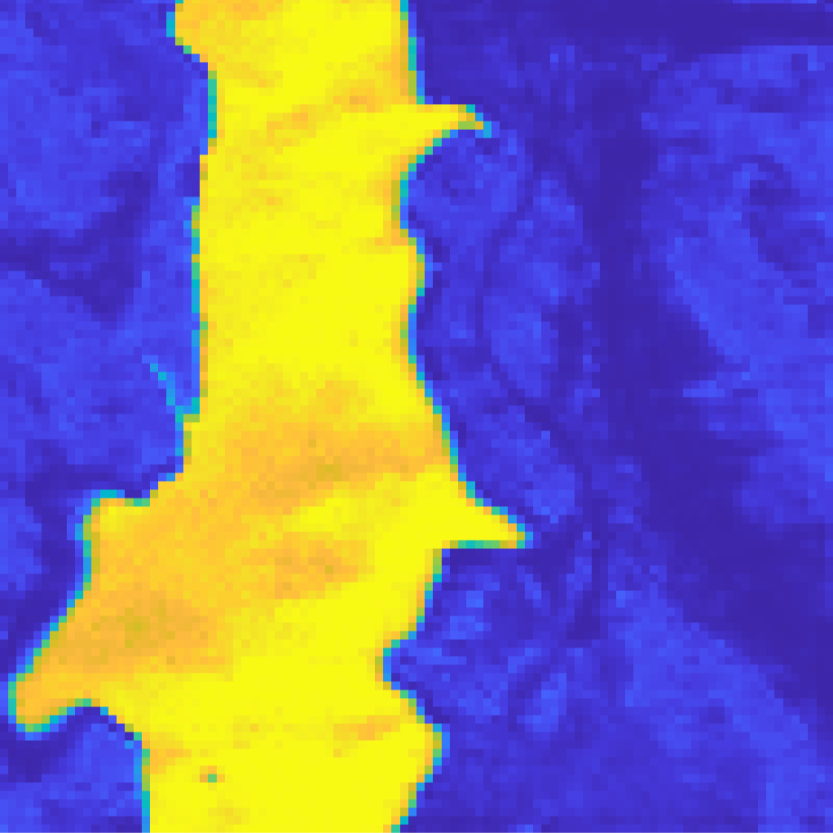}};
\filldraw [fill=white, draw=black] (0.2,-0.8) rectangle (1.35,-1.3);
\node at (0.2,-1.05) [anchor=west] {$23.91$dB};
\end{tikzpicture}\begin{tikzpicture}[outer sep=0pt,inner sep=1pt]\tikzstyle{every node}=[font=\small]
\node [align = center](img){\includegraphics[width=0.159\textwidth]{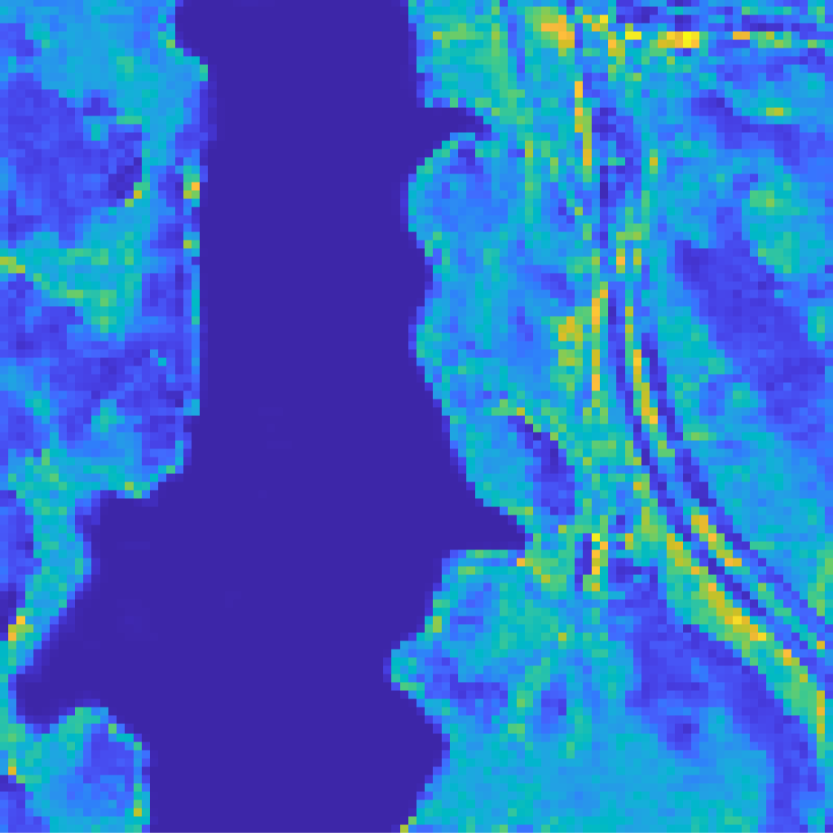}\baselineskip=1pt};
\filldraw [fill=white, draw=black] (0.2,-0.8) rectangle (1.35,-1.3);
\node at (0.2,-1.05) [anchor=west] {$15.00$dB};\end{tikzpicture}\hspace{-2.2pt}\begin{tikzpicture}[outer sep=0pt,inner sep=1pt]\tikzstyle{every node}=[font=\small]
\node [align = center](img){\includegraphics[width=0.159\textwidth]{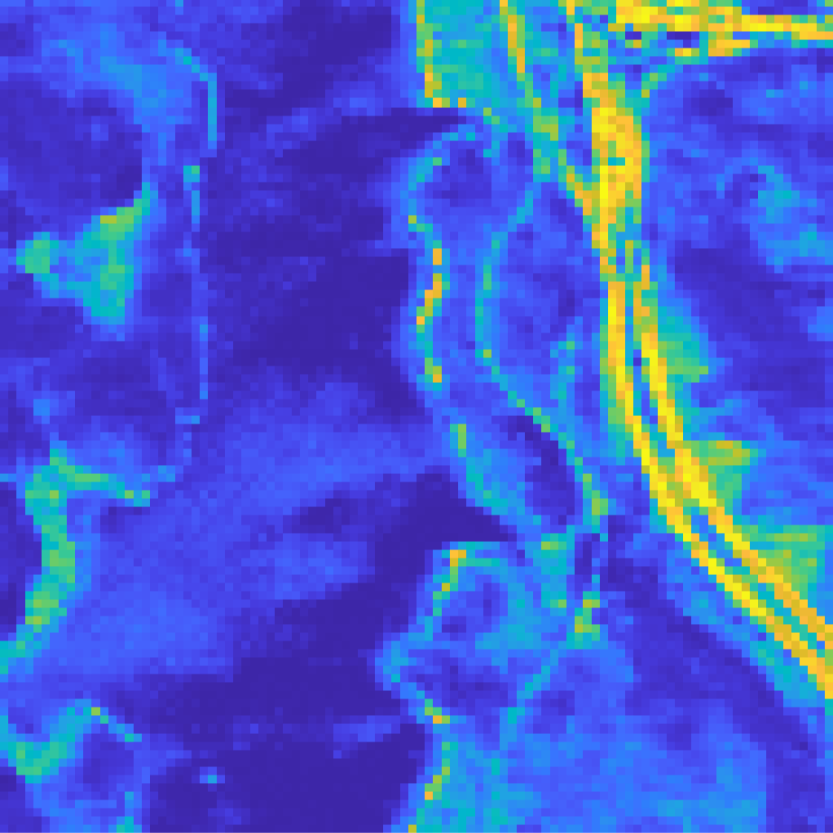}}; 
\filldraw [fill=white, draw=black] (0.2,-0.8) rectangle (1.35,-1.3);
\node at (0.2,-1.05) [anchor=west] {$15.63$dB};
\end{tikzpicture} 
\caption{Unmixing remote sensing task application resulting from the interpretable learned features and adjusted weights of Mixture-Net. RGB representation of the ground truth and the estimated recovered image, endmember, and abundance matrices.}
\label{fig:unmixing}
\end{figure*}

%% file: tikz/figure9.tex
\begin{figure}[tb!]
\begin{tikzpicture}
\tikzstyle{every node}=[font=\footnotesize]
\begin{scope}[node distance = 1mm, inner sep = 0pt, outer sep = 0pt,spy using outlines={rectangle, red, magnification=3.5, every spy on node/.append style={thick}}]
\node[align = center](img){\includegraphics[width=0.49\linewidth]{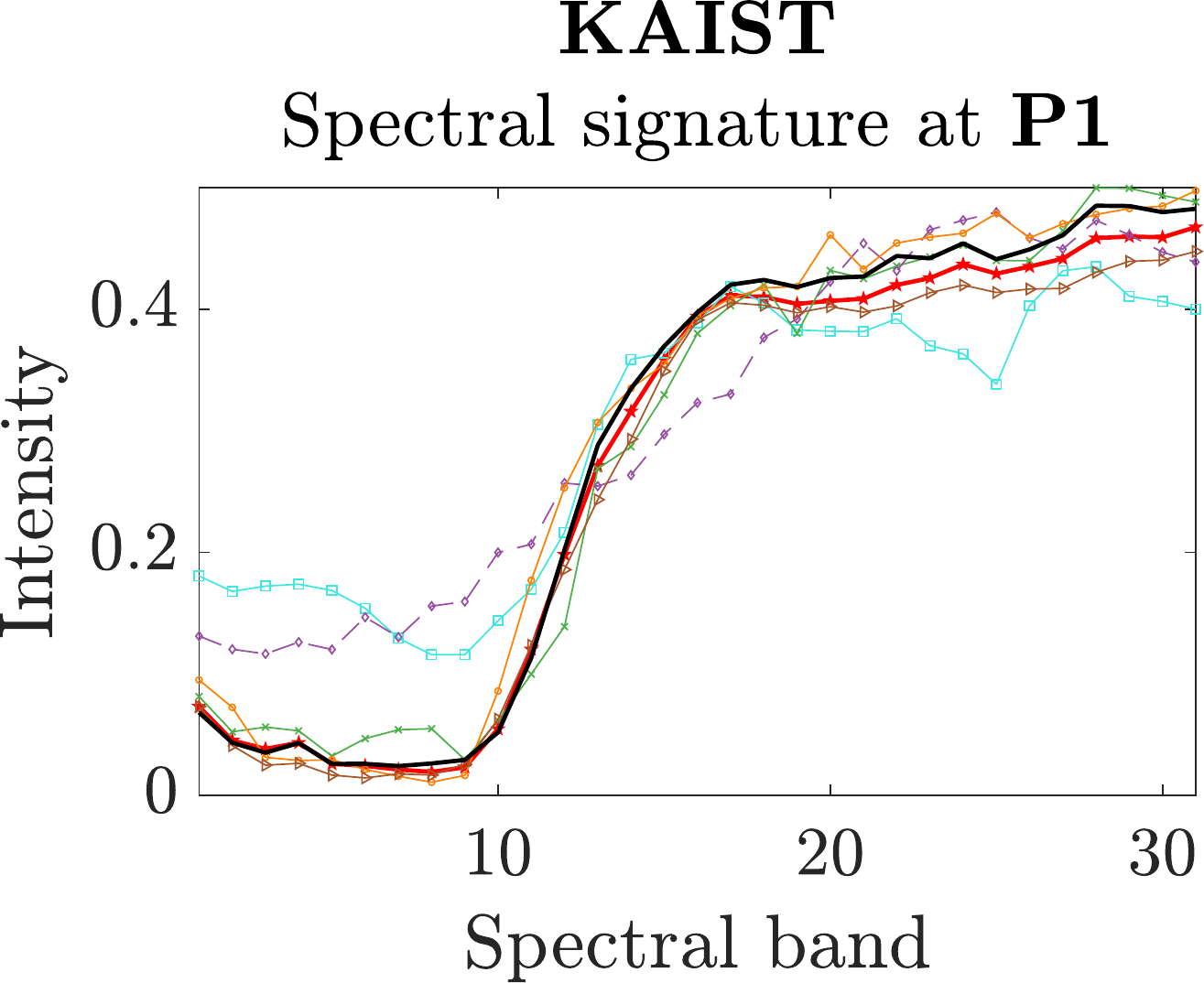}};
\spy [circle,dashed, blue, height=0.07\linewidth, width = 0.15\linewidth] on (0.0,0.1) in node at (1.3,-0.3);
\end{scope} 
\end{tikzpicture}
\begin{tikzpicture}
\tikzstyle{every node}=[font=\footnotesize]
\begin{scope}[node distance = 1mm, inner sep = 0pt, outer sep = 0pt,spy using outlines={rectangle, red, magnification=3.5, every spy on node/.append style={thick}}]
\node[align = center](img){\includegraphics[width=0.49\linewidth]{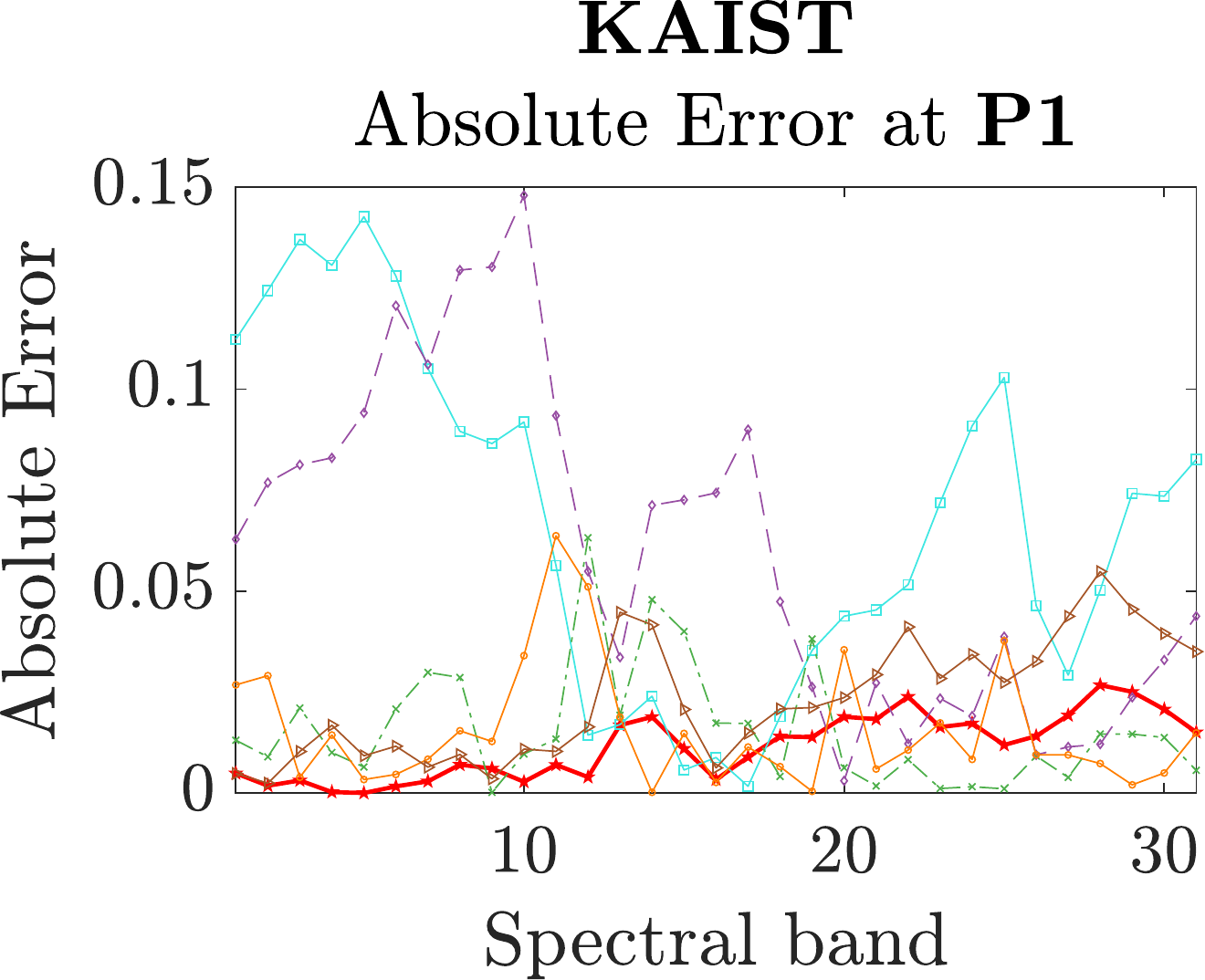}};
\end{scope} 
\end{tikzpicture}\\

\begin{tikzpicture}
\tikzstyle{every node}=[font=\footnotesize]
\begin{scope}[node distance = 1mm, inner sep = 0pt, outer sep = 0pt,spy using outlines={rectangle, red, magnification=3.0, every spy on node/.append style={thick}}]
\node[align = center](img){\includegraphics[width=0.49\linewidth]{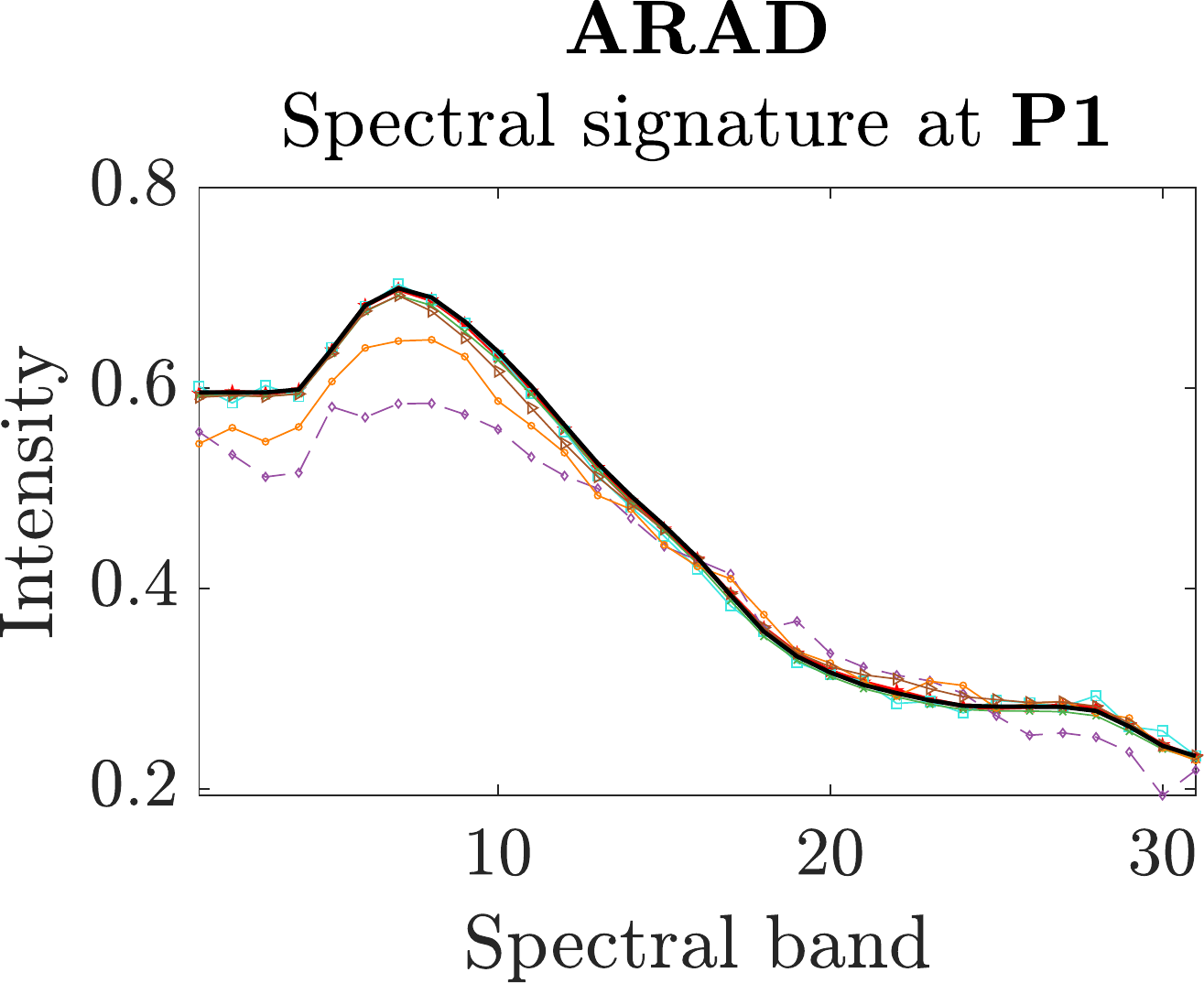}};
\spy [circle,dashed, blue, height=0.07\linewidth, width = 0.15\linewidth] on (-0.3,0.3) in node at (1.3,0.35);
\end{scope} 
\end{tikzpicture}
\begin{tikzpicture}
\tikzstyle{every node}=[font=\footnotesize]
\begin{scope}[node distance = 1mm, inner sep = 0pt, outer sep = 0pt,spy using outlines={rectangle, red, magnification=3.0, every spy on node/.append style={thick}}]
\node[align = center](img){\includegraphics[width=0.49\linewidth]{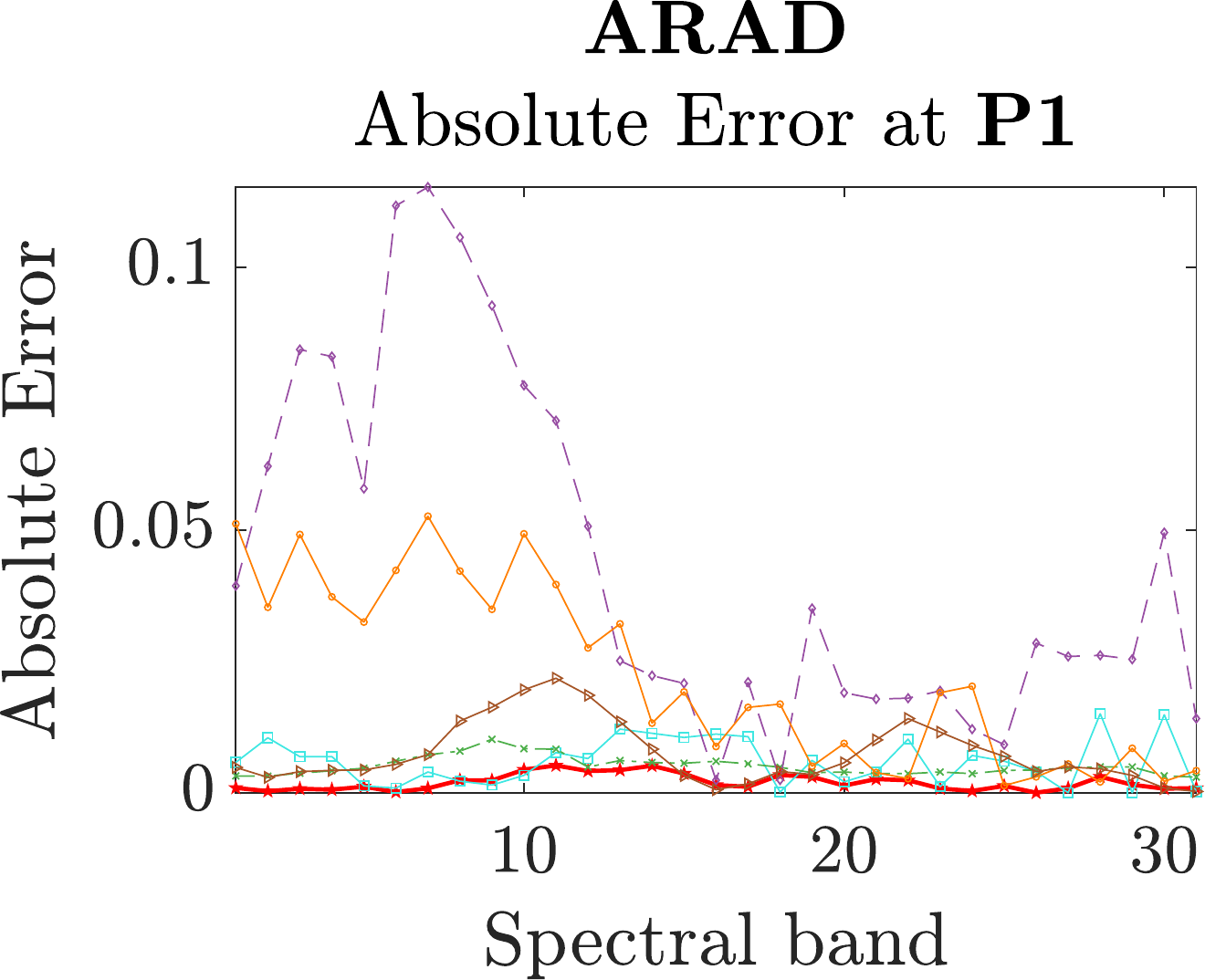}};
\end{scope} 
\end{tikzpicture}

\begin{tikzpicture}
\tikzstyle{every node}=[font=\footnotesize]
\node[align = center] at (0,0){\includegraphics[width=1.0\linewidth]{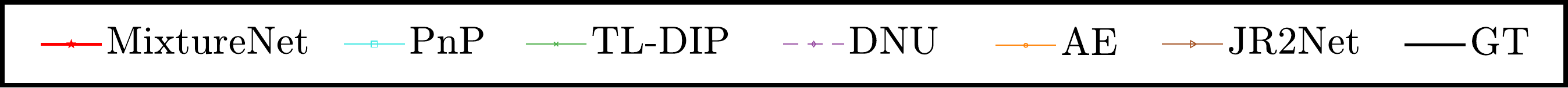}};
\end{tikzpicture}
\caption{Spectral signatures comparison at P1 for each dataset. The absolute error plots confirm that the spectral signatures obtained by the Mixture-Net method are more accurate than those obtained by the comparison methods.}
\label{fig:SpectralComparison}
\end{figure}

%% file: tikz/figure11.tex
\begin{figure}[tb!]
\centering
\includegraphics[width=1\linewidth]{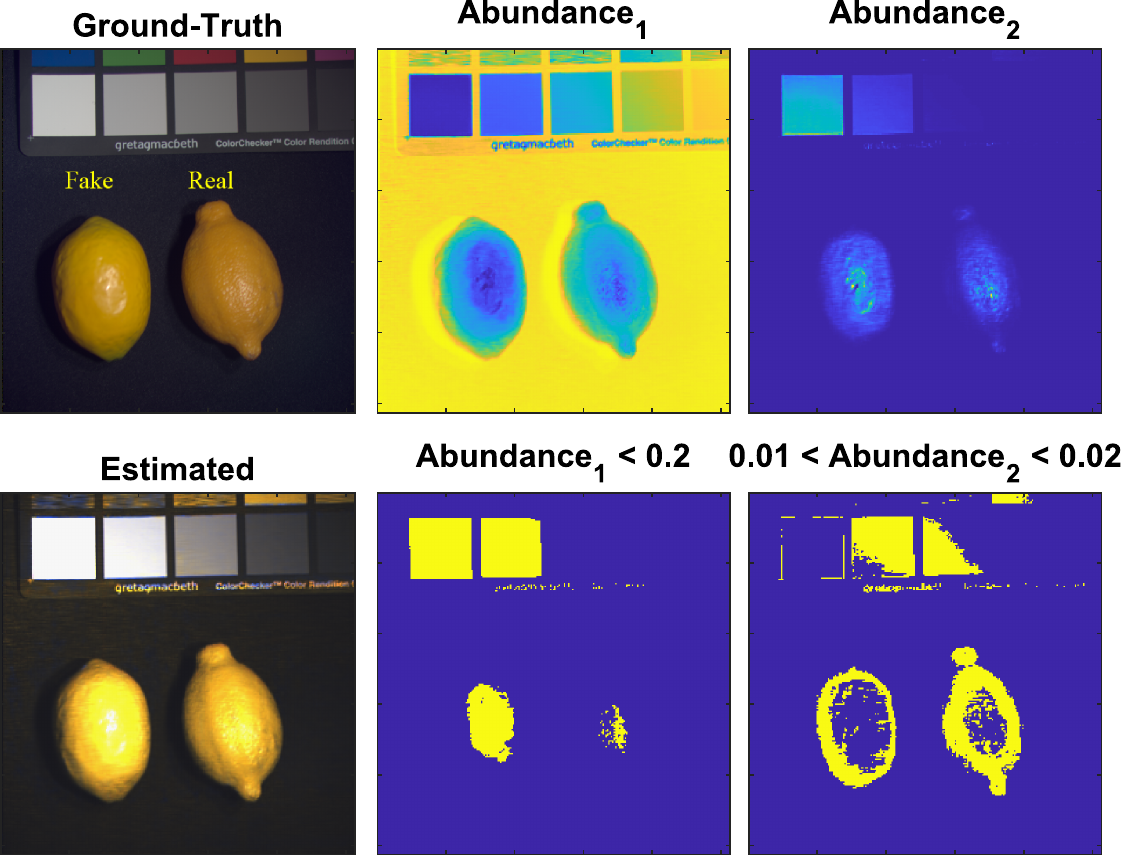}\caption{\textit{(left)} RGB Ground truth and estimated Fake image. \textit{(Top right)} Two obtained features interpreted as the abundances. \textit{(Bottom right)} Binary maps  after applying a threshold, identifying the differences in the materials. }\label{fig:kaistcomp}
\end{figure}

%% file: tikz/figure12.tex
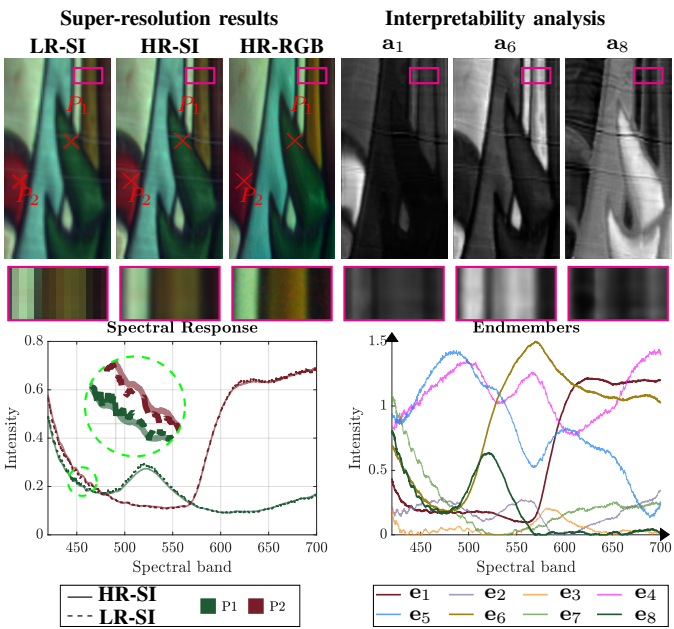
\begin{figure}[hb!]
\begin{tikzpicture}
\tikzstyle{every node}=[font=\footnotesize]
\node[anchor = west] at (0.1\linewidth,0){\textbf{Super-resolution results}};
\node[align = center] at (0.05\linewidth,0){\textcolor{white}{s}};
\end{tikzpicture}
\begin{tikzpicture}
\tikzstyle{every node}=[font=\footnotesize]
\node[anchor = west] at (0.15\linewidth,0){\textbf{Interpretability analysis}};
\node[align = center] at (0.05\linewidth,0){\textcolor{white}{s}};
\end{tikzpicture}
 	 
\begin{tikzpicture}
\tikzstyle{every node}=[font=\footnotesize]
\begin{scope}[node distance = 0mm, inner sep = 0pt, outer sep = 0,spy using outlines={rectangle, red, magnification=3.5, every spy on node/.append style={thick}}]
\node[align = center](img){\includegraphics[width=0.159\linewidth, height=0.30\linewidth]{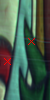}};
\spy [magenta, height=0.08\linewidth, width= .149\linewidth] on (0.4,1.1) in node at (0,-1.8);
\node at (0,1.5){\textbf{LR-SI}};
\node[red] at (0.03\linewidth,0.7){$P_1$};
\node[red] at (-0.4,-0.5){$P_2$};
\end{scope}
\end{tikzpicture}
\begin{tikzpicture}
\tikzstyle{every node}=[font=\footnotesize]
\begin{scope}[node distance = 0mm, inner sep = 0pt, outer sep = 0,spy using outlines={rectangle, red, magnification=3.5, every spy on node/.append style={thick}}]
\node[align = center](img){\includegraphics[width=0.159\linewidth, height=0.30\linewidth]{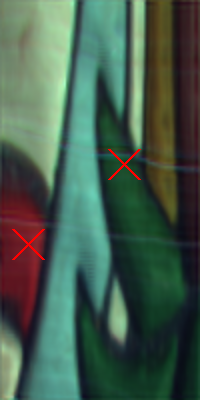}};
\spy [magenta, height=0.08\linewidth, width= .149\linewidth] on (0.4,1.1) in node at (0,-1.8);
\node at (0,1.5){\textbf{HR-SI}};
\node[red] at (0.03\linewidth,0.7){$P_1$};
\node[red] at (-0.4,-0.5){$P_2$};
\end{scope}
\end{tikzpicture}
\begin{tikzpicture}
\tikzstyle{every node}=[font=\footnotesize]
\begin{scope}[node distance = 0mm, inner sep = 0pt, outer sep = 0,spy using outlines={rectangle, red, magnification=3.5, every spy on node/.append style={thick}}]
\node[align = center](img){\includegraphics[width=0.159\linewidth, height=0.30\linewidth]{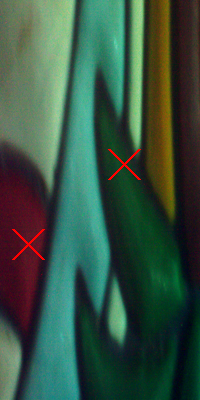}};
\spy [magenta, height=0.08\linewidth, width= .149\linewidth] on (0.4,1.1) in node at (0,-1.8);
\node at (0,1.5){\textbf{HR-RGB}};
\node[red] at (0.03\linewidth,0.7){$P_1$};
\node[red] at (-0.4,-0.5){$P_2$};
\end{scope}
\end{tikzpicture}
\begin{tikzpicture}
\tikzstyle{every node}=[font=\footnotesize]
\begin{scope}[node distance = 0mm, inner sep = 0pt, outer sep = 0,spy using outlines={rectangle, red, magnification=3.5, every spy on node/.append style={thick}}]
\node[align = center](img){\includegraphics[width=0.159\linewidth, height=0.30\linewidth]{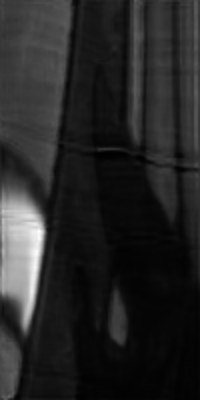}};
\spy [magenta, height=0.08\linewidth, width= .149\linewidth] on (0.4,1.1) in node at (0,-1.8);
\node at (0,1.5){$\mathbf{a}_1$};
\end{scope}
\end{tikzpicture}
\begin{tikzpicture}
\tikzstyle{every node}=[font=\footnotesize]
\begin{scope}[node distance = 0mm, inner sep = 0pt, outer sep = 0,spy using outlines={rectangle, red, magnification=3.5, every spy on node/.append style={thick}}]
\node[align = center](img){\includegraphics[width=0.159\linewidth, height=0.30\linewidth]{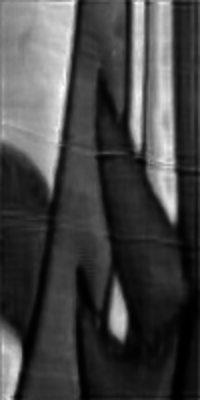}};
\spy [magenta, height=0.08\linewidth, width= .149\linewidth] on (0.4,1.1) in node at (0,-1.8);
\node at (0,1.5){$\mathbf{a}_6$};
\end{scope}
\end{tikzpicture}
\begin{tikzpicture}
\tikzstyle{every node}=[font=\footnotesize]
\begin{scope}[node distance = 0mm, inner sep = 0pt, outer sep = 0,spy using outlines={rectangle, red, magnification=3.5, every spy on node/.append style={thick}}]
\node[align = center](img){\includegraphics[width=0.159\linewidth, height=0.30\linewidth]{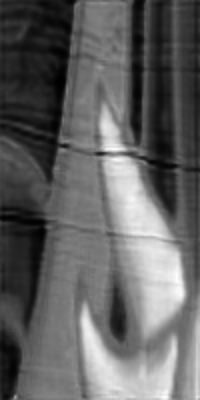}};
\spy [magenta, height=0.08\linewidth, width= .149\linewidth] on (0.4,1.1) in node at (0,-1.8);
\node at (0,1.5){$\mathbf{a}_8$};
\end{scope}
\end{tikzpicture}

\begin{tikzpicture}
\tikzstyle{every node}=[font=\footnotesize]
\begin{scope}[node distance = 0mm, inner sep = 0pt, outer sep = 0,spy using outlines={circle, red, magnification=3.5, every spy on node/.append style={thick}}]
\node[align = center](img){\includegraphics[width=1\linewidth]{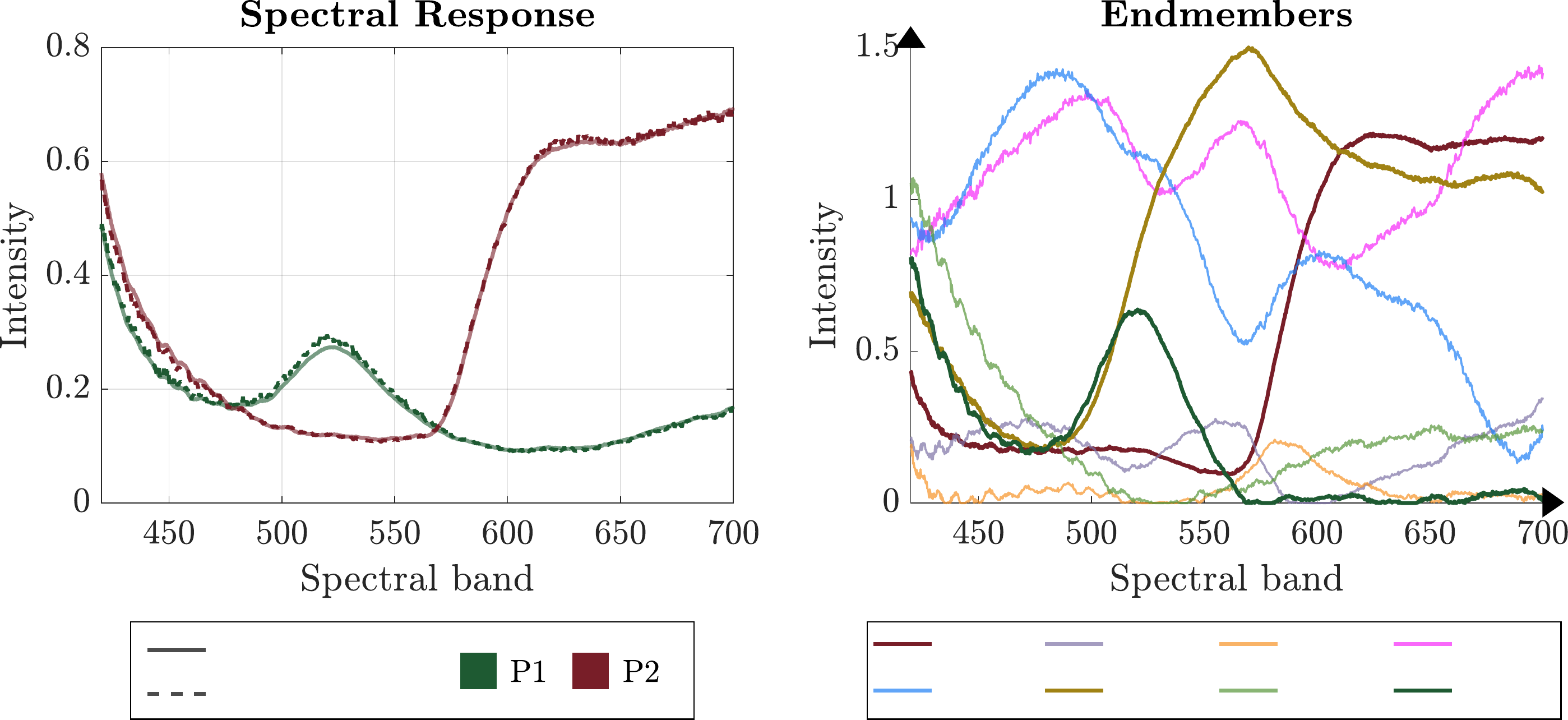}};
\spy [green, dashed, height=0.15\linewidth] on (-3.4,-0.1) in node at (-2.7,0.9);
\end{scope}
\node at (-2.8, -1.6){\textbf{HR-SI}};
\node at (-2.8, -1.9){\textbf{LR-SI}};
\node at (1.1, -1.6){$\mathbf{e}_1$};
\node at (2.1, -1.6){$\mathbf{e}_2$};
\node at (3.1, -1.6){$\mathbf{e}_3$};
\node at (4.1, -1.6){$\mathbf{e}_4$};
\node at (1.1, -1.9){$\mathbf{e}_5$};
\node at (2.1, -1.9){$\mathbf{e}_6$};
\node at (3.1, -1.9){$\mathbf{e}_7$};
\node at (4.1, -1.9){$\mathbf{e}_8$};
\end{tikzpicture}
\caption{\dcheck{Real-data super-resolution.\textit{(Left-top)} Spatial comparison of the estimated super-resolved image (HR-SI) against the reference RGB (HR-RGB) and the acquired SI (LR-SI).\textit{(Left-bottom)} LR-SI spectral signatures at two spatial locations $P_1 = (42,32)$ and $P_2 = (62,8)$ compared against the estimated HR-SI spectral signatures at the corresponding locations $P_1 = (493,373)$ and $P_2 = (733,85)$.\textit{(Right)} Mixture-Net interpretability analysis. The learned features and adjusted weights appear to be the abundance maps and the endmembers spanning the image. In particular, notice that the pairs $(\mathbf{a}_1,\mathbf{e}_1)$, $(\mathbf{a}_6,\mathbf{e}_6)$, $(\mathbf{a}_8,\mathbf{e}_8)$ can be interpreted as the maps and spectral signatures of the red, yellow, and green pixels.}}
\label{fig:RealDataCoregistered1}
\end{figure}